%% file: main.tex
\documentclass[10pt,twocolumn,letterpaper]{article}
\usepackage[T1]{fontenc}
\usepackage[pagenumbers]{cvpr}
\usepackage{amsthm}
\usepackage{tikz}
\usetikzlibrary{arrows.meta}
\graphicspath{{figures/}}
\definecolor{cvprblue}{rgb}{0.21,0.49,0.74}
\usepackage[breaklinks,colorlinks,allcolors=cvprblue]{hyperref}

\newtheorem{proposition}{Proposition}

\title{EgoOddball: Watching or Waiting?\\Auditing Early Warning in Robot Failure Detection}
\author{Zoe Li\textsuperscript{1}\thanks{Correspondence to Zoe Li at
\href{mailto:xli431@uw.edu}{\nolinkurl{xli431@uw.edu}}.}\quad
Tengwen Fan\textsuperscript{2}\quad Anze Wang\textsuperscript{3}\quad
Xu Chen\textsuperscript{1}\\[3pt]
\small \textsuperscript{1}MACS Lab, University of Washington\qquad
\textsuperscript{2}LAND Lab, University of Washington\\
\small \textsuperscript{3}Independent Researcher}
\date{}

\hypersetup{pdftitle={EgoOddball: Watching or Waiting? Auditing Early Warning in Robot Failure Detection},pdfauthor={Zoe Li, Tengwen Fan, Anze Wang, Xu Chen}}
\begin{document}
\maketitle
\input{sec0_abstract}
\input{fig_flowchart}

\section{Introduction}
\label{sec:intro}
Visual failure monitors should warn from camera evidence, such as slipping objects or trajectories drifting from success. The monitors audited here score visual embeddings, vision-language model (VLM) outputs or vision-language-action (VLA) representations. Under any-step alarms with stop-at-success rollouts, successful episodes can become unobservable earlier, so duration itself predicts outcomes. On PushT, a sensor-free calibrated clock reaches balanced accuracy (bAcc) .971 versus Failure Prediction at Runtime (FIPER)'s .712. Shortcut and Clever Hans predictors can score well without establishing the intended capability~\cite{geirhos2020shortcut,lapuschkin2019cleverhans}. Time is legitimate online information: what does vision add?

EgoOddball\footnote{\emph{Ego}: the robot's own observations and actions. \emph{Oddball}: as in the oddball paradigm~\cite{kappenman2021erpcore}, a warning should mark a departure from expectation, not elapsed time. Analogy only (Supp.~G).} contributes a reproducible attribution protocol: an analytic length-only ceiling, clocks inside each method's native pipeline including TWA, multi-landmark comparisons, task-mixture checks and held-out setting selection. A competitive clock measures what duration can earn without identifying a detector's internal mechanism.

SAFE~\cite{gu2025safe} and RESCUE~\cite{ahmedt2026rescue} already recognize duration effects; we do not claim their discovery. We test whether protocols change method preference or retain task-composition signals. Shortest-rollout truncation covers 16--24\% of the horizon on four FIPER tasks. At the quarter horizon, Sorting's ordering of five detectors and the clock reverses ($\tau_b=-.87$); the detector-only interval includes zero, and Stacking differs. SAFE's task-wise truncation removes within-task clock variation but leaves pooled clock area under the receiver operating characteristic curve (AUROC, also AUC) .587. These findings motivate horizon profiles and task references, without invalidating truncation.

The study makes four contributions:
\begin{itemize}
\item \textbf{Operational controls}: a length-only ceiling, native-pipeline clock, landmark risk sets and held-out selection (Sec.~\ref{sec:audit}).
\item \textbf{External audits}: five FIPER tasks plus Sentinel and SAFE. Clock bAcc is competitive on four FIPER tasks; it never significantly exceeds FIPER's TWA (Sec.~\ref{sec:experiments}).
\item \textbf{Protocol-sensitive comparisons}: same-metric ranks with uncertainty, task references and unscored insufficient risk sets (Sec.~\ref{sec:ranking}, exploratory).
\item \textbf{Mechanism and increment tests}: controlled termination-rule replication and frozen-monitor transfer tests; TEST-2 finds a small visual pipeline increment below its practical target (Sec.~\ref{sec:simpler}).

\end{itemize}
FIPER's TWA resisting the calibrated clock is a positive finding. More broadly, the audit can screen candidate monitors before deployment; it does not certify downstream safety.

\begin{figure*}[t]
\centering
\input{ranking_protocol_audit_20261109}
\caption{\textbf{Protocol sensitivity can alter method preference (exploratory).} (a) Five detectors plus the clock on Sorting: 14 of 15 pair orders reverse; the detector-only interval spans zero. (b) Fixed-landmark performance with 95\% episode-bootstrap intervals and success/failure risk counts; populations change, and invalid points are never interpolated. (c) GR00T image-distance proxy (not canonical FIPER): task-stratified AUROC with 95\% configuration-bootstrap intervals and success/failure counts. The any-step inset uses episode maxima; landmarks use the last logged score at or before $L$. Populations and summaries differ. Visual discrimination beyond a constant clock does not establish an increment over state/physics controls.}
\label{fig:findings}
\end{figure*}

\section{Related Work}
\paragraph{Runtime failure prediction for robot policies.}
FIPER~\cite{romer2025fiper} combines random network distillation (RND) in observation embeddings (RND-OE) with action-chunk entropy (ACE); statistical temporal action consistency (STAC)~\cite{agia2024stac} monitors temporal consistency. Other approaches use successful-rollout uncertainty~\cite{xu2025faildetect}, learned dynamics~\cite{liu2024siriusrm}, language reasoning~\cite{sinha2024realtime}, VLM explanations~\cite{duan2025aha,ye2025robofac}, conformal help/switching rules~\cite{ren2023robots,huang2023conformalpolicy}, or belief gating over traceable failure recall~\cite{li2026coresense}. SAFE~\cite{gu2025safe} truncates each task to its shortest rollout. RESCUE~\cite{ahmedt2026rescue} reports a chunk-boundary counter with AUROC 1.000 and an early-exposure control reducing a progress monitor from .979 to .637. Mechanistic Interpretability for Neural Diagnosis (MIND)~\cite{ahmed2026mind} compares sparse-feature monitors with a success-length alarm rule and scores RoboTwin on opening frames; ActProbe~\cite{huang2026actprobe} reads emitted action chunks. We audit evaluation credit, complementing these detector designs.

\paragraph{Shortcuts and evaluation artifacts.}
Zhang and An~\cite{zhang2026memoryclock} show that time-indexed VLA memory can confuse waiting with task progress. We study duration-dependent credit in monitor evaluation, rather than policy memory. Elapsed time remains legitimate online information, distinct from evaluation leakage~\cite{kapoor2023leakage}.

\paragraph{Information available at decision time.}
Streaming perception evaluates a prediction against the world at the time it is delivered~\cite{li2020streaming}, and early event detection trains and scores detectors on partial sequences~\cite{hoai2014mmed}. Landmark analysis in survival statistics conditions on subjects still at risk at a fixed time~\cite{anderson1983landmark,vanhouwelingen2007landmark}; time-dependent ROC methods define time-specific sensitivity and specificity~\cite{heagerty2005survival}, and concordance indices compare only pairs ordered in time~\cite{harrell1982yield,uno2011cstatistics}; our common observation window applies this idea to alarm evaluation. $\pi_{0.7}$ conditions training on speed, quality and mistake labels, using desired values at inference~\cite{pi2026pi07}; this legitimate supervision motivates monitor attribution controls.

\input{sec3_problem}
\input{sec4_method}
\input{sec5_experiments}

\section{Discussion and Limitations}
\paragraph{Level 1: useful predictive increment.}
A useful visual monitor must warn beyond task/time, available state/action history and nominal-physics residuals~\cite{lyu2026adapt}. The comparison should hold populations, supervision and model capacity fixed, with models, thresholds and cadence frozen before new episodes. Lead time should be measured after processing and delivery, against reviewed onset intervals, and reported with false-alarm burden, missing-image coverage and a state-only fallback. Dens tests state-transition surprise. Pixel-level errors from world or world-action models remain prospective visual signals and need forecast-span, update-cadence, clock and landmark controls (Supp.~T,Y). TEST-2 establishes a small pipeline increment within fixed configurations; useful failure-onset warning remains unconfirmed.

\paragraph{Levels 2--3: reliance and intervention utility.}
Only after Level 1 should controlled masking, stale-image and missing-image tests ask whether the monitor actually depends on vision. Holding the physical trajectory fixed can separate extra image evidence from visually estimated state. Evidence-gated execution interfaces~\cite{li2026pier} separate such checks from control; their utility still requires a controlled response study measuring costs, errors and delays (Supp.~T). Current DEV sensitivity and mechanism probes establish neither visual reliance nor intervention utility.

\paragraph{Policy efficiency and exposure.}
Stochastically shorter successes with unchanged failure lengths cannot reduce the clock ceiling (Prop.~\ref{prop:clock}; Supp.~T). Higher throughput~\cite{pi2026pi07} alone establishes neither this shift nor improved calibrated-clock performance; these premises remain untested for $\pi_{0.7}$.

\paragraph{Scope of the evidence.}
The clock's floating-point tie rule was fixed after inspection; learned-method cells are unchanged. Two FIPER tasks have only 20 test rollouts. Original intervals fix scores, settings and calibration, and therefore omit training and unknown acquisition clustering. The 25 releases are not independent score streams, and rank/task-mixture checks are exploratory without multiplicity correction. SAFE learned detectors were not rerun; real-robot evidence is released offline data. The first frozen-monitor test fails its risk-set gate. We establish no closed-loop or physical-safety benefit.

\section{Conclusion}Outcome-dependent exposure can inflate failure-warning accuracy. EgoOddball combines ceilings, native-pipeline clocks, landmarks and held-out selection to audit this effect. GR00T confirms the termination mechanism; TEST-2 finds a small visual pipeline increment below its practical target. The audit supports monitor comparison and evidence-based selection before deployment in structured or unstructured environments. Useful failure-onset warning, visual reliance and intervention benefits require separate evidence; these deployments were not evaluated.

\section*{LLM Usage Statement}
LLM-based tools (OpenAI ChatGPT/Codex and Anthropic Claude) were used under the authors' direction for implementation, experimentation support, figure preparation, writing and reviewer-style feedback. The authors determined the research design and conclusions, verified all results, and take full responsibility for this preprint.

\bibliographystyle{ieeenat_fullname}
\bibliography{refs}

\input{supplementary-analyses}

\end{document}

%% file: sec0_abstract.tex
\begin{abstract}
Robot failure monitors should earn credit for visual evidence, yet standard evaluation can also reward waiting: when alarms count at any step and rollouts stop at success, failures can remain observable longer than successes. EgoOddball audits this outcome-dependent exposure. It places a sensor-free clock inside each detector's calibration and alarm pipeline, bounds what episode length alone can earn, and compares methods on fixed-landmark risk sets and held-out settings, separating duration-driven scores from information available at decision time.

On five reproduced Failure Prediction at Runtime (FIPER) tasks, the clock matches or exceeds FIPER's balanced accuracy on four (PushT: .971 versus .712), yet never significantly exceeds its time-weighted accuracy. Across 25 released rollout sets from FIPER, Sentinel and SAFE, including real-robot rollouts, the length-only balanced-accuracy ceiling is at least .76 in 21. In a prespecified replication with a frozen GR00T N1.7 policy on 816 new episodes, changing only the termination rule raises clock-only balanced accuracy from .500 to .965 (difference .465; 95\% CI [.446,.481]). Frozen-monitor tests then quantify the visual pipeline increment: after a first 416-episode test fails its risk-set gate, a second 1,248-episode test finds $+.0035$ AUROC [$+.0007,+.0062$] beyond state/action feedback, below the prespecified $.02$ target (Carrot/Spoon, 24 fixed configurations per task).

Useful failure-onset warning and intervention benefits remain unestablished. EgoOddball gives monitor developers a reproducible check before claiming that alarms reflect vision rather than elapsed time.
\end{abstract}

%% file: fig_flowchart.tex
\begin{figure*}[t]
\centering
\begingroup
\definecolor{eoInk}{HTML}{263445}
\definecolor{eoBlue}{HTML}{0072B2}
\definecolor{eoOrange}{HTML}{D55E00}
\definecolor{eoPurple}{HTML}{655C96}
\definecolor{eoLine}{HTML}{CDD5DF}
\resizebox{\textwidth}{!}{%
\begin{tikzpicture}[x=1cm,y=1cm,font=\sffamily\fontsize{8}{9.5}\selectfont,text=eoInk,
 >={Stealth[length=4pt,width=3pt]},
 card/.style={draw=eoLine,fill=white,rounded corners=3pt,line width=.6pt},
 flow/.style={->,draw=eoInk!65,line width=.8pt},
 title/.style={anchor=west,font=\sffamily\bfseries\fontsize{9.5}{11}\selectfont},
 txt/.style={align=center,inner sep=2pt}]
\path[use as bounding box] (0,0) rectangle (17.4,6.45);
\node[title] at (.05,6.23) {(a) Why scores can mislead};
\node[title] at (4.80,6.23) {(b) EgoOddball attribution audit};
\node[title] at (11.65,6.23) {(c) What the evidence supports};
\draw[card,fill=eoOrange!3] (0,.60) rectangle (4.30,5.93);
\draw[card,fill=eoPurple!3] (4.75,.60) rectangle (11.15,5.93);
\draw[card] (11.60,.60) rectangle (17.40,5.93);

\node[txt] at (2.15,5.66) {\textbf{Sentinel PushChair: real robot}};
\node[inner sep=0pt] at (2.15,5.10) {\includegraphics[width=3.96cm]{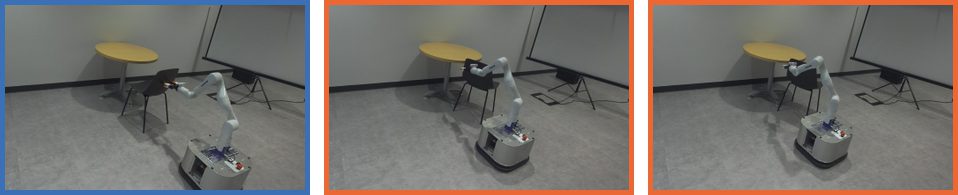}};
\node[txt] at (.82,4.33) {success\\ends: 16};
\node[txt] at (2.15,4.33) {failure\\alarm: 24};
\node[txt] at (3.48,4.33) {failure\\ends: 48};
\node[txt] at (2.15,3.87) {\textbf{One clock, unequal exposure}};
\node[anchor=east] at (1.02,3.44) {success};
\node[anchor=east] at (1.02,2.95) {failure};
\draw[eoInk!50,line width=2pt] (1.13,3.44)--(2.11,3.44);
\draw[eoInk,line width=2pt] (1.13,2.95)--(4.07,2.95);
\draw[eoOrange,dashed,line width=.8pt] (2.60,2.67)--(2.60,3.72);
\fill[eoOrange] (2.60,2.95) circle (2pt);
\node[anchor=north,text=eoOrange] at (2.60,2.68) {alarm at 24};
\draw[eoLine,->] (1.13,2.26)--(4.10,2.26);
\foreach \x/\v in {1.13/0,2.11/16,4.07/48}
 {\draw[eoLine] (\x,2.21)--(\x,2.31);\node[anchor=north] at (\x,2.18) {\v};}
\node[anchor=east] at (1.02,2.26) {step};
\draw[eoLine] (.20,1.79)--(4.10,1.79);
\node[txt] at (2.15,1.48) {\textbf{PushT: full-rollout bAcc}};
\node[txt,text=eoOrange] at (2.15,1.06) {Clock \textbf{.971}\quad vs.\quad FIPER \textbf{.712}};

\node[txt,text=eoPurple] at (7.95,5.64) {\textbf{A1: same procedure, different inputs}};
\node[card,draw=eoBlue,fill=eoBlue!5,txt,minimum width=3.05cm,minimum height=.76cm] (v) at (6.55,5.08) {};
\node[txt] at (6.79,5.08) {\textbf{Visual detector}\\available prefix only};
\draw[eoBlue,line width=.8pt,rounded corners=.5pt] (5.13,4.95) rectangle (5.51,5.20);
\draw[eoBlue,line width=.8pt] (5.20,5.20)--(5.24,5.26)--(5.40,5.26)--(5.44,5.20);
\draw[eoBlue,line width=.8pt] (5.32,5.075) circle (.075);
\node[card,draw=eoOrange,fill=eoOrange!5,txt,minimum width=3.05cm,minimum height=.76cm] (c) at (6.55,4.24) {};
\node[txt] at (6.79,4.24) {\textbf{Sensor-free clock}\\$c(t)=t+1$};
\draw[eoOrange,line width=.8pt] (5.32,4.24) circle (.18);
\draw[eoOrange,line width=.9pt,line cap=round] (5.32,4.36)--(5.32,4.24)--(5.42,4.19);
\node[card,txt,minimum width=2.48cm,minimum height=1.30cm] (p) at (9.70,4.66) {Native calibration\\and alarm rule\\\textbf{bAcc / TWA}};
\draw[flow,eoBlue] (v.east)--(v.east -| p.west);
\draw[flow,eoOrange] (c.east)--(c.east -| p.west);
\draw[flow] (4.30,5.08)--(4.95,5.08);
\draw[flow] (11.15,4.66)--(11.60,4.66);
\draw[card] (4.97,2.90) rectangle (10.93,3.68);
\node[anchor=west] at (5.12,3.29) {\textbf{Step 0: ceiling}};
\node[anchor=east] at (10.78,3.29) {bAcc $\leq\frac12(1+D^+)$};
\draw[card] (4.97,1.90) rectangle (10.93,2.72);
\node[anchor=west,align=left] at (5.12,2.31) {\textbf{A2: landmarks}\\and task strata};
\node[anchor=east,align=right] at (10.78,2.31) {Compare at the same $L$\\among rollouts still running};
\draw[card] (4.97,.89) rectangle (10.93,1.71);
\node[anchor=west,align=left] at (5.12,1.30) {\textbf{A3: held-out}\\setting selection};
\node[anchor=east,align=right] at (10.78,1.30) {Select on $A$, score on $B$\\then exchange folds};

\node[anchor=west,font=\sffamily\bfseries\fontsize{8.5}{10}\selectfont] at (11.82,5.61) {Termination mechanism confirmed};
\node[txt] at (14.50,5.22) {GR00T: 816 new episodes};
\node[txt,text=eoOrange,font=\sffamily\bfseries\fontsize{12}{13}\selectfont] at (14.50,4.76) {.500 $\longrightarrow$ .965};
\node[txt] at (14.50,4.35) {Clock bAcc; only termination changes};
\draw[eoLine] (11.82,4.04)--(17.18,4.04);
\node[anchor=west,font=\sffamily\bfseries\fontsize{8.5}{10}\selectfont] at (11.82,3.76) {Visual increment: small, bounded};
\node[txt] at (14.50,3.39) {TEST-2: 1,248 episodes; Carrot/Spoon};
\node[txt,text=eoBlue] at (14.50,3.02) {$\Delta$AUROC $+.0035$\quad 95\% CI $[.0007,.0062]$};
\draw[eoLine] (12.20,1.96)--(16.60,1.96);
\draw[eoInk!50,dashed] (12.20,1.96)--(12.20,2.68);
\draw[eoOrange,dashed] (16.20,1.96)--(16.20,2.68);
\draw[eoBlue,line width=1.6pt] (12.345882,2.36)--(13.430397,2.36);
\draw[eoBlue] (12.345882,2.26)--(12.345882,2.46);
\draw[eoBlue] (13.430397,2.26)--(13.430397,2.46);
\fill[eoBlue] (12.893348,2.36) circle (2.5pt);
\node[anchor=north] at (12.20,1.90) {0};
\node[anchor=north] at (14.20,1.90) {.01};
\node[anchor=north,text=eoOrange] at (16.20,1.90) {.02};
\node[anchor=south,text=eoOrange] at (16.20,2.49) {target};
\node[txt] at (14.50,1.30) {Beyond state/action; below practical target};
\node[txt] at (14.50,.93) {24 fixed configurations per task};

\node[anchor=west,text=eoPurple,font=\sffamily\bfseries\fontsize{8}{9.5}\selectfont] at (.08,.23) {Use: compare monitors before deployment};
\node[anchor=east] at (17.32,.23) {Still unestablished: useful onset warning $\cdot$ visual reliance $\cdot$ intervention benefit};
\end{tikzpicture}%
}
\endgroup

\caption{\textbf{Watching or waiting? From exposure bias to attributable evidence.} (a) Selected real-robot PushChair frames and recorded lengths from Sentinel~\cite{agia2024stac}; a calibration threshold of 22.2 first alarms at step 24. The PushT comparison is a separate FIPER result. (b) Complementary controls: a descriptive length-only ceiling (Prop.~\ref{prop:clock}), native-pipeline clocks, within-task landmarks and held-out selection. (c) GR00T confirms the termination mechanism. Separately, TEST-2 shows a small visual pipeline increment beyond state/action, below its prespecified practical target; the interval is conditional on the fixed configurations. These results do not establish useful failure-onset warning or intervention benefits.}
\label{fig:overview}\label{fig:protocol}
\end{figure*}

%% file: ranking_protocol_audit_20261109.tex
\definecolor{eaBlue}{HTML}{0072B2}
\definecolor{eaOrange}{HTML}{D55E00}
\definecolor{eaPurple}{HTML}{CC79A7}
\definecolor{eaGreen}{HTML}{009E73}
\definecolor{eaSky}{HTML}{56B4E9}
\definecolor{eaGray}{HTML}{444444}
\includegraphics[width=\textwidth]{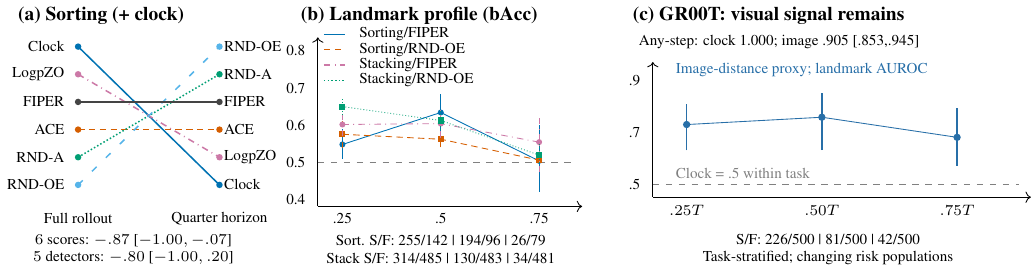}

%% file: sec3_problem.tex
\section{Decision-Time Evidence and Claims}
\label{sec:problem}
Consider an operator receiving an alarm at decision time $t$. Elapsed time is already known; the scene evidence and feedback available then determine what additional information can justify that alarm. A later feature or retrospective label cannot be used to reconstruct an unrealistically informed online decision.

Evidence $e_i=(x_i,\tau_i^{\rm obs},\tau_i^{\rm avail})$ separates the represented event from the time its required processing completes. A target or outcome $y_j$ has feedback time $b_j$. The decision may use only
\begin{equation}
\mathcal I_t=\{e_i:\tau_i^{\rm avail}\le t\}\cup\{(j,y_j):b_j\le t\}.
\label{eq:info_set}
\end{equation}
For raw measurements, availability can be receipt; for a computed feature, it is feature readiness, not the earlier image receipt time. Models and calibration statistics depend only on this information and designated training/development data. Later arrivals cannot enter retroactively. Alarm utility in the control loop is a separate endpoint from offline discrimination; it is not established by this audit.

\paragraph{Prediction, reliance and utility.}
We distinguish three evidence levels (Table~\ref{tab:claim-levels}). A predictive increment asks what a specified visual pipeline adds conditional on available time, state and action inputs. Actions can already carry policy vision, so this is not the causal value of all vision. Reliance requires interventions on observations; utility requires a controlled response study. The audit addresses the first level's attribution problem, without assuming either later conclusion.
\input{claim_levels_20261114}

A world-action model (WAM) may propose actions and a future image. That image is a prediction, not a new direct observation. Its residual becomes usable only after the matching real image arrives and comparison completes. Source cutoff, readiness, decision and target times stay separate (Supp.~T).

\paragraph{Observation horizon is a separate estimand.}
For the duration audit, let $T$ be the observed rollout length in decision steps, with zero-based indices $0,\ldots,T-1$, and $Y$ its eventual outcome. The elapsed-count clock $c(t)=t+1$ with integer threshold $q$ satisfies
\begin{equation}
\Pr\!\left(\max_{0\le t<T}c(t)>q\mid Y=y\right)=\Pr(T>q\mid Y=y).
\label{eq:clock-exposure}
\end{equation}
An any-step clock alarm therefore separates outcomes exactly as well as their observation durations do. If successful rollouts stop at success and failed ones run to a cap, a clock can look like an early-warning detector. This is an exposure effect, not future-input leakage: elapsed time is available online. It means that a protocol's termination rule is part of what a detector is credited for.

\begin{proposition}[Clock ceiling]
\label{prop:clock}
Let $F_S$ and $F_F$ be the distribution functions of the scored lengths of successful and failed test rollouts, with any-step alarms and no patience. A detector whose alarm depends only on the step index has balanced accuracy at most $\tfrac12(1+D^+)$ with $D^+=\max_\tau[F_S(\tau)-F_F(\tau)]$, the one-sided Kolmogorov--Smirnov distance, and attains it by alarming at the maximizing step. With detection time normalized by $T_{\max}-1$, its TWA is at most $\max_\tau\tfrac12\big[(1-\tfrac{\tau}{T_{\max}-1})(1-F_F(\tau))+F_S(\tau)\big]$.
\end{proposition}
\noindent\emph{Proof.} With true-positive rate (TPR) and true-negative rate (TNR), such a detector alarms at a fixed step $\tau$ on exactly the rollouts with $T>\tau$, so $\mathrm{TPR}=1-F_F(\tau)$ and $\mathrm{TNR}=F_S(\tau)$ (never alarming gives $0.5$); maximize over $\tau$. $\square$

The ceiling needs no detector: it is computable from rollout lengths alone, before any method is run, and states how much of an any-step score duration could explain.

\paragraph{Landmarks neutralize the specified clock.}
At a common observation-count landmark $L$, restrict evaluation within task to the risk set $\{T>L\}$ and ask whether the detector has alarmed by $L$. The elapsed-count clock gives every rollout the same decision, so its balanced accuracy is exactly $0.5$. This does not remove progress, task priors or variable wall-clock availability; coarse time bins need not even neutralize the specified clock.

%% file: claim_levels_20261114.tex
\begin{table}[t]
\centering\footnotesize
\setlength{\tabcolsep}{3pt}
\begin{tabular}{p{.21\linewidth}p{.73\linewidth}}
\toprule
Claim & Required evidence\\
\midrule
1. Predictive increment & Frozen visual versus time/state/physics controls on identical future risk sets; reviewed onset and delivered lead for an early-warning claim.\\
2. Visual reliance & Paired observation interventions with the physical trajectory and other inputs fixed; matched corruption controls and unavailable-image cases.\\
3. Intervention utility & A separately controlled response study measuring outcomes and costs, including response errors and delays.\\
\bottomrule
\end{tabular}
\caption{\textbf{Three distinct evidence levels.} Prediction, reliance and utility answer different questions. A score increment does not establish the latter two; image sensitivity alone can reflect harmful corruption. This paper audits predictive attribution and does not confirm useful visual early warning, causal visual reliance or intervention utility.}
\label{tab:claim-levels}
\end{table}

%% file: sec4_method.tex
\section{Audit Protocol}
\label{sec:audit}
The protocol leaves the audited method unchanged and adds three references. Original audits follow plans hashed before their metrics were read; the ranking extension below was specified after earlier results were inspected and is exploratory. Later changes are dated amendments.

\paragraph{A1: clock reference inside the method's own pipeline.}
We add a method whose per-step score is the elapsed step count. It reads no sensor, embedding or action. It passes through the identical window aggregation, calibration rollouts, threshold styles, quantiles, metrics and setting selection as the audited method. Its behavior follows from Eq.~\ref{eq:clock-exposure}: with a constant threshold it alarms after a calibration quantile of successful-rollout lengths; with a time-varying threshold it alarms after the longest calibration success. The clock is a reference for how much of a score the termination rule alone can produce, not a competitor.

\paragraph{A2: common observation window.}
For each task with maximum test length $T_{\rm task}$, landmarks are $L=\lfloor fT_{\rm task}\rfloor$ for $f\in\{0.25,0.5,0.75\}$. The risk set contains test rollouts with length $>L$, labeled by their final outcome. A method alarms by $L$ if its normalized score exceeds its threshold at a step $<L$, using its selected setting. We report balanced accuracy in the risk set only when it contains at least five successes and five failures. Rollouts that ended before $L$ are counted by outcome, not deleted or padded.

\paragraph{Continuous scores, ranks and task mixture.}
For available continuous scores, the primary prefix summary is $S_i(L)=\max_{t<L}s_i(t)$; the last available score before $L$ is a sensitivity analysis, not an interpolated observation. Within task we estimate
\begin{equation}
\begin{aligned}
\mathrm{AUC}_L=\Pr[&S_i(L)>S_j(L)\mid\\
&Y_i=1,Y_j=0,T_i>L,T_j>L],
\end{aligned}
\end{equation}
with half credit for ties. This is final-outcome discrimination within a landmark risk set, inspired by landmark analysis~\cite{vanhouwelingen2007landmark}; it is not the event-time cumulative/dynamic survival AUC. Binary-only monitor outputs contribute balanced accuracy, not continuous-score AUROC. We compare like metrics using Kendall $\tau_b$ and pairwise rank reversals. Each landmark changes the population and question; a reversal does not establish which method optimizes deployment utility.

Pooled AUROC is accompanied by task-wise scores and equal-task macro AUROC. We also report a task-prior reference and a ridge logistic task-plus-elapsed-step reference, both obtained by fixed five-fold episode cross-fitting. These diagnostic references use labeled episodes in other folds, not successful-only calibration, and establish no unseen-task generalization. Their AUCs are not additive contributions. Pooled and macro comparisons identify when eligible task sets differ; matched-task pooled scores are supplied as a control.

\paragraph{A3: selection without test reuse.}
If window and threshold type are chosen on test-rollout metrics, the reported number includes selection optimism. We split test rollouts into two folds by a hash of task and file name, select on one fold with the original criterion, evaluate on the other, and average. The comparator is the original selection evaluated on the same folds, so the two arms differ only in which rollouts select. We call the difference selection-procedure sensitivity.

\paragraph{Uncertainty.}
We report 95\% confidence intervals (CIs). The original audit uses 2000 outcome-stratified episode-bootstrap draws; the ranking extension uses 1000. Shared original-episode draws are restricted to each risk set; seeds are averaged, not independent rollouts. Original intervals fix scores, calibration and selected settings, excluding their uncertainty and unknown acquisition clustering. Supp.~X adds exploratory setting reselection, still fixing training and calibration.

\begin{center}
\fbox{\begin{minipage}{.92\columnwidth}\small
\textbf{Before crediting a visual early-warning score}\par
1. State termination rules, caps and outcome-specific cap shares.\par
2. Report the length-only ceiling (Prop.~\ref{prop:clock}).\par
3. Run the calibrated clock through the same pipeline (A1).\par
4. Report landmark bAcc/AUROC, risk counts and rank uncertainty (A2); distinguish pooled and task-macro scores.\par
5. Disclose test-based selection and its held-out estimate (A3).
\end{minipage}}
\end{center}
Items 1--2 need no detector. Missing score streams cannot support ranking. No checklist item alone establishes deployment or intervention value.

\subsection{Outcome signals for \mbox{SIMPLER}}
Dens uses five heteroscedastic multilayer perceptrons (MLPs) (width 64, 20 epochs) that predict joint and end-effector increments from the preceding state and executed action. On fully observed transitions,
\begin{equation}
z_t^2=\frac{1}{d}\sum_{j=1}^d\frac{(\Delta x_{t,j}-\bar\mu_{t,j})^2}{v^{\rm data}_{t,j}+v^{\rm model}_{t,j}}.
\label{eq:dens-score}
\end{equation}
Five features summarize the last ten transitions. The baseline A comprises task indicators, elapsed step and kinematics. A and Dens use balanced logistic classifiers fitted within ten configuration-group folds; intervals resample saved out-of-fold scores.

%% file: sec5_experiments.tex
\section{Experiments}
\label{sec:experiments}
The Octo learned-signal contrast and GR00T termination replication follow frozen CONFIRMATORY plans; released-data and protocol diagnostics are EXPLORATORY-AUDIT. The release inventory is in Supp.~R; FastUMI is in Supp.~I.

\subsection{External re-evaluation of FIPER}
\label{sec:fiper-audit}
\paragraph{Setting.} We ran the official FIPER code (learnsyslab/fiper at 13d79c5; five tasks; seeds 0, 1, 2, 42, 43) on the authors' released rollouts, converted from pickle to arrays in an isolated offline container. Selection follows the paper (S-paper): metrics are averaged over quantiles 0.90--0.99 and seeds, and the window and threshold type with the highest across-task mean TWA are chosen. As verified in the code, this selection uses test-rollout metrics. Time-weighted accuracy (TWA) credits a detected failure with $1-\mathrm{DT}$, where normalized detection time (DT) is the first alarm step divided by the task's maximum episode length minus one.

\paragraph{Reproduction.} Here ``matches'' means a gap no greater than 0.05, not a statistical equivalence test. FIPER reproduces on all five tasks within $\pm0.05$ of the published accuracy and TWA, with the published setting (window 25/50, time-varying quantile threshold). FIPER's components ACE and RND-OE and the flow-based observation-likelihood baseline logpZO reproduce on 5/5 tasks, the action-based RND baseline (RND-A) on 1/5.

\paragraph{Clock tie rule.} The clock's score equals its own time-varying threshold step by step, so its normalized score is exactly 1 in exact arithmetic; floating-point residues ($<10^{-12}$) produced spurious alarms in the raw output. A tie rule (normalized score $>1+10^{-9}$), fixed after results were seen, removes them; it changes no cell for any learned method. Both versions are kept; Table~\ref{tab:fiper-audit} uses the tie-corrected clock.

\begin{table}[t]
\centering\small
\setlength{\tabcolsep}{2pt}
\begin{tabular}{lrrrrrrr}
\toprule
 & \multicolumn{2}{c}{FIPER} & \multicolumn{2}{c}{Clock} & \multicolumn{2}{c}{Clock $-$ FIPER} & Quarter\\
Task & bAcc & TWA & bAcc & TWA & bAcc & TWA & bAcc\\
\midrule
Sorting & .661 & .542 & .726 & .511 & $+.064$ & $-.031$ & .547\\
Stacking$^\dagger$ & .732 & .617 & .500 & .500 & $-.232$ & $-.117$ & .601\\
PushT & .712 & .554 & .971 & .525 & $+.259$ & $-.030$ & .533\\
Pretzel & .850 & .676 & .850 & .498 & $+.000$ & $-.178$ & .669\\
PushChair & .969 & .836 & 1.000 & .833 & $+.031$ & $-.002$ & n/a\\
\bottomrule
\end{tabular}
\caption{FIPER external re-evaluation (exploratory audit), S-paper selection. Paired 95\% intervals from 2000 outcome-stratified rollout bootstraps, selection fixed, are preserved in Supp.~V and Fig.~\ref{fig:fiper-audit}. Quarter: FIPER bAcc among rollouts still running at $\lfloor0.25T\rfloor$; n/a denotes fewer than five outcomes per class. The within-task clock is 0.5. $^\dagger$Stacking TWA uses normalization 120 although its longest test rollout has 100 steps.}
\label{tab:fiper-audit}
\end{table}

\begin{figure*}[t]
\centering
\includegraphics[width=\textwidth]{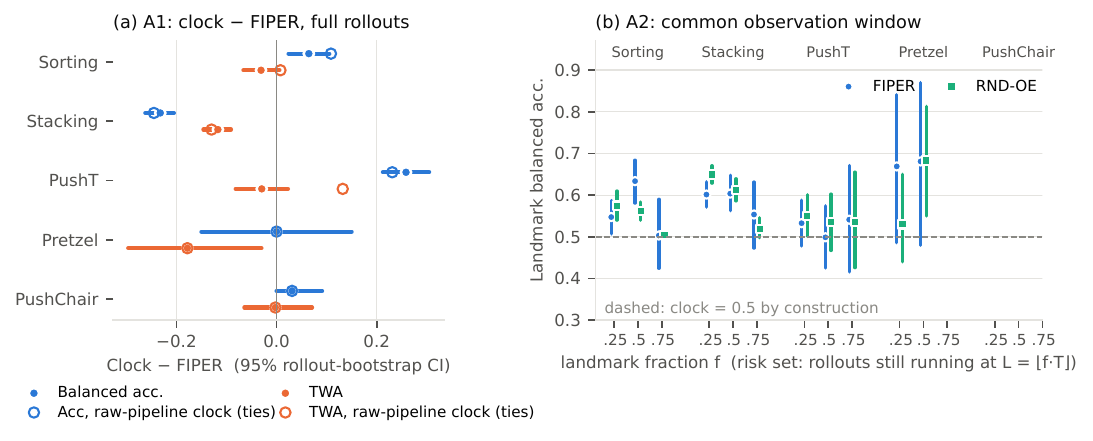}
\caption{\textbf{FIPER audit.} (a) Clock minus FIPER, full rollouts, 95\% rollout-bootstrap CI; hollow markers: raw-pipeline clock with floating-point ties. (b) Balanced accuracy among rollouts still running at landmarks $L=\lfloor fT\rfloor$; FIPER circles vs RND-OE squares; dashed: clock $=0.5$ by construction. NA denotes an unevaluable risk set.}
\label{fig:fiper-audit}
\end{figure*}

\paragraph{Findings.} Fig.~\ref{fig:fiper-audit} shows the contrasts and the landmark analysis. (i) \emph{Balanced accuracy is sensitive to duration.} The clock's accuracy is within 0.05 of, or above, FIPER's on four of five tasks, and clearly above on Sorting ($+0.064$) and PushT ($+0.259$; clock 0.971). On PushT every failed test rollout runs to the 38-step cap. The clock's advantage there primarily reflects specificity (TNR .941 versus .441), with similar sensitivity (TPR 1.000 versus .983), not a claim that FIPER alarms late (Supp.~M). (ii) \emph{TWA guards against it on these data.} The clock's TWA is never significantly above FIPER's and is clearly below on Stacking and Pretzel: the clock alarms late, and TWA charges for lateness. We stated this outcome in advance as a possible positive audit finding. (iii) \emph{Signal among running rollouts.} At $0.25T$, FIPER exceeds chance on Sorting and Stacking but at no landmark on PushT, although its full-rollout accuracy there is 0.712. On Sorting and Stacking the RND-OE component alone is at least as high (0.575, 0.650). (iv) \emph{Selection.} Choosing window and threshold type on one fold and evaluating on the other lowers FIPER's TWA by 0.040 on average (Pretzel $-0.153$; accuracy $-0.003$).

\paragraph{Clock ceiling and rankings (post hoc, exploratory).} Supp.~V evaluates Prop.~\ref{prop:clock} on the test lengths. Across all 2,550 clock settings, every alarm fired at one common step on exactly the rollouts longer than it, and no setting exceeded the ceiling. Duration alone could reach balanced accuracy 0.76--1.00, above FIPER's on every task; on Stacking the calibrated clock stays at 0.50 although lengths separate almost perfectly ($D^+=0.98$), because its thresholds come from successful calibration rollouts. TWA's ceiling is lower (0.52--0.83) but exceeds FIPER's TWA on Stacking (0.646 vs.\ 0.617) and PushT (0.570 vs.\ 0.554): TWA protected against the calibrated clock, not against every duration-only rule. Rankings also move. By full-rollout accuracy the clock ranks first on Sorting, PushT and PushChair; on Sorting and PushT, the ordering of the six scores at $0.25T$ runs largely opposite to their accuracy ordering (Kendall rank correlation $-0.87$ and $-0.60$). Many landmark differences lie within their intervals, so we read this as a change in what the metrics reward, not as a new ranking of methods.

\paragraph{Notes.} On Stacking, DT is normalized by a maximum episode length of 120 while the longest test rollout has 100 steps, which softens TWA's lateness penalty. Our run used one process per seed and dropped the unreported OR combination to avoid memory exhaustion; FIPER's computation is unchanged.

\subsection{A second external audit: Sentinel, including real-robot rollouts}
\label{sec:sentinel}
Sentinel~\cite{agia2024stac} released per-step scores of its temporal-consistency monitor (STAC), a diffusion-ensemble baseline, embedding baselines and VLM monitors, with calibration rollouts, for a real-robot PushChair task and simulated PushT and Cover tasks. We converted the released files in isolation, reimplemented Sentinel's detection rule from its released code, and reproduced every stored episode count. Its main setting compares a running sum of the per-step score with the 95\% quantile of the calibration rollouts' final sums and alarms at any step. We pass the clock through the same rule: its running sum is the elapsed step, so its threshold is the 95\% quantile of calibration success lengths. The plan was hashed before any length or score was read; results are exploratory.

\begin{table}[t]
\centering\footnotesize\setlength{\tabcolsep}{2.4pt}
\begin{tabular}{lccccc}
\toprule
Set & S/F & Clock & STAC & Best other & STAC@$\frac{T}{4}$\\
\midrule
PushChair & 10/10 & .90 & .75--.85 & .95 (GPT-4o) & ---\\
PushT (10) & 7--21/15--36 & .89--1.0 & .57--1.0 & .93--1.0$^\dagger$ & .49--.63\\
\bottomrule
\end{tabular}
\caption{Sentinel's released rollouts (exploratory): real PushChair and ten simulated PushT settings. All failures reach the step cap except on PushChair, where every failure is longer than every success. The clock and quantitative monitors use the cumulative rule; VLMs retain released verdicts. Entries are balanced accuracy. STAC: range over released variants. $^\dagger$Embedding Mahalanobis, available in 6 sets. Last column: STAC at the $0.25T$ landmark, where the clock is $0.5$. PushChair uses raw endpoint 84, not dense-score endpoint 36; its quarter risk set has two successes and ten failures, below the reporting gate (Supp.~H). Two Cover sets have one class with a single scored rollout and are omitted; the third Cover set and three PushT sets have no released detector scores (all 17 sets are in Supp.~H).}
\label{tab:sentinel}
\end{table}

\paragraph{Findings.} In every interpretable set failed rollouts outlast successes: all reach the step cap except on PushChair, where every failure is longer than every success, and almost no success reaches it, so $D^+\ge0.94$ (Supp.~H). Under Sentinel's own rule the clock reaches balanced accuracy 0.89--1.00, and no STAC or VLM variant has a paired 95\% interval above it in any set; on the real robot GPT-4o is highest (0.95 vs.\ 0.90; difference $+0.05$ [$-0.10$, $0.20$]). At the $0.25T$ landmark STAC falls to 0.49--0.63. The running sum in the main setting grows with elapsed time by construction, which is one route by which duration enters. As with FIPER, this concerns what the released evaluation rewards, not whether these monitors are useful.

\paragraph{PushChair timing (exploratory).}
Using raw span 84, GPT-4o/clock TWA is .745/.757. Restricting clock decisions to VLM query times changes the clock to .643. Both paired difference intervals include zero (Supp.~H): the point reversal depends on cadence.

\paragraph{Truncation versus landmarks.} SAFE's shortest-rollout truncation~\cite{gu2025safe} evaluates all rollouts on the common prefix $[0,T_{\min})$; later landmarks condition on rollouts still running. On FIPER, $T_{\min}$ is 16--24\% of the horizon on four tasks; on Sorting, FIPER's balanced accuracy is 0.538 there and 0.634 at $0.5T$. On real-robot PushChair only truncation is usable, as later risk sets hold too few successes. Across 25 released rollout sets (5 FIPER, 17 Sentinel, 3 SAFE), the length-only ceiling is at least 0.76 in 21 (FIPER's and Sentinel's PushChair sets may share rollouts); the four exceptions, like our fixed-horizon \mbox{SIMPLER} runs, mostly continue after success (Supp.~H).

\subsection{Protocol-sensitive ranks and task composition}
\label{sec:ranking}
\paragraph{Available evidence and specification.}
The extension was specified after the earlier audits were read and remains exploratory. Of the 25 released sets, four Sentinel sets have no detector scores. FIPER's existing numerical export supplies first-alarm indices for six scores, not continuous confidence values: it supports balanced-accuracy ranks only. Sentinel supplies cumulative-score series and, for some VLMs, binary verdicts. SAFE supplies four training-free token metrics; learned SAFE detectors were not rerun. The supplement lists every release, calibration availability and excluded risk set. We preserve native orientations and use maximum available prefix score; missing decision positions are not interpolated.

\paragraph{Rank changes at a fixed metric.}
On Sorting, full-rollout versus quarter-horizon balanced-accuracy ordering across six FIPER scores has $\tau_b=-.867$ $[-1.000,-.067]$: 14 of 15 method pairs reverse. PushT has $\tau_b=-.600$ $[-.867,.200]$, a descriptive reversal with uncertainty; Stacking is positive ($+.333$). Sorting's quarter-horizon risk set contains 255 successes and 142 failures; PushT's contains 98 and 198. On all ten evaluable Sentinel PushT sets, the corresponding balanced-accuracy rank correlation is negative; five intervals lie below zero. Continuous-score AUROC behaves differently: eight of ten point correlations are positive, and all ten intervals include zero. Thresholded-alarm ranking and continuous-score ranking cannot be interchanged. All these intervals are exploratory, conditional on stored scores and settings, without multiplicity correction. A same-cohort check retains Sorting's six-score reversal on the identical 397 quarter-horizon rollouts ($\tau_b=-.87$ $[-1.00,-.07]$). Removing the clock gives $-.80$ $[-1.00,.20]$; restricting all horizons to the 105 late survivors gives $-.14$ $[-.55,.55]$ with the clock. The result is therefore not a universal, cohort-independent ordering (Supp.~N).

\paragraph{Task composition after truncation.}
In SAFE's LIBERO release (ten tasks, 500 episodes), task-wise shortest-rollout truncation gives a pooled clock AUROC of .587 despite constant clock scores within each task. A five-fold diagnostic task-prior reference reaches .843 $[.835,.851]$; adding elapsed steps gives .843. These references use other-fold outcome labels and do not establish unseen-task generalization or deployment calibration. The released \texttt{max\_token\_prob} field scores .634 pooled. Restricting to the same seven tasks with at least five outcomes of each class, pooled and macro AUROC are .586 and .583; the other tasks are not silently included in the macro denominator. Thus much of this pooled--macro difference is a population-composition issue, not a detector improvement. Neither these AUCs nor their differences are additive or causal contributions.

\paragraph{Constructive output.}
Sorting illustrates the audit chain: full-rollout bAcc favors the clock (.726 vs.\ .661); at the quarter landmark the clock is .500 and FIPER .547, while six-score ranks reverse. This makes selection by full-rollout bAcc alone questionable; it does not establish a new winner (Supp.~N). Figure~\ref{fig:findings} summarizes the diagnostics. Report risk-set counts rather than select favorable landmarks. On small real-robot PushChair sets, later risk sets are too small; unscorable points remain absent. Shared original-episode bootstrap draws are restricted separately to each risk set. These original intervals hold calibration, selection and training fixed. Supplementary calibration-plus-test (CAL+TEST) resampling propagates threshold uncertainty for reconstructable Sentinel variants, but not training or setting selection (Supp.~N).

\subsection{Termination rules on \mbox{SIMPLER} and a prespecified confirmation}
\label{sec:simpler}
\mbox{SIMPLER}~\cite{li2024simpler} runs one frozen Octo policy~\cite{octo2024} on four WidowX tasks, with 640 development and 640 confirmation episodes at a fixed horizon.

\paragraph{Termination-rule control.} On 480 non-stacking confirmation episodes, the clock-only any-step detector has true-positive/false-positive rates (TPR/FPR) $0/0$ at the fixed horizon and $1.00/0.111$ $[0.061, 0.171]$ when the same episodes stop at success ($q=0.90$). The control is constructed and post hoc; it isolates the mechanism of Eq.~\ref{eq:clock-exposure}. A prespecified replication with a frozen GR00T N1.7 policy~\cite{bjorck2025gr00t} (816 new episodes) confirms it: balanced accuracy $0.500\to0.965$ $[0.946, 0.981]$ at $q=0.95$ (Supp.~O); an SO-101 replication is in Supp.~B.

\paragraph{Confirmation.} The hash-frozen primary contrast, Dens minus A, reaches $\Delta$AUC $+0.0099$ (configuration-bootstrap 95\% interval $[0.0009, 0.0183]$): above zero, but below the prespecified 0.02 practical-effect target. The registered secondary image OOD contrast is $+0.0257$ $[0.0063, 0.0426]$, without multiplicity control. An ensemble advantage is not established: Dens exceeds a matched-data ridge model by only $+0.0018$ $[-0.0024, 0.0060]$, and the residual signal is not above elapsed time within configuration and time bin (0.609 versus 0.641; post hoc). The target is eventual task failure, not forecast error. Forward models use successful development (DEV) rollouts; outcome classifiers use configuration-group cross-validation within confirmation. This endpoint confirms the prespecified cross-validated pipeline, not a DEV-trained detector transferred without refitting.

\paragraph{Remaining visual information (exploratory).}
Historical GR00T image-distance scores retain task-stratified landmark AUROC .731/.759/.682 (Fig.~\ref{fig:findings}c). Populations change, weighting is by outcome pairs, and aggregation was post hoc; this is not confirmation of a detector (Supp.~O).

\paragraph{Prospective frozen-monitor test.}DEV-trained task/time, state/action and image-novelty heads transfer without refitting. TEST-1 freezes weights, references and thresholds before 416 new GR00T episodes. Its quarter-horizon gate fails: Eggplant retains four successes, below five per class. Descriptive macro AUROC is .500/.715/.726; visual minus state/action is $+.011$ [$-.005,.028$] (1967/2000 valid configuration-bootstrap draws). No tasks or episodes are added (Supp.~S). Separately, TEST-2 uses 1,248 fixed-60-step episodes on Carrot/Spoon and 24 fixed configurations per task. Its paired macro AUROC increment is $+.0035$ (95\% within-configuration episode-bootstrap interval [$+.0007,+.0062$]), below the prespecified $.02$ target; Carrot decreases and Spoon improves. The supplement's TEST-2 section reports the post-launch repository push and limited preflight. The success bank supplies DEV supervision; actions may carry visual information. Thus this frozen visual pipeline adds a small amount of predictive information on the specified tasks and configurations, but misses the prespecified practical target. Useful failure-onset warning and causal visual reliance remain unestablished.

%% file: supplementary-analyses.tex
\onecolumn
\appendix
\section*{Appendix: Supplementary Analyses}
\addcontentsline{toc}{section}{Appendix: Supplementary Analyses}
\setcounter{figure}{0}
\setcounter{table}{0}
\setcounter{equation}{0}
\renewcommand{\thefigure}{S\arabic{figure}}
\renewcommand{\thetable}{S\arabic{table}}
\renewcommand{\theequation}{S\arabic{equation}}
\renewcommand{\theHfigure}{app.\arabic{figure}}
\renewcommand{\theHtable}{app.\arabic{table}}
\renewcommand{\theHequation}{app.\arabic{equation}}
\paragraph{Reading guide.}
For the mathematical basis and duration-only ceilings, read Sections~\ref{app:ceiling} and~\ref{app:design-principles}.
For released-data audits and their evidence units, read Sections~\ref{app:ranking}, \ref{app:robustness-extension} and~\ref{app:evidence-ledger}; Section~\ref{app:robustness-extension} includes a worked Sorting example.
For the termination-mechanism replication, read Section~\ref{app:gr00t}.
For the separate frozen-monitor tests, read Sections~\ref{app:prospective-monitors} (TEST-1) and~\ref{sec:prospective-test2} (TEST-2).

\paragraph{Terminology.}
Colour images use red--green--blue (RGB) channels. Data partitions are model fitting (FIT), training (TRAIN), validation (VAL), development (DEV), calibration (CAL), test (TEST) and evaluation (EVAL); their roles are study-specific. Metrics include balanced accuracy (bAcc), time-weighted accuracy (TWA), area under the receiver operating characteristic curve (AUROC, also AUC), area under the risk--coverage curve (AURC), mean squared error (MSE), and mean absolute error (MAE). CI denotes confidence interval; TPR, TNR and FPR denote true-positive, true-negative and false-positive rates. Model terms include multilayer perceptron (MLP), out-of-distribution (OOD), Kullback--Leibler (KL) divergence, cumulative sum (CUSUM), and adaptive conformal inference (ACI). Neural measurements use electroencephalography (EEG), event-related potential (ERP) and mismatch negativity (MMN); FCz is an electrode label. Learning algorithms include proximal policy optimization (PPO) and success-guided sampling (SGS). Runtime terms include vision-language model (VLM), vision-language-action (VLA), random network distillation (RND), its observation-embedding variant (RND-OE), action-chunk entropy (ACE), and statistical temporal action consistency (STAC). External method and model names retain their published spelling; no undocumented acronym expansion is inferred.

\input{sec_appendix_integrated_20261105}
\input{robustness_extension_20261110}
\input{sae_representation_extension_20261111}
\input{evidence_inventory_round2}
\input{prospective_monitor_20261112}
\input{constructive_design_principles}
\input{systems_representation_pilot_20261113}
\input{detailed_audit_estimates_20261113}
\input{physical_object_memory_20261113}
\input{evidence_ladder_diagnostics_20261114}
\input{scope_monitor_signals_20261114}
\input{prospective_test2_20261116}

%% file: sec_appendix_integrated_20261105.tex
\section{Egocentric RGB forecast and selection studies (exploratory)}
\label{app:egocentric}
These studies test whether a latent residual or a working-belief surprise adds reliability information beyond quality and arrived historical error when forecasting a wrist reference 0.5\,s ahead from egocentric RGB. They are small and mostly negative; they motivated the strong-predictor control of Sec.~5.4 of the main paper.

\begin{figure}[!ht]
\centering
\includegraphics[width=\textwidth]{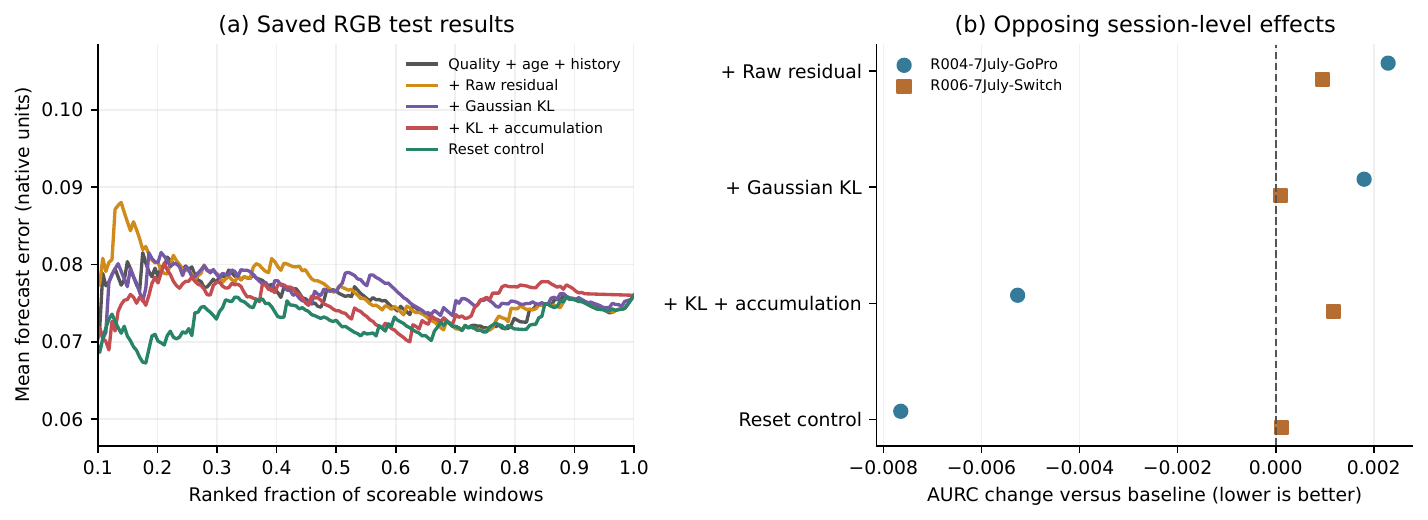}
\caption{\textbf{HoloAssist: pooled and per-recording views.} Descriptive risk--ranked-acceptance curves and AURC differences from quality/age/history. Opposing effects across two TEST recordings prevent a stable-benefit claim; realized CAL-fixed acceptance differs. Saved-output reconstruction, not independent replication.}
\label{fig:rgb-audit}
\end{figure}

\paragraph{HoloAssist.}
\label{sec:completed-bounded}
Four FIT, two DEV, two CAL and two TEST recording sessions, first 60 seconds each, give 337, 124, 93 and 194 scoreable windows; TEST contains 232 issued queries. A frozen V-JEPA~2 checkpoint~\cite{assran2025vjepa2} (\texttt{facebook/vjepa2-vitl-fpc64-256}, revision \texttt{b3c1679}) is adapted to 16 past and 8 predicted frames, and a shared head predicts the left wrist at $+0.5$\,s in launch-camera coordinates. Reference poses never enter the RGB predictor; error feedback is released at target time plus an assumed 0.1\,s. All reliability arms share the same point forecasts (mean endpoint error 0.076004, native units). Neither raw latent residual nor Gaussian surprise consistently improves the quality/history baseline (Table~\ref{tab:bounded-rgb}). Accumulation lowers pooled AURC but reduces coverage and acceptance, with opposite effects in the two TEST sessions (Fig.~\ref{fig:rgb-audit}); a reset control has the lowest pooled AURC. The cached-RGB implementation had 102.74\,ms median latency and missed the 100\,ms deadline on every query. A later quantile-boosting reanalysis lowered pooled AURC (.06938 to .06574) but raised error at 80\% ranked acceptance (.07066 to .07174); the TEST sessions had already been inspected.

\begin{table}[!ht]
\centering\small
\begin{tabular}{lrrrr}
\hline
Arm & Cov. & Accept. & Sel. error & AURC\\
\hline
Constant scale & .9021 & .1164 & .11034 & .07600\\
Quality, age, history & .8247 & .6207 & .07865 & .07594\\
+ Raw residual & .8299 & .6552 & .07814 & .07741\\
+ Gaussian KL & .8299 & .5948 & .08129 & .07642\\
+ KL, accumulation & .7990 & .5388 & .07510 & .07440\\
Reset control & .8299 & .6638 & .07716 & .07303\\
\hline
\end{tabular}
\caption{Bounded HoloAssist RGB replay. Acceptance uses all 232 issued queries; coverage, error and AURC use 194 scoreable windows. Native units.}
\label{tab:bounded-rgb}
\end{table}

\paragraph{HOT3D.}
\label{sec:hot3d-dev}
From a pinned HOT3D release~\cite{banerjee2025hot3d} (\texttt{30fe9674}) we select 16 five-second TRAIN clips (13 participants; 9 Aria, 7 Quest~3). Frozen ResNet-18~\cite{he2016resnet} and DINOv2-small~\cite{oquab2023dinov2} features at frames $t-30,t-15,t$ feed a ridge head predicting the left reference wrist at $t+15$; outer folds hold out one participant (560 scoreable queries). DINOv2 lowers point error by 4.3\% relative to ResNet-18 (paired participant bootstrap $-0.01036$ $[-0.03115, 0.00743]$). A post-hoc FIT-only constant centroid has error .22553, below both image heads, so this setup has not shown useful image-conditioned forecasting, and its reliability results cannot be read as reliability evidence (Table~\ref{tab:hot3d-dev}). A MediaPipe hand-geometry adaptation gives .21816, not stably better than the centroid ($-.00737$ $[-.02650, .00884]$). With past UmeTrack wrist inputs, ignoring an injected 0.25\,s delay raises error from .08557/.09755 to .11889/.14307 (left/right), and extrapolating by the known age worsens it further (.14653/.17900): knowing age does not fix an inaccurate motion rule.

\begin{table}[!ht]
\centering\small
\begin{tabular}{lrrrr}
\hline
Head / reliability cue & Point & Risk@80 & Accept. & Cov.\\
\hline
ResNet / quality + history & .23920 & .24165 & .5681 & .8505\\
ResNet / + latent residual & same & .24258 & .6495 & .8505\\
DINO / quality + history & .22884 & .23025 & .5956 & .9253\\
DINO / + latent residual & same & .23043 & .5868 & .9264\\
FIT centroid (post-hoc) & .22553 & -- & -- & --\\
\hline
\end{tabular}
\caption{HOT3D TRAIN-pool development, participant-equal means in native units.}
\label{tab:hot3d-dev}
\end{table}

\section{Other robot development checks (exploratory)}
\label{app:robotdev}
AgiBot state forecasting (120 episodes, three tasks; 60 FIT, 18 DEV, 18 CAL, 24 TEST) improves mean error over constant velocity by 16.2\%, 3.1\% and 6.9\%; state innovation and accumulation show no stable extra benefit. RAX (6,563 Franka command records, five sessions; 2,771 commands with 44 anomalies after availability rules) gives leave-one-session-out AUC .50204 without and .51628 with early innovation/path deviation; the row-bootstrap interval $[-.01730, .04553]$ ignores session clustering. ArmnetBench, a set of SO-101 single-arm and bimanual LeRobot policy rollouts that stop at success, shows the duration effect post hoc: the clock exceeds every sensor-based detector at equal calibration (TPR $+0.24$ to $+0.41$, intervals excluding zero), whereas under time-stratified evaluation the ranking reverses (elapsed time 0.581 versus the ensemble signal 0.650, difference $+0.070$ $[0.008, 0.122]$). An ensemble-signal development analysis there reports Dens minus a kinematic/history baseline $+.023$ $[.003, .046]$ after repeated inspection; the matched ridge contrast includes zero and CUSUM lowers AUC by .005.

\section{Decision records, age gates and the original reliability model}
\label{app:gates}
For required modality $m$, an evidence-age upper bound is
\begin{equation}
A^{\max}_{t,m}=(t+\epsilon_t)-(\widetilde\tau^{\rm obs}_{t,m}-\epsilon_m),
\label{eq:age-upper}
\end{equation}
with mapped observation time $\widetilde\tau$ and justified error bounds $\epsilon$; each stream keeps its own age. A decision-record schema with a receipt watchdog (G0), an evidence-age/quality gate (G1) and candidate-score gates (G2) has been tested only in replay: on \mbox{SIMPLER} episodes with injected delays, G1 rejects 100\% of 100--200\,ms-delayed decisions that G0 passes, and one bounded re-observation recovers none of the persistent faults. The original candidate scored a frozen predictor's latent residual $e_t$ with a Gaussian working belief and its posterior/prior KL~\cite{itti2009bayesian}, accumulated into a bounded index that scaled a delayed-feedback prediction set; Table~\ref{tab:bounded-rgb} reports its exploratory performance. Withholding a forecast is not a physical safety certificate.
\section{RGB diagnostics and reference checks}
\label{app:rgb}
These exploratory diagnostics complement Tables~\ref{tab:bounded-rgb} and~\ref{tab:hot3d-dev}. Neither pooled gains nor a backbone comparison establishes a stable incremental reliability benefit.
\begin{figure}[!ht]
\centering
\includegraphics[width=\textwidth]{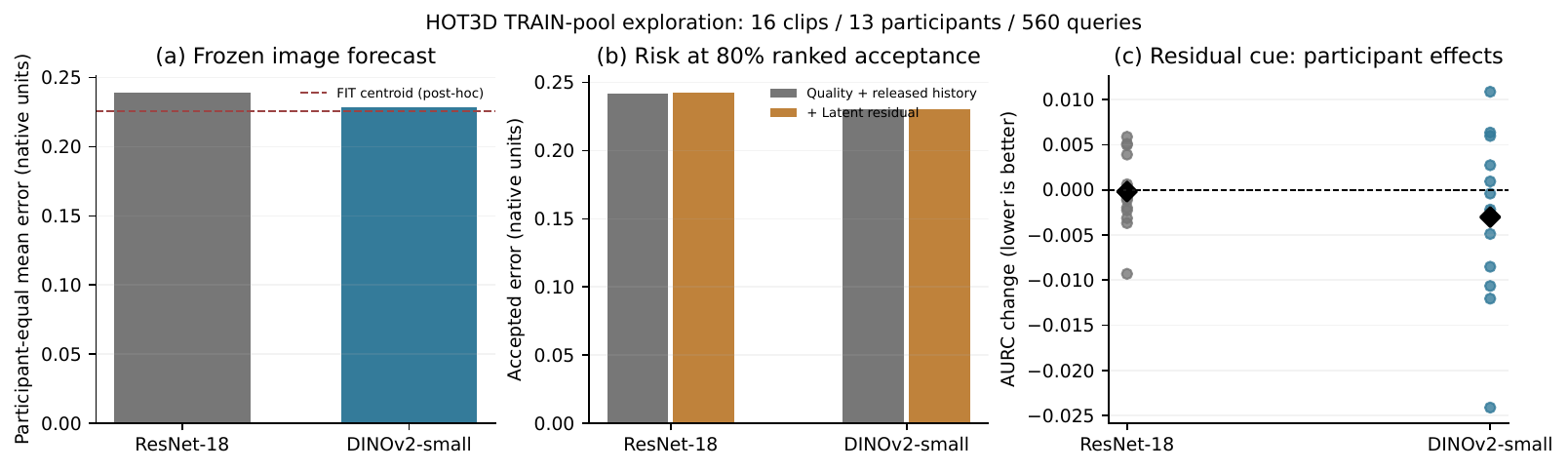}
\caption{\textbf{HOT3D TRAIN-pool diagnostics.} Image heads do not beat the post-hoc FIT centroid; latent residual does not improve error at 80\% ranked acceptance. Participant AURC changes vary; black diamonds show their mean. Native units, adapted exploratory models, not official challenge results. Derived hand-reference statistics retain HOT3D noncommercial share-alike terms; source: Banerjee et al.~\cite{banerjee2025hot3d}.}
\label{fig:hot3d-visual}
\end{figure}
\clearpage
\section{FastUMI selection analysis, training and local prediction errors}
\label{app:training}
\begin{figure}[!ht]
\centering
\includegraphics[width=\textwidth]{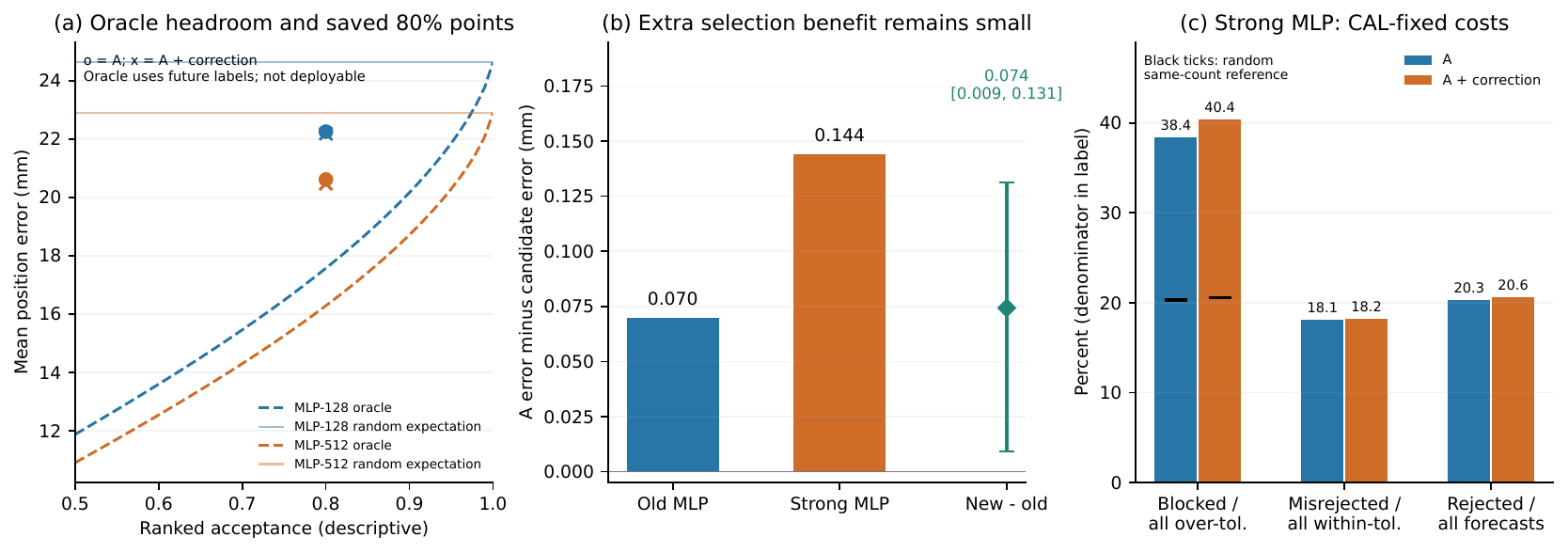}
\caption{\textbf{Small incremental selection gains have a rejection cost (exploratory FastUMI DEV).} (a) Offline oracle headroom and 80\%-ranked operating points on 73,775 common windows. (b) Correction-cue improvement and paired difference, 50-episode block-bootstrap intervals. (c) Each rule's own CAL-fixed threshold: blocking errors also rejects valid outputs. The oracle uses future labels and is not deployable.}
\label{fig:fastumi_selection}
\end{figure}
\begin{figure}[!ht]
\centering
\includegraphics[width=.94\textwidth]{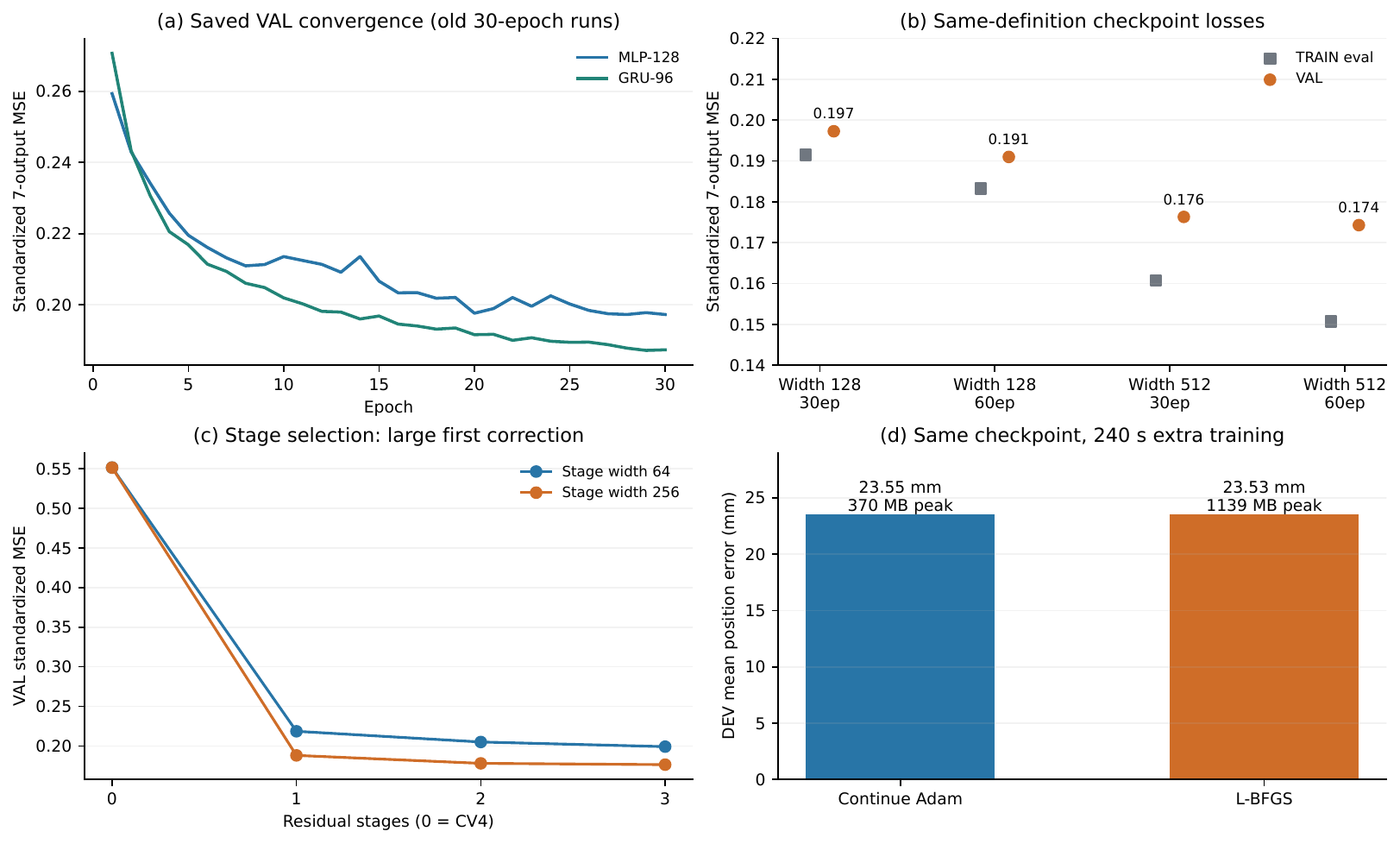}
\caption{\textbf{Training diagnostics (exploratory DEV).} Saved VAL convergence; same-definition TRAIN/VAL checkpoint losses; stage selection with unmatched budgets; and equal-time optimizer branches from one checkpoint. Standardized seven-output MSE differs from position error. One seed; no optimizer equivalence or original MSNN reproduction.}
\label{fig:fastumi_training}
\vspace{5pt}
\includegraphics[width=.94\textwidth]{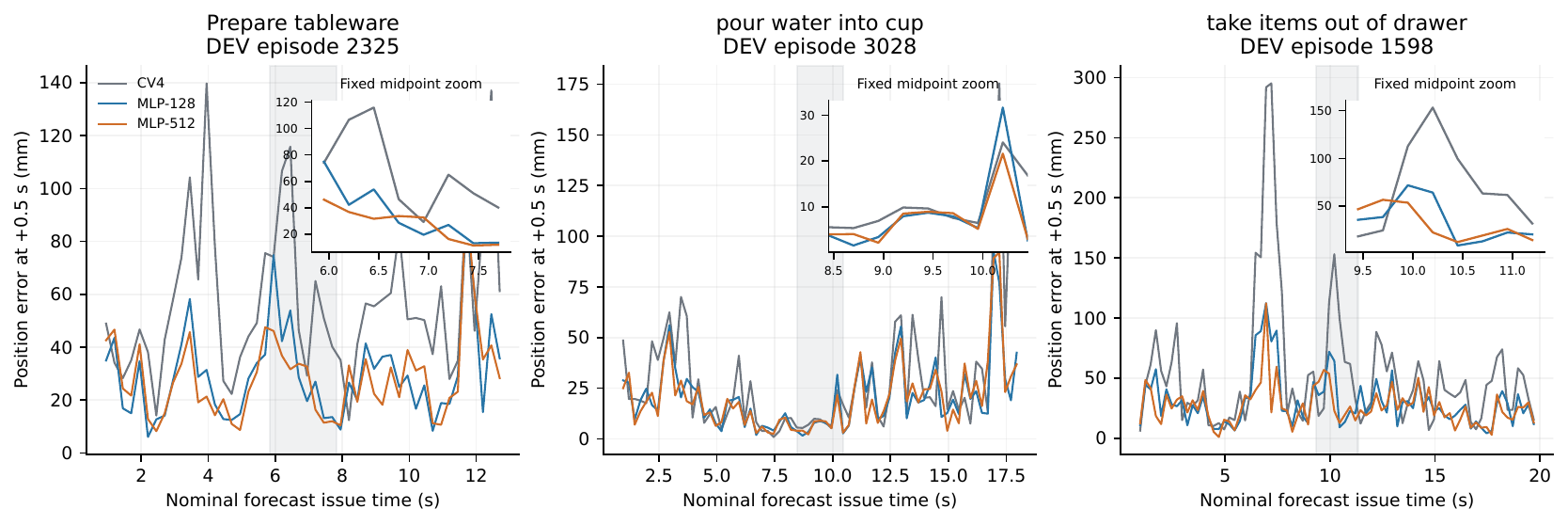}
\caption{\textbf{Whole-record and local behavior on the same diagnostic page.} One DEV episode per task selected by minimum fixed-key SHA-256 among records with at least 40 windows; two-second insets are centered on each record midpoint. Position-error norms concern a $+0.5$-s forecast. Nominal frame-index/20 timing does not measure sensor latency. These cases illustrate variation, not aggregate efficacy.}
\label{fig:fastumi_local}
\end{figure}
\clearpage
\section{Predictor architecture and prospective deployment}
\label{app:deployment}
\begin{figure}[!ht]
\centering
\includegraphics[width=\textwidth]{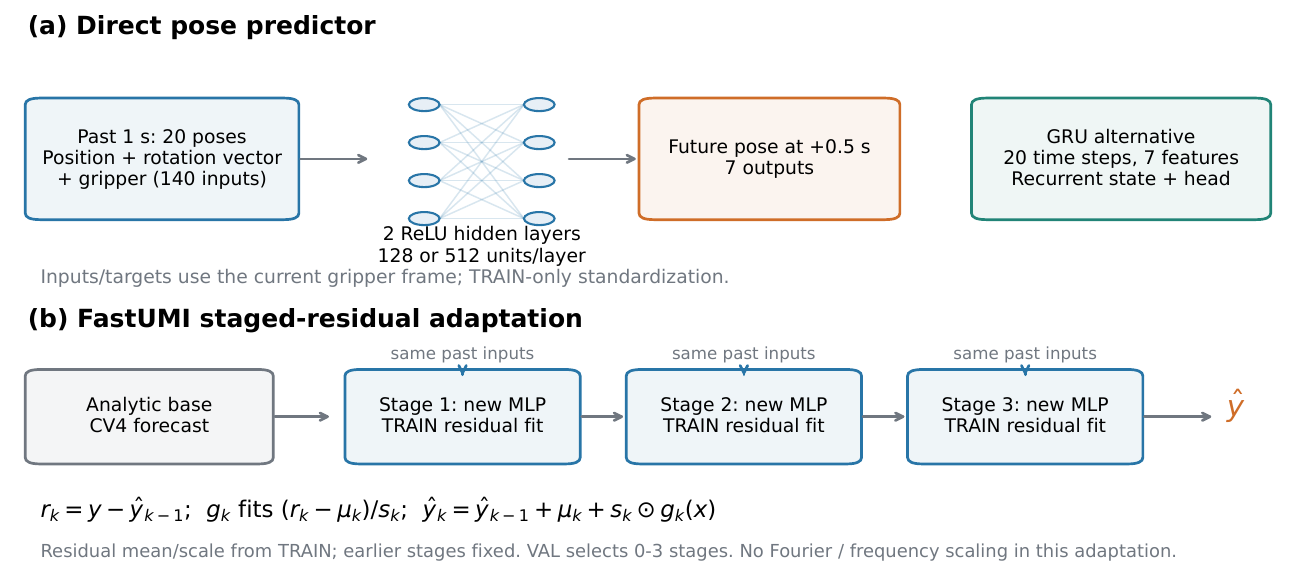}
\caption{\textbf{Predictor internals and staged-residual control.} MLP/GRU share the pose history. Each residual network receives the same past input; arrows denote forecast corrections, not hidden-state transfer. TRAIN defines residual mean/scale and VAL selects 0--3 stages. Nodes are illustrative. This seven-output adaptation implements no Fourier scaling or machine-precision reproduction~\cite{wang2024multistage}.}
\label{fig:fastumi_model}
\end{figure}
\paragraph{Prospective frozen-policy validation loop.}
Table~\ref{tab:real_sim_cycle} defines the prospective loop. The photograph-estimated tabletop is not a calibrated twin. ACT would use real WidowX AI demonstrations after adapter acceptance; simulated pretraining is a separate variant. For that variant, policies trained in massively parallel simulation, e.g., with success-guided sampling of task configurations~\cite{zhang2026balanced}, would be a further source of rollouts for the same audit; none has been trained here. Freeze design before real CAL and lock thresholds before TEST. Gates remain advisory; causal intervention benefits require a new paired experiment. Repeated values alone do not establish tracking loss, and persistent-injection recovery does not establish general re-observation utility.

\begin{table}[!ht]
\centering\small
\begin{tabular}{p{.12\textwidth}p{.40\textwidth}p{.39\textwidth}}
\hline
Stage & Planned work & Boundary / required evidence\\
\hline
Real 0 & Base/table/object measurements, camera calibration and 3--5 short teleop recordings & Engineering inputs only; record units, frames, uncertainty and actual timing. No policy or efficacy claim.\\
Sim 1 & Measured-scene headless rehearsal; fixed step/event and termination rules & Validate camera projection and trajectory replay; estimated geometry remains flagged until replaced.\\
Real 1 & Demonstrations for ACT; operator-wrist and robot-view shadow development & Separate targets, episode/session splits and TRAIN/VAL selection. These are development records.\\
Sim 2 & Stress tests informed by real development observations & Simulator state labels only modeled events; freeze policy, detector design and calibration procedure.\\
Real 2 & Frozen ACT, real CAL thresholds, then new TEST sessions & Gates remain advisory; report risk, rejection and latency. Intervention/task benefit is a separate experiment.\\
\hline
\end{tabular}
\caption{Prospective real--sim--real--sim--real validation loop. No row denotes a completed physical deployment. Measuring geometry does not establish contact-dynamics or sensor-latency fidelity.}
\label{tab:real_sim_cycle}
\end{table}

\paragraph{Prospective spatial evidence-age check.}
The attribution question of the main paper also applies within a frame: a global stream check can pass while the evidence behind one target is old.
The check here measures the age of the evidence behind each decision. A prospective check reports, for each decision, (i) the age of the direct observation supporting each queried target, (ii) the time skew among the inputs actually selected, and (iii) whether required perception outputs were ready at decision time (Fig.~\ref{fig:spatial-age}). A missing clock mapping or observation is reported as unknown, never as zero age; during a stream outage the last known observation time is retained, so age keeps growing.

The first recording protocol is drafted but not yet frozen, and no result is reported here. A fixed RGB-D camera with hardware timestamps observes a tagged target; a second camera tracking a tag on the same target, registered through a shared anchor, provides an independent position reference. Conditions are a static target (old but correct), a moving target, occlusion by another object, a fresh but mislocalized distractor, a stream interruption, and an unstaged tabletop task without instructed occlusion. Staged and unstaged sessions are reported separately, and the frequency of ``global check passes, local evidence stale'' is quoted from unstaged sessions only. The time budget is fixed before data from tolerance and target speed, $B=\delta/v_{\max}$ (10\,mm at 0.2\,m/s gives 50\,ms). As in the main audit, the per-target check earns credit only if it flags more over-tolerance decisions than a global receipt timeout and an elapsed-task clock at an equal false-flag rate set on calibration sessions. No robot is actuated; the check reports reasons and leaves any response to the existing controller.

\input{fig_spatial_age}

\clearpage
\section{Separate neural-data boundary study}
\label{app:eeg}
\emph{On the name.} \emph{Ego} denotes robot-centered evidence: what the robot itself observes and executes (camera frames, proprioception and commanded actions), as opposed to external annotations or privileged simulator state. \emph{Oddball} refers to the oddball paradigm of cognitive neuroscience, in which rare deviant stimuli presented among frequent standards evoke a measurable surprise response (e.g., the P3 and mismatch-negativity tasks of ERP CORE~\cite{kappenman2021erpcore}). Analogously, a failure warning should mark a departure from the robot's expected course rather than merely elapsed time. The oddball name is a motivating analogy to departures from an explicit expectation. Rarity, violated expectation, task error and future failure are distinct targets; the analogy does not identify our duration audit with a neural mismatch mechanism. The robot results do not depend on this analogy.

Change blindness supplies a complementary motivation, not evidence of a shared mechanism. Human observers can miss large scene changes, and failure to report a change need not imply that no prior representation was retained; it can also reflect a failure to attend to or compare the relevant pre- and post-change states~\cite{simons2005change,simons2002preserved}. By analogy, a robot monitor may retain visual information yet fail to compare stored state with current evidence, or may allocate limited processing to task-irrelevant change. EgoOddball therefore does not equate high prediction error with failure: a deviation must be interpreted relative to an explicit expectation, causally available evidence and task outcome, while occluded or stale evidence may warrant abstention. This framing motivates future physical-quadruplet studies crossing expectation violation with task correctness under controlled occlusion and staleness, followed by a separate Aria Gen~2 intervention study. These are prospective designs, not results of this paper.

An exploratory processed ERP CORE~\cite{kappenman2021erpcore} mismatch-negativity analysis tested whether contextual or accumulated surprisal improves single-trial voltage prediction beyond stimulus/history controls. Neither lowers the fixed participant-held-out mean squared error. Its derived-table regressions were reproduced, but end-to-end preprocessing was not independently repeated. The recordings are unpaired with robot/video data. This separate analysis supplies no evidence of a shared neural mechanism, failure-monitor accuracy, or robot-learning benefit; it is not an additional main-paper contribution.

\section{Clock ceiling across released rollout sets (exploratory)}
\label{app:ceiling}
Table~\ref{tab:ceiling-all} applies Prop.~1 of the main paper to every released rollout set we could obtain, using only rollout lengths and outcomes; no detector is run. Lengths are policy steps as recorded in each release (FIPER: scored steps; Sentinel: recorded timesteps of the raw episode files). FIPER's and Sentinel's PushChair sets have identical outcome counts and cap shares and may be the same real-robot rollouts; we list them separately, but they may not be independent. SAFE's $\pi_0$-FAST LIBERO-10 release (500 rollouts; length in policy inference steps) was statically scanned and converted under the same isolation, as were its two real-robot releases: OpenVLA on a WidowX (per-step CSV logs; its pickles, which hold torch objects, were scanned but not opened) and $\pi_0$-FAST on a Franka in the DROID setup. FAIL-Detect releases code but no rollouts.

\begin{table}[ht]
\centering\small
\begin{tabular}{llccccc}
\toprule
Release & Set & S/F & F at cap & S at cap & $D^+$ & Ceiling\\
\midrule
FIPER & Sorting & 255/145 & 0.52 & 0.00 & 0.517 & 0.76\\
FIPER & Stacking & 315/485 & 0.00 & 0.00 & 0.983 & 0.99\\
FIPER & PushT & 102/198 & 1.00 & 0.01 & 0.990 & 1.00\\
FIPER & Pretzel & 10/10 & 0.20 & 0.00 & 0.800 & 0.90\\
FIPER & PushChair & 10/10 & 0.10 & 0.00 & 1.000 & 1.00\\
Sentinel & PushChair (real) & 10/10 & 0.10 & 0.00 & 1.000 & 1.00\\
Sentinel & PushT, 13 sets & 7--21/15--36 & 1.00 & 0.00--0.06 & 0.94--1.00 & 0.97--1.00\\
Sentinel & Cover, ss & 5/45 & 1.00 & 0.00 & 1.000 & 1.00\\
Sentinel & Cover, na & 47/3 & 1.00 & 0.94 & 0.064 & 0.53\\
Sentinel & Cover, ca & 45/5 & 1.00 & 0.96 & 0.044 & 0.52\\
SAFE & $\pi_0$-FAST LIBERO-10 & 298/202 & 1.00 & 0.00 & 1.000 & 1.00\\
SAFE & OpenVLA WidowX (real) & 244/288 & 1.00 & 1.00 & 0.000 & 0.50\\
SAFE & $\pi_0$-FAST Franka (real) & 558/906 & 0.94 & 0.70 & 0.245 & 0.62\\
Ours & \mbox{SIMPLER}, 4 tasks & 640 total & \multicolumn{2}{c}{fixed horizon} & 0 & 0.50\\
\bottomrule
\end{tabular}
\caption{Length-only clock ceiling $\frac12(1+D^+)$ (exploratory). \emph{At cap}: share of each outcome whose length equals the set's maximum. Sentinel PushT spans three randomization conditions and three execution horizons. Franka: 13 tasks with horizons of 25--63 steps; \emph{at cap} is relative to each task's maximum.}
\label{tab:ceiling-all}
\end{table}

The ceiling is at least 0.76 in 21 of the 25 released sets. In the four exceptions successful rollouts also largely run to the cap (two Cover sets, the WidowX set with a fixed 50 steps, and the Franka set), so lengths depend little on the outcome; our \mbox{SIMPLER} runs are fixed-horizon by design. Stacking shows that the ceiling can be high even when no failure reaches the cap: failures are simply longer. A high ceiling does not mean a detector exploits duration; it bounds how much of a reported score duration alone could account for, and should be reported next to any-step alarm metrics.

\paragraph{SAFE under its own protocol.} SAFE's headline metric scores each rollout by its maximum over the first $T_{\min}$ steps of its task~\cite{gu2025safe}, which removes length within a task. In its $\pi_0$-FAST LIBERO-10 release every failure runs to the 104-step cap and no success does, so the untruncated clock has ROC-AUC 1.00. Under SAFE's truncated metric pooled over tasks, the clock scores 0.59 [0.53, 0.64]: tasks differ in $T_{\min}$ and in failure rate, and in one task all 50 rollouts fail and $T_{\min}$ equals the cap; without that task (post hoc) it is 0.45. For SAFE's hand-crafted token scores, which need no training, the untruncated variant is inflated: maximum token $-\log p$ has ROC-AUC 0.88 on full rollouts and 0.63 truncated (difference [0.20, 0.29]); maximum token entropy 0.81 and 0.54. At the $0.5T$ landmark they score 0.63 and 0.58, and the truncated-minus-landmark interval contains 0 for all eight token scores. We use all rollouts rather than SAFE's trained test splits, and learned detectors are not rerun. These results support SAFE's choice of truncation; the residual pooled clock signal is a reason to report per-task scores as well.

\paragraph{PushChair length semantics and scoring coverage.}
The twenty raw Sentinel PushChair episodes have recorded elapsed-step endpoints 12--24 for successes and 40--84 for failures. Their maximum endpoint is 84. Dense detector tables instead stop at 36; that is their scoring-prefix endpoint, not the end of every failed episode. The VLM table has two query positions per episode and reaches 44 overall. Array row counts, last score time and raw episode endpoint are three different quantities. All 328 score-table/raw RGB comparisons match by episode and timestep. We retain native decision positions without interpolation and use the raw endpoint axis for landmark eligibility: $L=\lfloor .25\times 84\rfloor=21$ leaves two successes and ten failures, below the five-per-class reporting rule. The episode-wise raw/dense/VLM ledger is retained with the numeric analysis. The earlier score-table horizon 36 remains useful for reproducing its native output coverage, but is not substituted for 84 in the current risk sets.

\paragraph{Recorded-step TWA and query cadence (exploratory).}
This added endpoint reuses the same exposed ten successes and ten failures; no model is rerun. With raw elapsed-step span $H=84$, an alarm at recorded step $a_i$ receives failure credit $1-a_i/H$, an undetected failure receives zero, and a successful episode receives one only if it has no alarm:
\begin{equation}
 \mathrm{TWA}_{H}=\frac 12\left[
 \frac{1}{n_F}\sum_{i\in F}\mathrm{I}(\mathrm{alarm}_i)(1-a_i/H)
 +\frac{1}{n_S}\sum_{i\in S}\mathrm{I}(\mathrm{no\ alarm}_i)\right].
\end{equation}
The same span is used for every arm; this is an auxiliary metric rather than a published Sentinel endpoint. On a zero-origin axis, span 84 corresponds to an inclusive index count 85. Using 83 as an endpoint-minus-one sensitivity leaves the point orderings below unchanged. Normalizing each method by its own last query would change the metric and can place a VLM query after its supposed horizon.

The original cumulative clock uses the .95 quantile of ten calibration-success endpoints, 22.2, and first alarms at 24 on its four-step decision grid. A required cadence sensitivity keeps that threshold but permits clock alarms only at the VLM's logged query positions. It therefore changes alarm opportunities, not calibration or the physical episodes. All 28 released prediction variants are retained; Table~\ref{tab:pushchair-twa} summarizes the five cumulative quantitative variants, three VLMs and both clocks. Released success labels, failure-alarm flags and detected first times match the export exactly. A non-alarming stored detection-time entry is the last scoring time, not an alarm, and receives no failure credit.

\begin{table}[ht]
\centering\small
\setlength{\tabcolsep}{7pt}
\begin{tabular}{lrrrrr}
\toprule
Arm & bAcc & $\mathrm{TWA}_{84}$ & TP/10 & FP/10 & Mean TP alarm step\\
\midrule
Clock, native four-step grid & .900 & .757 & 10 & 2 & 24.0\\
Clock, VLM query grid & .900 & .643 & 10 & 2 & 43.2\\
Diffusion ensemble, cumulative & .750 & .636 & 6 & 1 & 32.0\\
STAC, MSE & .750 & .631 & 7 & 2 & 28.6\\
STAC, MMD & .850 & .729 & 9 & 2 & 22.7\\
STAC, forward KL & .850 & .705 & 9 & 2 & 27.1\\
STAC, reverse KL & .850 & .710 & 8 & 1 & 29.5\\
GPT-4o & .950 & .745 & 9 & 0 & 38.2\\
Claude 3.5 Sonnet & .850 & .669 & 7 & 0 & 43.4\\
Gemini 1.5 Pro & .700 & .567 & 9 & 5 & 24.9\\
\bottomrule
\end{tabular}
\caption{Exploratory recorded-step PushChair timing on the same 20 exposed episodes. Quantitative variants retain their released cumulative calibration and VLMs their released verdicts. Mean times condition on detected failures, so their denominators differ. MMD denotes maximum mean discrepancy; KL direction follows the release keys. Clock cadence changes while its threshold stays fixed. These are eventual-outcome scores, with physical onset and response-readiness times unmeasured.}
\label{tab:pushchair-twa}
\end{table}

GPT-4o minus the native-grid clock is $+.050$ in bAcc but $-.0119$ in TWA, with paired 95\% interval $[-.1381,.1357]$. Against the query-matched clock its TWA difference is $+.1024$ $[-.0262,.2524]$. Both intervals include zero. The point reversal against the denser clock disappears under matched query opportunities, illustrating cadence sensitivity rather than a validated method ranking or universal protection from duration effects. Intervals use 20000 paired outcome-stratified episode-bootstrap draws with training, calibration and scores fixed, unknown acquisition clustering and no multiplicity correction. Independent per-episode credit arithmetic reconstructs all 60 metric cells to $2.3\times 10^{-16}$ and both contrast intervals; three boundary tests cover horizon credit, retained missed/false alarms and score-time/end-time confusion. Neither logged alarm time nor a time-weighted endpoint alone establishes pre-onset warning or delivered intervention utility.

\paragraph{SAFE real-robot releases.} In the WidowX release every rollout lasts exactly 50 steps, so the clock scores 0.50 under every protocol, and SAFE's eight training-free token scores reach ROC-AUC 0.45--0.54. In the Franka release the raw ceiling is 0.62; SAFE's default keeps only rollouts of full length, which fixes the horizon within each task. Pooled over tasks, the clock still scores 0.59 [0.56, 0.62] on that subset, because horizons and failure rates differ by task; SAFE's rebalancing of every task to a 0.5 success rate removes this channel, since a score that is constant within each task then has pooled AUC exactly 0.5. Token scores on the subset reach 0.42--0.55. The recorded outcome disagrees with the file-name suffix in 12 of 1,464 Franka rollouts; we use the recorded flag.

\clearpage
\section{Frozen FastUMI forecast-selection test (separate target)}
\label{sec:fastumi}
We use the gripper-pose streams of three single-arm FastUMI-100K~\cite{fastumi100k} tasks. Action columns equal the next-frame state and are never used as inputs. One-second histories predict the pose 0.5\,s ahead. Episodes are split by hash into TRAIN, VAL (early stopping only), DEV and TEST. Predictors were chosen on TRAIN/VAL and their weights frozen; selector features, ridge model and thresholds were developed on DEV. The TEST plan, with one primary endpoint and its decision rules, was hashed before TEST was read, and TEST was analyzed once (2,394 episodes; 37,050 FIT, 38,649 CAL and 76,817 EVAL windows within TEST). Predictor weights and selector definitions are not changed on TEST; the DEV procedure is rerun once, refitting the ridge selector on TEST-FIT episodes, setting its threshold on TEST-CAL and scoring TEST-EVAL. The section applies the paper's principle to selection: before a cue is credited, it is compared with the simpler alternative of a stronger predictor.

\begin{table}[t]
\centering\small
\resizebox{\columnwidth}{!}{%
\begin{tabular}{lrrrr}
\hline
Predictor & All & A at CAL thr. & A+cue at CAL thr. & $\Delta_{\rm cal}$ [95\% CI]\\
\hline
MLP-128 & 24.24 & 21.77 (.801) & 21.77 (.808) & $-.003$ [$-.087$, $.079$]\\
MLP-512 (main) & 22.47 & 20.13 (.801) & 20.07 (.806) & $+.059$ [$-.024$, $.138$]\\
GRU-256 & 22.17 & 19.67 (.798) & 19.62 (.805) & $+.049$ [$-.042$, $.135$]\\
\hline
\end{tabular}}
\caption{FastUMI frozen TEST. Mean position error (mm) adopting all forecasts and among forecasts accepted by each rule at its own calibration-fixed threshold (realized acceptance in parentheses). $\Delta_{\rm cal}$: A minus A+cue; positive favors the cue. Block bootstrap over 50-episode index blocks within task. Only the MLP-512 row is the prespecified primary endpoint.}
\label{tab:fastumi_strength}
\end{table}

\paragraph{Primary result.} For the main predictor, adding the correction-magnitude cue (position and rotation distance between the predictor's forecast and a constant-velocity extrapolation) changes accepted error by $+0.059$\,mm, 95\% CI $[-0.024, 0.138]$ (Table~\ref{tab:fastumi_strength}). By the frozen rules, an increment is \emph{not established} (lower bound $\le0$), and any increment is \emph{practically negligible} (upper bound below the 1\,mm minimum meaningful effect). Acceptance is non-inferior: the cue's rule accepts $+0.5$ percentage points more ($[-0.05, 1.09]$). On DEV the same contrast was $+0.177$ $[0.091, 0.265]$.

\paragraph{Secondary results (exploratory).} Strengthening the predictor lowers error by 1.77\,mm $[1.65, 1.88]$, replicating DEV (1.74). At exactly 80\% ranked acceptance the cue's increment is $+0.113$ $[0.049, 0.189]$\,mm; the deployable rule differs because the cue's rule also accepts slightly more forecasts. The DEV observation that the increment grows with predictor strength does not replicate: the MLP-512 minus MLP-128 difference in increment is $+0.029$ $[-0.033, 0.105]$\,mm (DEV $+0.074$ $[0.009, 0.131]$). Selection itself has a clear cost: at A's threshold the main predictor's rule rejects 19.9\% of forecasts, including 17.9\% of within-tolerance ones (50\,mm or $10^\circ$), while blocking 38.0\% of over-tolerance forecasts versus 19.9\% for same-count random rejection; exceedance among accepted forecasts falls from 10.2\% to 7.9\%. On DEV, an offline oracle showed that the decision-time selector captures 52\% of removable error for constant-velocity extrapolation but only 33\% for learned predictors; that analysis and the training and optimizer checks are in Supp.~E, the staged-residual control in Supp.~F.

\subsection{Forecast-selection protocol}
\label{sec:selective}
For continuous forecasts, the endpoint is the realized error of the adopted forecasts, not an outcome label, so the duration channel does not apply by construction. A frozen point predictor issues every forecast; selection cannot change its all-query error. A selector scores each forecast from decision-time evidence only: errors of earlier forecasts whose targets have already arrived, and current kinematics (group A). A candidate cue is added on top (A$+$cue). Selectors are fit on one subset, the adoption threshold is the 80th percentile of scores on a disjoint calibration subset, and evaluation uses a third subset. We report (i) error adopting all, (ii) error among adopted forecasts at the calibration-fixed threshold with realized acceptance, (iii) the increment $\mathrm{err}_{80}(A)-\mathrm{err}_{80}(A{+}\mathrm{cue})$ in mm (positive is better), and (iv) rejection cost: the share of within-tolerance forecasts rejected and of over-tolerance forecasts blocked, against a same-count random rejection.

\section{Earlier combined evidence overview (archival)}
\begin{figure}[ht]
\centering\includegraphics[width=\textwidth]{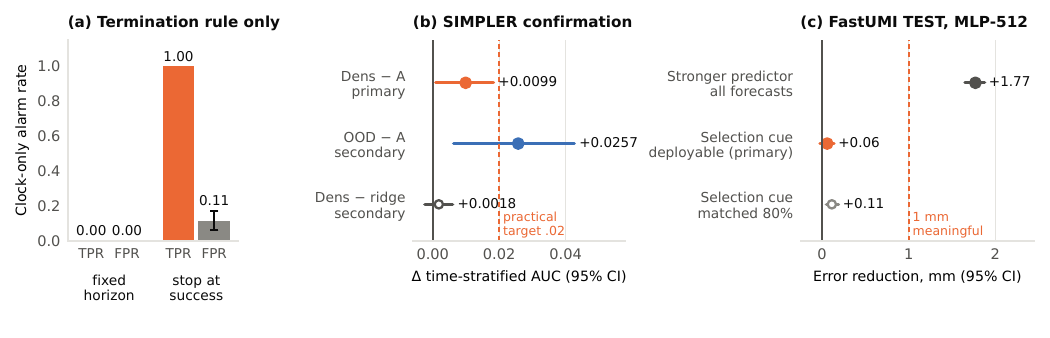}
\caption{Earlier combined overview retained for traceability. FastUMI forecasts and robot failure outcomes are distinct targets and are not pooled.}
\end{figure}

\clearpage
\section{Ranking audit: availability, estimand and uncertainty}
\label{app:ranking}
\begin{table}[ht]
\centering\small
\begin{tabular}{llccc}
\toprule
Release & Set & Full S/F & Quarter S/F & Calibration\\
\midrule
Sentinel & Cover 0527 h4 ca & --- & --- & not checked\\
Sentinel & Cover 0826 h4 na & 43/1 $^\dagger$ & 43/1 $^\dagger$ & 38\\
Sentinel & Cover 0904 h4 ss & 1/33 $^\dagger$ & 1/33 $^\dagger$ & 38\\
Sentinel & PushChair (real) & 10/10 & 2/10 $^\dagger$ & 10\\
Sentinel & PushT 0525 h2 ca & --- & --- & not checked\\
Sentinel & PushT 0525 h2 hh & 9/21 & 8/21 & 15\\
Sentinel & PushT 0525 h2 na & 15/15 & 13/15 & 15\\
Sentinel & PushT 0525 h4 ca & --- & --- & not checked\\
Sentinel & PushT 0525 h4 hh & 7/23 & 6/23 & 11\\
Sentinel & PushT 0525 h4 na & 15/15 & 13/15 & 11\\
Sentinel & PushT 0525 h8 ca & --- & --- & not checked\\
Sentinel & PushT 0525 h8 hh & 14/36 & 13/36 & 21\\
Sentinel & PushT 0525 h8 na & 21/29 & 17/29 & 21\\
Sentinel & PushT 0526 h8 hh & 14/36 & 13/36 & 21\\
Sentinel & PushT 0526 h8 na & 21/29 & 16/29 & 21\\
Sentinel & PushT 0527 h8 hh & 16/34 & 14/34 & 21\\
Sentinel & PushT 0527 h8 na & 16/34 & 16/34 & 21\\
FIPER & sorting & 255/145 & 255/142 & official run\\
FIPER & stacking & 315/485 & 314/485 & official run\\
FIPER & push\_t & 102/198 & 98/198 & official run\\
FIPER & pretzel & 10/10 & 10/10 & official run\\
FIPER & push\_chair & 10/10 & 4/10 $^\dagger$ & official run\\
SAFE-LIBERO & 10 tasks & 7/10 tasks valid & 7/10 tasks valid & not used\\
SAFE-WidowX & 8 tasks & 8/8 tasks valid & 8/8 tasks valid & not used\\
SAFE-Franka & 13 tasks & 13/13 tasks valid & 13/13 tasks valid & not used\\
\bottomrule
\end{tabular}
\caption{All 25 released sets, including missing-score and invalid-risk-set cases. Sentinel names retain task, release date, execution horizon (h) and randomization condition; full scored source paths are in \texttt{input/inventory.json}. S/F counts concern scored episodes, not always the complete raw release. $^\dagger$Fewer than five of a class: not evaluable. SAFE entries count eligible task cohorts; its calibration is not used by our training-free adapter, not asserted absent. Four Sentinel sets have no detector scores (---). FIPER exports first-alarm indices, not continuous scores. FIPER and Sentinel PushChair may share episodes. Full counts and later landmarks are in the accompanying inventory CSV.}
\end{table}

The numerical interchange contract requires \texttt{rollout\_id, task, outcome, length, step, score, is\_calibration}, plus optional method and benchmark identifiers. Failure is positive. Duplicate episode steps, nonfinite scores and split/label disagreements are rejected. The generic adapter fits a successful-calibration q95 episode-max threshold; official-release adapters retain the original method's calibration. A quantile rule is not a certified safety guarantee.

\paragraph{FIPER implementation note.} We ran the official FIPER implementation unchanged, adding only a loader for the converted arrays. Beyond the longest calibration rollout, time-varying thresholds are extended as in the released code. The clock score is identical across calibration rollouts and strictly increasing in the step index, so by construction its first alarm index does not depend on how the threshold is extended; we did not rerun the audit with an alternative extension.

The AUC target is final failure conditional on still running at landmark $L$, not a cumulative/dynamic event-time survival AUC. Multiple failure times and intervention utility are not observed by this analysis. A full-to-landmark rank change therefore describes protocol sensitivity, not a choice of the optimal deployment monitor. For first-alarm-only FIPER exports, we average balanced accuracy over the original ten quantiles and five seeds; binary verdicts are never reported as continuous confidence scores. For Sentinel, raw converted episode timestamps define the observation horizon: a sparse VLM's last scoring time is not the episode's end. No scores are imputed between native decision positions.

The same original-episode bootstrap draw is used for every score and protocol, then filtered to its risk set; 1000 draws are stratified by outcome within task. Pointwise intervals are exploratory and not corrected for the tested tasks, methods and horizons. Native methods and seeds are correlated variants, not independent replications. Task-mixture intervals stratify by task and outcome and keep cross-fitted task references fixed; they exclude fitting and task-generalization uncertainty. All task-macro means require at least five outcomes of each class. Full CSV/JSON outputs list omitted points and their class counts. The last available prefix score is supplied as a sensitivity analysis; no best-scoring summary is selected after inspection.

\begin{figure}[ht]
\centering\input{all_quarter_rankings}
\caption{All evaluable quarter-horizon balanced-accuracy rank correlations, including weak or stable findings. Pointwise paired episode-bootstrap intervals; overlapping settings are not independent discoveries. Complete source-set identifiers are in \texttt{rankings.csv}.}
\end{figure}

\section{RGB sensitivity: visual novelty is not outcome reliability}
\label{app:rgb-sensitivity}
This exploratory pilot uses 20 already-inspected real PushChair episodes from Sentinel~\cite{agia2024stac}, with 134 available RGB frames and 39 successful reference frames. A frozen torchvision ResNet-18 ImageNet-1K V1 backbone supplies normalized 512-dimensional pooled features. The score is mean cosine distance to the five nearest successful reference frames; episode scores are maximum available prefix scores. The inclusive common-prefix cutoff $L=13$ includes recorded steps through 12, the shortest raw endpoint, with observations strictly before $L$. It retains ten successes and ten failures, so the clock has no within-prefix discrimination. Brightness and contrast controls have AUROC .22 and .34 in their prespecified lower-quality orientation. This is not a comparison against a newly trained full policy-state baseline.

We hold reference images, encoder and recorded outcome labels fixed, changing only evaluation RGB: intensity factor .3, Gaussian blur (21-pixel kernel, standard deviation 3), or a central black rectangle covering one quarter of image area. The specification was hashed before this computation but after inspection of the release; it is not data-naive preregistration. No physical or policy rerun occurs. Zero exact RGB-byte hashes are shared between reference and evaluation frames; this check does not rule out near-duplicates or shared acquisition conditions.

\begin{center}\begin{minipage}{.95\textwidth}
\centering
\input{vision_sensitivity_20261109}
\captionsetup{hypcap=false}
\captionof{figure}{\textbf{Higher visual novelty can coincide with worse outcome ranking.} Same recorded episodes and common prefix: visual distance rises under darkening, blur and occlusion, while final-failure AUROC falls. Intervals use 1,000 paired outcome-stratified episode bootstrap draws with fixed encoder/reference. Small-sample exploratory illustration; image interventions do not create new failure labels and are not closed-loop robustness tests.}
\label{fig:rgb-sensitivity}
\end{minipage}\end{center}

Original AUROC is .94 [.79, 1.00]; dark, blur and occlusion score .77 [.52, .95], .47 [.21, .75] and .33 [.10, .60]. Paired differences from original have 95\% intervals $[-.39,-.02]$, $[-.74,-.21]$ and $[-.86,-.34]$. No multiplicity correction or reference-refitting uncertainty is included. This result supports separating sensitivity to image appearance from discrimination of the recorded task outcome; it does not show that every visual novelty monitor fails, nor that the original .94 would generalize. The 0.25/0.5/0.75-horizon risk sets are invalid (2/10, 0/8, 0/1 success/failure); no curves are filled in. Cached preprocessing/encoding timings exclude capture, transport, queuing and delivery, and are not reported as real-time end-to-end latency.

\clearpage
\section{Alarm overlap with successful-calibration duration}
\label{app:late-alarms}
This exploratory diagnostic fixes the new computation after prior FIPER results have been inspected. It uses only previously converted NPZ/JSON, inherited S-paper window/style settings, and quantile .95. For each task, $c$ is the higher empirical .95 quantile of successful calibration episode lengths. A zero-based first-alarm index $a\ge c$ is called late relative to that boundary; a clock alarms at $c$ only when episode length exceeds $c$. Calibration success durations do not guarantee that 95\% of test successes end before $c$.

\begin{table}[ht]
\centering\small
\begin{tabular}{lrrrrr}
\toprule
Task & $c$ & Failures & Late/detected failures [95\% CI] & Early failure TPR & Clock TPR/FPR\\
\midrule
Sorting & 52 & 145 & 22.4\% [15.1, 30.8] & 0.574 & 0.559/0.161\\
Stacking & 91 & 485 & 3.4\% [1.9, 5.3] & 0.810 & 0.990/0.035\\
Push T & 29 & 198 & 21.0\% [15.5, 27.3] & 0.778 & 1.000/0.108\\
Pretzel & 54 & 10 & 34.0\% [8.0, 62.0] & 0.660 & 0.900/0.200\\
Push Chair & 7 & 10 & 70.0\% [40.0, 100.0] & 0.300 & 1.000/0.000\\
\bottomrule
\end{tabular}
\caption{FIPER's alarm overlap with calibration-success duration, fixed .95 quantile (not the earlier quantile-averaged main table). Five-seed mean ratios and paired outcome-stratified episode-bootstrap intervals, conditional on calibration and selected settings. Early TPR divides by all failed episodes, not just detected ones. PushChair and Pretzel each have ten failed test episodes.}
\end{table}

The overlap varies markedly by task: it does not support claiming that most FIPER alarms are universally late or that late alarms have zero incremental value. Sorting's clock FPR is .161 despite its calibration-success .95 boundary; the threshold is not a guarantee of test false-alarm control. All six methods and both late-alarm denominators are supplied in the numerical artifact. First-alarm episode lengths and outcomes match the raw numerical exports for every analyzed file. Bootstrap resamples episodes once across all seeds (1,000 outcome-stratified draws); thresholds and selected settings are not refitted. This descriptive statistic is neither a conditional-independence test nor an intervention-value estimate. A matched metadata-plus-score experiment requires a separately fixed fitting and evaluation protocol.

\paragraph{A separate test-duration diagnosis and an exact metric decomposition.}
The preceding calibration-boundary computation is not the test-success-duration diagnostic. Recomputing with the .95 quantile of \emph{test} success lengths and all ten quantiles under inherited settings gives late shares of 0\%, 14.4\%, 0.9\%, 0\% and 70\% for Sorting, Stacking, PushT, Pretzel and PushChair. These use test outcomes and remain purely descriptive. Early alarms do not by themselves establish useful visual reasoning.

On PushT, the clock and FIPER have TPR 1.000/.983 and TNR .941/.441. Since balanced accuracy is $(\mathrm{TPR}+\mathrm{TNR})/2$, its clock-minus-FIPER difference .259 is the sum of .008 from sensitivity and .250 from specificity (rounded). On Sorting the corresponding components are $-.107$ and $+.171$: the clock misses more failures, but raises fewer false alarms. On Stacking it never alarms and loses overall despite perfect specificity. Thus a high clock score does not establish that the visual detector waits until success episodes end. No new uncertainty interval or causal claim is attached to this arithmetic decomposition. Both diagnostics retain their own threshold definitions and denominators. An alarm-specific clock chosen using test-success specificity is also an oracle diagnostic, not a deployable replacement or intervention benefit.

%% file: fig_spatial_age.tex
\begin{figure}[t]
\centering
\begin{tikzpicture}[font=\scriptsize, x=1cm, y=1cm]
\node[anchor=west, font=\scriptsize\bfseries] at (0,2.75) {(a) Evidence age per target};
\node[anchor=west, align=left, text=black!70] at (0,2.3) {stream arriving, receipt gap 33\,ms;\\ latest frame age 20\,ms};
\draw[black!40] (0,0.15) -- (3.2,0.15) -- (3.7,0.75) -- (0.5,0.75) -- cycle;
\fill[blue!65] (0.6,1.15) circle (0.17);
\node[text=blue!70!black] at (0.6,1.55) {A: 20\,ms};
\fill[orange!85!black] (3.2,1.15) circle (0.17);
\node[text=orange!85!black] at (3.2,1.55) {B: 800\,ms};
\fill[black!40] (1.9,1.15) circle (0.17);
\node[text=black!55] at (1.9,1.55) {C: unknown};
\begin{scope}[xshift=4.15cm]
\node[anchor=west, font=\scriptsize\bfseries] at (0,2.75) {(b) Evidence used at decision};
\def\sc{0.0034}
\draw[black!60] (0,0.05) -- (3.4,0.05);
\foreach \t/\lab in {0/0, 500/500, 1000/1000\,ms} {\draw[black!60] (\t*\sc,0.05) -- (\t*\sc,-0.02) node[below=-1pt] {\lab};}
\draw[dashed, black!70] (1000*\sc,0.05) -- (1000*\sc,2.4);
\node[anchor=east, text=black!70] at (1000*\sc-0.05,2.4) {decision};
\node[anchor=west] at (0,2.0) {A};
\fill[blue!65] (980*\sc,2.0) circle (0.06);
\fill[blue!65] (996*\sc-0.05,1.95) rectangle ++(0.1,0.1);
\node[anchor=west] at (0,1.25) {B};
\draw[orange!85!black, line width=1.2pt] (200*\sc,1.25) -- (1000*\sc,1.25);
\fill[orange!85!black] (200*\sc,1.25) circle (0.06);
\fill[orange!85!black] (260*\sc-0.05,1.2) rectangle ++(0.1,0.1);
\node[orange!85!black, below=1pt] at (230*\sc,1.2) {observed, ready};
\node[anchor=west] at (0,0.5) {C};
\node[anchor=west, text=black!55] at (0.3,0.5) {no direct observation};
\end{scope}
\end{tikzpicture}
\caption{\textbf{Prospective: the scene has a clock (schematic, synthetic values).} (a) Global checks pass because frames keep arriving, yet the evidence supporting target B was last observed 800\,ms before the decision, and target C has no direct observation (unknown, not fresh). (b) Observation (dot) and readiness (square) times of the evidence each target estimate actually uses, relative to the decision (dashed). Ages describe evidence freshness, not perception correctness or safety. No measured result is shown.}
\label{fig:spatial-age}
\end{figure}

%% file: all_quarter_rankings.tex
\resizebox{.85\textwidth}{!}{\begin{tikzpicture}[x=3cm,y=.3cm,font=\scriptsize]
\draw[dashed,gray](0,-1)--(0,14);
\draw[blue] (-0.4140,0)--(0.6901,0);\fill[blue] (0.0000,0) circle(1.3pt);\node[anchor=east] at(-1.1,0){FIPER/pretzel};
\draw[blue] (-0.8667,1)--(0.2000,1);\fill[blue] (-0.6000,1) circle(1.3pt);\node[anchor=east] at(-1.1,1){FIPER/push\_t};
\draw[blue] (-1.0000,2)--(-0.0667,2);\fill[blue] (-0.8667,2) circle(1.3pt);\node[anchor=east] at(-1.1,2){FIPER/sorting};
\draw[blue] (-0.0667,3)--(0.4667,3);\fill[blue] (0.3333,3) circle(1.3pt);\node[anchor=east] at(-1.1,3){FIPER/stacking};
\draw[blue] (-0.7616,4)--(0.7348,4);\fill[blue] (-0.7011,4) circle(1.3pt);\node[anchor=east] at(-1.1,4){Sentinel/PushT 0525 h2 hh};
\draw[blue] (-0.6523,5)--(-0.4698,5);\fill[blue] (-0.5427,5) circle(1.3pt);\node[anchor=east] at(-1.1,5){Sentinel/PushT 0525 h2 na};
\draw[blue] (-0.6895,6)--(0.6773,6);\fill[blue] (-0.5120,6) circle(1.3pt);\node[anchor=east] at(-1.1,6){Sentinel/PushT 0525 h4 hh};
\draw[blue] (-0.7210,7)--(0.4472,7);\fill[blue] (-0.1723,7) circle(1.3pt);\node[anchor=east] at(-1.1,7){Sentinel/PushT 0525 h4 na};
\draw[blue] (-0.7663,8)--(-0.3511,8);\fill[blue] (-0.7208,8) circle(1.3pt);\node[anchor=east] at(-1.1,8){Sentinel/PushT 0525 h8 hh};
\draw[blue] (-0.4978,9)--(-0.3494,9);\fill[blue] (-0.4523,9) circle(1.3pt);\node[anchor=east] at(-1.1,9){Sentinel/PushT 0525 h8 na};
\draw[blue] (-0.6816,10)--(0.3038,10);\fill[blue] (-0.1162,10) circle(1.3pt);\node[anchor=east] at(-1.1,10){Sentinel/PushT 0526 h8 hh};
\draw[blue] (-0.5872,11)--(-0.1230,11);\fill[blue] (-0.5240,11) circle(1.3pt);\node[anchor=east] at(-1.1,11){Sentinel/PushT 0526 h8 na};
\draw[blue] (-0.7090,12)--(0.4143,12);\fill[blue] (-0.1989,12) circle(1.3pt);\node[anchor=east] at(-1.1,12){Sentinel/PushT 0527 h8 hh};
\draw[blue] (-0.7454,13)--(-0.2474,13);\fill[blue] (-0.6835,13) circle(1.3pt);\node[anchor=east] at(-1.1,13){Sentinel/PushT 0527 h8 na};
\draw[->](-1,-1)--(1.1,-1);\foreach \x in {-1,-.5,0,.5,1}\node[below] at(\x,-1){\x};
\node at(0,-3){Kendall $\tau_b$: full vs quarter-horizon balanced-accuracy rank};
\end{tikzpicture}}

%% file: vision_sensitivity_20261109.tex
\resizebox{.95\textwidth}{!}{\begin{tikzpicture}[x=1cm,y=1cm,font=\small]
\node[font=\bfseries] at (2.5,3.7){(a) Visual novelty increases};
\node[font=\bfseries] at (9.1,3.7){(b) Outcome ranking deteriorates};
\draw[->](.3,.4)--(5.6,.4);\draw[->](.3,.4)--(.3,3.3);
\draw[->](6.7,.4)--(12,.4);\draw[->](6.7,.4)--(6.7,3.3);
\draw[dashed,gray](6.7,1.8)--(11.8,1.8);
\fill[blue!60](0.65,.4) rectangle (1.35,0.605);
\node[above] at(1.0,0.605){0.033};
\draw[blue](7.4,2.612)--(7.4,3.200);\fill[blue](7.4,3.032) circle(2pt);
\node[below,font=\scriptsize] at(1.0,.35){Original};
\node[below,font=\scriptsize] at(7.4,.35){Original};
\fill[blue!60](1.9,.4) rectangle (2.6,0.973);
\node[above] at(2.25,0.973){0.092};
\draw[blue](8.65,1.856)--(8.65,3.060);\fill[blue](8.65,2.556) circle(2pt);
\node[below,font=\scriptsize] at(2.25,.35){Dark};
\node[below,font=\scriptsize] at(8.65,.35){Dark};
\fill[blue!60](3.15,.4) rectangle (3.85,1.891);
\node[above] at(3.5,1.891){0.240};
\draw[blue](9.9,0.988)--(9.9,2.501);\fill[blue](9.9,1.716) circle(2pt);
\node[below,font=\scriptsize] at(3.5,.35){Blur};
\node[below,font=\scriptsize] at(9.9,.35){Blur};
\fill[blue!60](4.4,.4) rectangle (5.1,3.093);
\node[above] at(4.75,3.093){0.433};
\draw[blue](11.15,0.679)--(11.15,2.080);\fill[blue](11.15,1.324) circle(2pt);
\node[below,font=\scriptsize] at(4.75,.35){Occlusion};
\node[below,font=\scriptsize] at(11.15,.35){Occlusion};
\node[rotate=90]at(-.3,1.8){Cosine distance};\node[rotate=90]at(6.1,1.8){AUROC};
\foreach \v in {0,.5,1}\node[anchor=east]at(6.65,{.4+\v*2.8}){\v};
\end{tikzpicture}}

%% file: robustness_extension_20261110.tex
\clearpage
\section{Additional robustness checks (exploratory)}
\label{app:robustness-extension}
These computations were specified and hashed after the prior audits and RGB pilot were inspected; they are exploratory and leave the frozen \mbox{SIMPLER} endpoint unchanged. The artifact contains code, aggregate tables, hashes and unavailable cases.

\subsection{Holding the evaluation population fixed}
At $L=\lfloor fT_{\max}\rfloor$, $f\in\{.25,.50,.75\}$, both full-rollout and prefix rankings use identical episodes with $T_i>L$. Another diagnostic fixes $T_i>\lfloor .75T_{\max}\rfloor$ across all prefixes: a retrospective survivor cohort, not a deployed population. Original settings and variant averaging remain fixed, including FIPER's test-selected S-paper settings. bAcc and continuous-only AUROC remain separate. Evaluable cells require five outcomes per class and three methods; 1000 outcome-stratified, paired episode draws give pointwise intervals. Constant rankings remain undefined; missing scores are not interpolated.

\begin{figure}[!ht]
\centering
\input{robustness_extension_figure_20261110}
\caption{\textbf{Robustness of attribution (exploratory).} (a) Sorting full-versus-quarter-prefix bAcc ranks, with and without the clock; bars: 95\% episode-bootstrap intervals. Late survivors form a different population. (b) TEST-only versus CAL+TEST interval widths on 132 Sentinel dataset/variant combinations sharing episodes; the diagonal denotes equal width. (c) Same 20 real-robot PushChair episodes under fixed image edits; pointwise AUROC intervals are tabulated below. Severity is within-family and outcomes are unchanged.}
\label{fig:robustness-extension}
\end{figure}
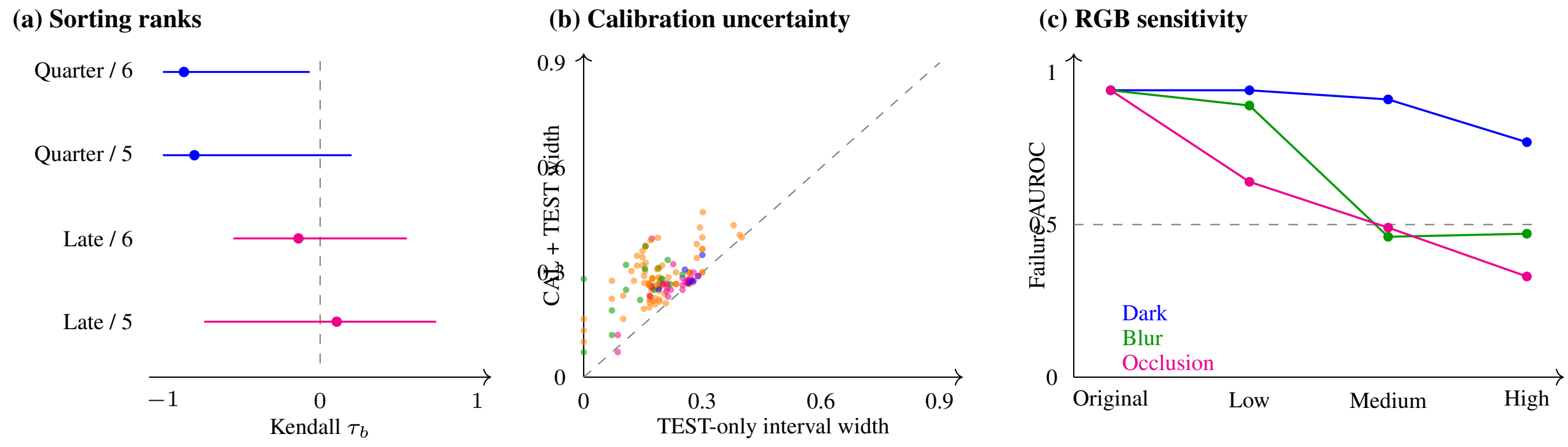

\begin{table}[!ht]
\centering\small
\begin{tabular}{llrrll}
\hline
Task & Shared population & Success & Failure & Six scores: $\tau_b$ [CI] & Five detectors: $\tau_b$ [CI]\\
\hline
Sorting & Quarter risk set & 255 & 142 & $-0.87$ [-1.00, -0.07] & $-0.80$ [-1.00, 0.20]\\
Sorting & Late survivors & 26 & 79 & $-0.14$ [-0.55, 0.55] & $0.11$ [-0.74, 0.74]\\
PushT & Quarter risk set & 98 & 198 & $-0.60$ [-0.87, 0.20] & $-0.40$ [-0.80, 0.80]\\
PushT & Late survivors & 13 & 198 & $-0.73$ [-0.73, 0.47] & $-0.60$ [-0.80, 1.00]\\
Stacking & Quarter risk set & 314 & 485 & $0.33$ [-0.07, 0.47] & $0.00$ [-0.40, 0.40]\\
Stacking & Late survivors & 34 & 481 & $0.60$ [-0.47, 0.73] & $0.80$ [-0.40, 1.00]\\
\hline
\end{tabular}
\caption{Full-versus-quarter-prefix bAcc ranking on identical episodes. Six scores include the clock; five detectors exclude it. Both columns use the same episodes. Pointwise exploratory intervals, without multiplicity adjustment.}
\end{table}

Sorting's six-score reversal persists on its identical 397-episode quarter risk set. Excluding the clock gives $\tau_b=-.80$ with interval $[-1.00,.20]$, so a detector-only reversal is not established at this precision. Restricting all horizons to the 105 late survivors weakens the six-score quarter-prefix comparison to $-.14$ $[-.55,.55]$. On Sentinel PushT, the same-quarter-cohort bAcc comparison remains negative in all ten evaluable sets, with five intervals below zero. No Sentinel PushT common late-survivor set meets the outcome-count rule. These checks distinguish protocol sensitivity from a universal ordering, not causal utility or a universally corrected leaderboard.

\clearpage
\paragraph{Worked example: what the Sorting audit changes.}
All numbers in this example concern the released FIPER Sorting task; it introduces no new analysis or independent validation.
Under the original full-rollout protocol, FIPER has bAcc .661 and TWA .542. Passing a sensor-free clock through calibration and alarm evaluation gives bAcc .726 but TWA .511: a bAcc advantage need not mean better timely evidence.
At the quarter-horizon landmark, among 255 successes and 142 failures still running, FIPER has bAcc .547 and the clock .500. Across six scores, the full-rollout/quarter-prefix rank correlation is $-.87$; the reversal persists when both rankings use these same 397 episodes (Table above).
This goes beyond asking whether a clock scores well: it checks whether the method ordering survives an equal-time comparison on the same population. However, the detector-only interval crosses zero, and the late-survivor comparison weakens. The practical implication is to withhold a universal method choice based on full-rollout bAcc alone and carry multiple candidates into held-out evaluation, not to declare a replacement winner. This is a descriptive, exploratory selection diagnostic; no new selection was independently validated here.

\subsection{Propagating calibration uncertainty}
For every eligible released Sentinel cumulative-score variant, the threshold is the .95 quantile of successful calibration episodes' \emph{final} cumulative scores. We verify reconstructed per-step decisions against the released predictions. One variant disagrees at one step (score 46.00353 versus threshold 46.00258); its cause is unresolved. Its two protocol rows remain in the artifact but are excluded from E2 summaries. No tolerance is tuned to remove the discrepancy. The other variants have zero mismatches.

CAL+TEST intervals independently resample successful calibration episodes and outcome-stratified test episodes (1000 draws). Within each replicate, detector and clock share the CAL and TEST draws. The clock uses observed score-grid steps for this check, not synthesized dense control ticks; these rows do not replace the original raw-horizon audit. Repeated split-local episode IDs do not establish overlap or independence, and no session-level independence is claimed. This check conditions on released scores: it propagates threshold uncertainty, not training or setting-selection uncertainty. The FIPER first-alarm export cannot support recalibration, and the current SAFE export has no score-calibration split.

\begin{table}[!ht]
\centering\small
\begin{tabular}{lrrrr}
\hline
Family & Dataset/variant rows & Median TEST-only width & Median CAL+TEST width & Wider rows\\
\hline
Diffusion ensemble & 7 & 0.268 & 0.276 & 5\\
Embedding similarity & 18 & 0.156 & 0.276 & 18\\
Loss function & 24 & 0.238 & 0.269 & 21\\
Temporal consistency & 83 & 0.186 & 0.269 & 78\\
\hline
\end{tabular}
\caption{Descriptive bAcc interval widths at the full horizon after reconstruction-parity checks. The 132 rows reuse releases and episodes; they are not 132 independent datasets. Threshold uncertainty widens 122 rows, leaves seven unchanged and narrows three. It need not widen every percentile interval.}
\end{table}

For temporal consistency, median interval width changes from .186 to .269 across 83 dataset/variant rows. This illustrates uncertainty omitted by conditional TEST-only intervals; it is not a significance test on interval widths, proof of calibration failure, or evidence that every original finding disappears.

\subsection{Fixed RGB corruption severities}
We retain the prior pilot's frozen ResNet-18, successful demo reference, five-neighbour cosine score and common observed prefix before the shortest raw episode horizon. All 20 test episodes (ten successes, ten failures) appear under every condition. Brightness multiplies RGB by .7/.5/.3; Gaussian blur uses kernel 21 and $\sigma=1/2/3$; centered black-square side fractions .25/.375/.50 cover 6.25/14.06/25\% of the image. There is no retraining or severity selection.

\begin{table}[!ht]
\centering\small
\begin{tabular}{llrll}
\hline
Edit & Fixed parameter & Mean max novelty & Failure AUROC [CI] & Paired AUROC change [CI]\\
\hline
Original & --- & 0.033 & 0.94 [0.79, 1.00] & +0.00 [+0.00, +0.00]\\
Brightness & $\times0.7$ & 0.040 & 0.94 [0.79, 1.00] & +0.00 [-0.04, +0.04]\\
Brightness & $\times0.5$ & 0.054 & 0.91 [0.75, 1.00] & -0.03 [-0.14, +0.06]\\
Brightness & $\times0.3$ & 0.092 & 0.77 [0.54, 0.95] & -0.17 [-0.37, +0.00]\\
Blur & $\sigma=1$ & 0.058 & 0.89 [0.69, 1.00] & -0.05 [-0.15, +0.00]\\
Blur & $\sigma=2$ & 0.140 & 0.46 [0.21, 0.72] & -0.48 [-0.73, -0.24]\\
Blur & $\sigma=3$ & 0.240 & 0.47 [0.20, 0.76] & -0.47 [-0.73, -0.19]\\
Occlusion & 6.30\% area & 0.158 & 0.64 [0.39, 0.88] & -0.30 [-0.54, -0.11]\\
Occlusion & 14.03\% area & 0.284 & 0.49 [0.22, 0.78] & -0.45 [-0.71, -0.19]\\
Occlusion & 25.00\% area & 0.433 & 0.33 [0.10, 0.58] & -0.61 [-0.86, -0.31]\\
\hline
\end{tabular}
\caption{Pointwise, paired episode-bootstrap intervals (1000 draws), without multiplicity correction. Mean novelty is the mean of episode-prefix maxima. Severity levels are fixed, not optimized for a favorable result.}
\end{table}

Mild darkening preserves the .94 AUROC; stronger blur and occlusion reduce discrimination while novelty rises. The blur AUROC is not monotonic between $\sigma=2$ and 3. This is measurement sensitivity on a previously inspected small release, not a physical perturbation benchmark, a new task-failure label, sensor-to-action timing, or evidence of intervention or safety gains.

\section{\mbox{SIMPLER} replication with GR00T N1.7}
\label{app:gr00t}

\paragraph{Setup.} We repeat the termination control of Sec.~5.4 of the main paper with the released GR00T N1.7 Bridge checkpoint~\cite{bjorck2025gr00t}, used frozen; nothing is trained. Simulator, tasks, horizons, logging and resampling unit are unchanged. The plan, including the primary endpoint and the sample-size rule, was hash-frozen before any analysis rollout. Sample sizes were then fixed from 640 development episodes (targets: at least 40 failures and 20 successes per task). The 816 confirmation episodes were generated once and analysed once. The two policies are not pooled.

\paragraph{Primary endpoint.} For the clock-only any-step detector ($q=0.95$), balanced accuracy is $0.500$ at the fixed horizon and $0.965$ $[0.946, 0.981]$ when the same episodes stop at success (TPR $1.00$, FPR $0.070$). The difference is $+0.465$ $[+0.446, +0.481]$, confirmed under the prespecified rule (lower bound above zero). The endpoint is nearly fixed by construction: every failure runs to the horizon, and the threshold is a quantile of success lengths. Its empirical content is that successes end before the horizon. On the three non-stacking tasks at $q=0.90$, TPR/FPR is $1.00/0.108$ $[0.063, 0.163]$, against $1.00/0.111$ for Octo. Table~\ref{tab:gr00t} gives per-task values.

\begin{table}[h]
\centering
\small
\begin{tabular}{lrrrc}
\hline
Task & $N$ & Success & Failure & bAcc difference \\
\hline
Carrot & 160 & 74 & 86 & $+0.459$ $[+0.417, +0.488]$ \\
Spoon & 176 & 126 & 50 & $+0.472$ $[+0.439, +0.493]$ \\
StackCube & 320 & 28 & 292 & $+0.429$ $[+0.289, +0.500]$ \\
Eggplant & 160 & 88 & 72 & $+0.472$ $[+0.447, +0.494]$ \\
\hline
Pooled & 816 & 316 & 500 & $+0.465$ $[+0.446, +0.481]$ \\
\hline
\end{tabular}
\caption{GR00T N1.7 on \mbox{SIMPLER}, confirmation episodes. Clock-only any-step detector, $q=0.95$: balanced accuracy when episodes stop at success minus balanced accuracy at the fixed horizon. Intervals are 95\% configuration-bootstrap intervals.}
\label{tab:gr00t}
\end{table}

\paragraph{Secondary analyses (no confirmation claim).} The frozen Dens-minus-A contrast is $+0.036$ $[0.013, 0.059]$. It is carried by Carrot and StackCube, and its interval includes zero once the first 10\% of steps are dropped ($+0.026$ $[-0.019, 0.073]$). The simplified FIPER-style score minus A is $+0.043$ $[0.016, 0.072]$. Under stop-at-success, any-step AUROC is $1.000$ for episode length and $0.905$ $[0.853, 0.945]$ for the FIPER-style score. Among episodes still running at $0.25T$, $0.5T$ and $0.75T$, where the clock is uninformative, the FIPER-style score reaches $0.731$ $[0.632, 0.809]$, $0.759$ $[0.633, 0.851]$ and $0.682$ $[0.574, 0.794]$. These risk sets hold 500 failures with 226, 81 and 42 successes; Eggplant contributes one success at the last two landmarks. Within configuration and time bin, the FIPER-style score reaches $0.723$ $[0.679, 0.769]$ against $0.666$ $[0.655, 0.680]$ for elapsed time. With this policy the visual score therefore keeps signal that waiting does not explain, while the any-step protocol still ranks the clock first.

\paragraph{Deviations.} Fixed-horizon any-step values for the two detector scores were not computed, because the frozen row table ends at first success. The primary-endpoint script was written after generation but before any confirmation outcome was read. The landmark script was written after the primary result was seen. One episode that succeeds at the first step has no logged rows and is excluded from the secondary analyses.

\section{Simulation engineering scope note}
\label{app:m1-skai}
Separate M1 wheel-legged locomotion training and validation of a supplied simulation scene are engineering development, not additional failure-monitor benchmarks. They establish no obstacle-traversal, hardware-transfer or safety benefit. These proximal policy optimization (PPO) runs did not use success-guided sampling (SGS)~\cite{zhang2026balanced}; SGS remains a related training-allocation method, not the source of PPO. Engineering run details are retained separately from the audit evidence.

%% file: robustness_extension_figure_20261110.tex
\resizebox{\textwidth}{!}{%
\begin{tikzpicture}[x=1cm,y=1cm,font=\scriptsize]
\node[anchor=west,font=\footnotesize\bfseries] at (0,4.4) {(a) Sorting ranks};
\draw[->] (1.6,.55)--(5.3,.55);
\draw[dashed,gray] (3.45,.7)--(3.45,4.05);
\node at (1.75,.3) {$ -1 $};
\node at (3.45,.3) {$ 0 $};
\node at (5.15,.3) {$ 1 $};
\node at (3.45,0) {Kendall $\tau_b$};
\node[anchor=east] at (1.55,3.850) {Quarter / 6};
\draw[blue,line width=.6pt] (1.7500,3.850)--(3.3367,3.850);
\fill[blue] (1.9767,3.850) circle (1.6pt);
\node[anchor=east] at (1.55,2.950) {Quarter / 5};
\draw[blue,line width=.6pt] (1.7500,2.950)--(3.7900,2.950);
\fill[blue] (2.0900,2.950) circle (1.6pt);
\node[anchor=east] at (1.55,2.050) {Late / 6};
\draw[magenta,line width=.6pt] (2.5115,2.050)--(4.3885,2.050);
\fill[magenta] (3.2154,2.050) circle (1.6pt);
\node[anchor=east] at (1.55,1.150) {Late / 5};
\draw[magenta,line width=.6pt] (2.1956,1.150)--(4.7044,1.150);
\fill[magenta] (3.6292,1.150) circle (1.6pt);
\node[anchor=west,font=\footnotesize\bfseries] at (5.8,4.4) {(b) Calibration uncertainty};
\draw[->] (6.3,.55)--(10.4,.55);
\draw[->] (6.3,.55)--(6.3,4.05);
\draw[dashed,gray] (6.3,.55)--(10.15,3.95);
\node at (8.35,0) {TEST-only interval width};
\node[rotate=90] at(5.95,2.3){CAL + TEST width};
\fill[blue,opacity=.55] (7.5833,1.8722) circle (1pt);
\fill[orange,opacity=.55] (7.5833,1.6833) circle (1pt);
\fill[orange,opacity=.55] (7.5833,1.6833) circle (1pt);
\fill[orange,opacity=.55] (7.5833,2.0611) circle (1pt);
\fill[orange,opacity=.55] (8.0111,2.0611) circle (1pt);
\fill[orange,opacity=.55] (7.1148,1.3895) circle (1pt);
\fill[orange,opacity=.55] (7.0130,1.3595) circle (1pt);
\fill[orange,opacity=.55] (7.2167,1.6294) circle (1pt);
\fill[orange,opacity=.55] (7.5231,1.8392) circle (1pt);
\fill[orange,opacity=.55] (7.1148,1.4810) circle (1pt);
\fill[orange,opacity=.55] (7.1148,1.5394) circle (1pt);
\fill[orange,opacity=.55] (7.1148,1.6294) circle (1pt);
\fill[orange,opacity=.55] (7.5231,1.9892) circle (1pt);
\fill[orange,opacity=.55] (7.0130,1.3895) circle (1pt);
\fill[orange,opacity=.55] (7.1148,1.5394) circle (1pt);
\fill[orange,opacity=.55] (7.0130,1.5394) circle (1pt);
\fill[orange,opacity=.55] (7.5562,2.1690) circle (1pt);
\fill[orange,opacity=.55] (7.1556,1.5606) circle (1pt);
\fill[orange,opacity=.55] (7.2981,1.5574) circle (1pt);
\fill[orange,opacity=.55] (7.4407,1.6833) circle (1pt);
\fill[orange,opacity=.55] (7.0130,1.5574) circle (1pt);
\fill[orange,opacity=.55] (7.4407,1.6833) circle (1pt);
\fill[orange,opacity=.55] (7.4407,1.5574) circle (1pt);
\fill[orange,opacity=.55] (7.4407,1.5574) circle (1pt);
\fill[orange,opacity=.55] (7.1556,1.5574) circle (1pt);
\fill[orange,opacity=.55] (7.2981,1.6833) circle (1pt);
\fill[orange,opacity=.55] (7.2981,1.5574) circle (1pt);
\fill[orange,opacity=.55] (7.2981,1.5574) circle (1pt);
\fill[orange,opacity=.55] (7.1556,1.5574) circle (1pt);
\fill[orange,opacity=.55] (7.1370,1.6176) circle (1pt);
\fill[orange,opacity=.55] (7.0440,1.6176) circle (1pt);
\fill[orange,opacity=.55] (7.9208,2.1925) circle (1pt);
\fill[orange,opacity=.55] (7.0440,1.6176) circle (1pt);
\fill[orange,opacity=.55] (7.0440,1.6176) circle (1pt);
\fill[orange,opacity=.55] (7.0463,1.5355) circle (1pt);
\fill[orange,opacity=.55] (7.5886,2.3333) circle (1pt);
\fill[orange,opacity=.55] (6.9510,1.2891) circle (1pt);
\fill[orange,opacity=.55] (7.1370,1.5355) circle (1pt);
\fill[orange,opacity=.55] (7.0463,1.3713) circle (1pt);
\fill[orange,opacity=.55] (7.9872,2.0884) circle (1pt);
\fill[orange,opacity=.55] (7.0130,1.4315) circle (1pt);
\fill[orange,opacity=.55] (7.0130,1.5574) circle (1pt);
\fill[orange,opacity=.55] (6.3000,1.1796) circle (1pt);
\fill[orange,opacity=.55] (7.5833,1.9352) circle (1pt);
\fill[orange,opacity=.55] (7.0130,1.3056) circle (1pt);
\fill[orange,opacity=.55] (7.0130,1.4315) circle (1pt);
\fill[orange,opacity=.55] (6.7278,1.1796) circle (1pt);
\fill[orange,opacity=.55] (7.4443,1.6833) circle (1pt);
\fill[orange,opacity=.55] (6.3000,1.0537) circle (1pt);
\fill[orange,opacity=.55] (6.7278,1.4315) circle (1pt);
\fill[orange,opacity=.55] (6.3000,0.9309) circle (1pt);
\fill[orange,opacity=.55] (7.5833,1.9352) circle (1pt);
\fill[blue,opacity=.55] (7.3949,1.7126) circle (1pt);
\fill[green!60!black,opacity=.55] (6.7583,1.7643) circle (1pt);
\fill[green!60!black,opacity=.55] (7.1233,1.5701) circle (1pt);
\fill[green!60!black,opacity=.55] (6.7583,1.4944) circle (1pt);
\fill[magenta,opacity=.55] (7.4800,1.5769) circle (1pt);
\fill[magenta,opacity=.55] (7.4289,1.5994) circle (1pt);
\fill[magenta,opacity=.55] (7.3694,1.6221) circle (1pt);
\fill[magenta,opacity=.55] (7.2421,1.4954) circle (1pt);
\fill[orange,opacity=.55] (6.9705,1.7868) circle (1pt);
\fill[orange,opacity=.55] (6.9366,1.7568) circle (1pt);
\fill[orange,opacity=.55] (6.9366,1.9142) circle (1pt);
\fill[orange,opacity=.55] (6.9535,1.6444) circle (1pt);
\fill[orange,opacity=.55] (6.9366,1.8471) circle (1pt);
\fill[orange,opacity=.55] (6.8772,1.7568) circle (1pt);
\fill[orange,opacity=.55] (6.8772,1.8617) circle (1pt);
\fill[orange,opacity=.55] (7.1742,1.4502) circle (1pt);
\fill[blue,opacity=.55] (7.4452,1.5862) circle (1pt);
\fill[green!60!black,opacity=.55] (6.9111,1.3843) circle (1pt);
\fill[green!60!black,opacity=.55] (7.0375,1.5270) circle (1pt);
\fill[green!60!black,opacity=.55] (6.6056,1.2718) circle (1pt);
\fill[magenta,opacity=.55] (7.4274,1.5704) circle (1pt);
\fill[magenta,opacity=.55] (7.0200,1.4247) circle (1pt);
\fill[magenta,opacity=.55] (7.2132,1.4216) circle (1pt);
\fill[magenta,opacity=.55] (7.1956,1.4782) circle (1pt);
\fill[orange,opacity=.55] (7.1218,1.3660) circle (1pt);
\fill[orange,opacity=.55] (7.0308,1.5022) circle (1pt);
\fill[orange,opacity=.55] (7.0481,1.5022) circle (1pt);
\fill[orange,opacity=.55] (6.8163,1.6976) circle (1pt);
\fill[orange,opacity=.55] (7.1223,1.5270) circle (1pt);
\fill[orange,opacity=.55] (7.0036,1.5673) circle (1pt);
\fill[orange,opacity=.55] (6.9568,1.5673) circle (1pt);
\fill[orange,opacity=.55] (6.9638,1.6980) circle (1pt);
\fill[blue,opacity=.55] (7.5394,1.6445) circle (1pt);
\fill[green!60!black,opacity=.55] (7.3694,1.6593) circle (1pt);
\fill[green!60!black,opacity=.55] (7.2082,1.8175) circle (1pt);
\fill[green!60!black,opacity=.55] (7.0639,1.4944) circle (1pt);
\fill[magenta,opacity=.55] (7.4634,1.5994) circle (1pt);
\fill[magenta,opacity=.55] (7.4883,1.6824) circle (1pt);
\fill[magenta,opacity=.55] (7.4543,1.6219) circle (1pt);
\fill[magenta,opacity=.55] (7.2676,1.7718) circle (1pt);
\fill[orange,opacity=.55] (7.2082,1.5469) circle (1pt);
\fill[orange,opacity=.55] (7.1233,1.5994) circle (1pt);
\fill[orange,opacity=.55] (7.1488,1.7568) circle (1pt);
\fill[orange,opacity=.55] (7.0724,1.3370) circle (1pt);
\fill[blue,opacity=.55] (7.1113,1.5022) circle (1pt);
\fill[green!60!black,opacity=.55] (6.6056,1.0059) circle (1pt);
\fill[green!60!black,opacity=.55] (6.9638,1.7224) circle (1pt);
\fill[green!60!black,opacity=.55] (6.3000,0.8198) circle (1pt);
\fill[magenta,opacity=.55] (6.6688,1.0059) circle (1pt);
\fill[magenta,opacity=.55] (6.6688,0.8212) circle (1pt);
\fill[magenta,opacity=.55] (7.0375,1.5270) circle (1pt);
\fill[magenta,opacity=.55] (7.0375,2.0481) circle (1pt);
\fill[orange,opacity=.55] (6.8444,1.5921) circle (1pt);
\fill[orange,opacity=.55] (6.6056,1.3967) circle (1pt);
\fill[orange,opacity=.55] (6.6056,1.5921) circle (1pt);
\fill[orange,opacity=.55] (7.1113,1.7224) circle (1pt);
\fill[blue,opacity=.55] (7.4874,1.5917) circle (1pt);
\fill[green!60!black,opacity=.55] (7.1054,1.7306) circle (1pt);
\fill[green!60!black,opacity=.55] (7.2358,1.5569) circle (1pt);
\fill[green!60!black,opacity=.55] (6.3000,1.6125) circle (1pt);
\fill[magenta,opacity=.55] (7.3694,1.5444) circle (1pt);
\fill[magenta,opacity=.55] (7.3694,1.4944) circle (1pt);
\fill[magenta,opacity=.55] (7.4245,1.5986) circle (1pt);
\fill[magenta,opacity=.55] (7.1573,1.5569) circle (1pt);
\fill[orange,opacity=.55] (7.0077,1.5432) circle (1pt);
\fill[orange,opacity=.55] (7.0864,1.5500) circle (1pt);
\fill[orange,opacity=.55] (7.0942,1.7167) circle (1pt);
\fill[orange,opacity=.55] (7.0710,1.4043) circle (1pt);
\fill[blue,opacity=.55] (7.4333,1.5710) circle (1pt);
\fill[green!60!black,opacity=.55] (6.9684,1.9667) circle (1pt);
\fill[green!60!black,opacity=.55] (7.1495,1.6125) circle (1pt);
\fill[green!60!black,opacity=.55] (6.9684,1.9667) circle (1pt);
\fill[magenta,opacity=.55] (7.5348,1.6406) circle (1pt);
\fill[magenta,opacity=.55] (7.4011,1.5780) circle (1pt);
\fill[magenta,opacity=.55] (7.2043,1.5160) circle (1pt);
\fill[magenta,opacity=.55] (7.1964,1.5641) circle (1pt);
\fill[orange,opacity=.55] (7.0236,2.0292) circle (1pt);
\fill[orange,opacity=.55] (6.9684,1.9667) circle (1pt);
\fill[orange,opacity=.55] (7.1021,2.0571) circle (1pt);
\fill[orange,opacity=.55] (7.1886,1.3490) circle (1pt);
\node[anchor=north] at (6.300,.5) {0};
\node[anchor=east] at (6.23,0.550) {0};
\node[anchor=north] at (7.583,.5) {0.3};
\node[anchor=east] at (6.23,1.683) {0.3};
\node[anchor=north] at (8.867,.5) {0.6};
\node[anchor=east] at (6.23,2.817) {0.6};
\node[anchor=north] at (10.150,.5) {0.9};
\node[anchor=east] at (6.23,3.950) {0.9};
\node[anchor=west,font=\footnotesize\bfseries] at (11.1,4.4) {(c) RGB sensitivity};
\draw[->] (11.6,.55)--(16.9,.55);
\draw[->] (11.6,.55)--(11.6,4.05);
\draw[dashed,gray] (11.6,2.2)--(16.65,2.2);
\node[rotate=90] at(11.2,2.3){Failure AUROC};
\node at (12.00,.3) {Original};
\node at (13.50,.3) {Low};
\node at (15.00,.3) {Medium};
\node at (16.50,.3) {High};
\node[anchor=east] at (11.55,0.550) {0};
\node[anchor=east] at (11.55,2.200) {0.5};
\node[anchor=east] at (11.55,3.850) {1};
\fill[blue] (12.000,3.652) circle (1.5pt);
\fill[blue] (13.500,3.652) circle (1.5pt);
\fill[blue] (15.000,3.553) circle (1.5pt);
\fill[blue] (16.500,3.091) circle (1.5pt);
\draw[blue,line width=.65pt] (12.000,3.652)--(13.500,3.652)--(15.000,3.553)--(16.500,3.091);
\fill[green!60!black] (12.000,3.652) circle (1.5pt);
\fill[green!60!black] (13.500,3.487) circle (1.5pt);
\fill[green!60!black] (15.000,2.068) circle (1.5pt);
\fill[green!60!black] (16.500,2.101) circle (1.5pt);
\draw[green!60!black,line width=.65pt] (12.000,3.652)--(13.500,3.487)--(15.000,2.068)--(16.500,2.101);
\fill[magenta] (12.000,3.652) circle (1.5pt);
\fill[magenta] (13.500,2.662) circle (1.5pt);
\fill[magenta] (15.000,2.167) circle (1.5pt);
\fill[magenta] (16.500,1.639) circle (1.5pt);
\draw[magenta,line width=.65pt] (12.000,3.652)--(13.500,2.662)--(15.000,2.167)--(16.500,1.639);
\node[text=blue,anchor=west] at(12.0,1.25){Dark};
\node[text=green!60!black,anchor=west] at(12.0,.98){Blur};
\node[text=magenta,anchor=west] at(12.0,.71){Occlusion};
\end{tikzpicture}}

%% file: sae_representation_extension_20261111.tex
\section{Representation benefit boundary (exploratory)}
\label{app:sae-representation}
This separate study asks whether sparse autoencoder (SAE) features improve prediction, beyond reconstructing or perturbing a video representation. It does not evaluate robot failure detection or neural correspondence.

\paragraph{Data and frozen representation.}
Egocentric monochrome Assembly101~\cite{sener2022assembly101} clips are passed through frozen V-JEPA~2~\cite{assran2025vjepa2}. Twelve training-pool participants provide two sessions each; eight fit the SAEs and four supply repeatedly inspected internal validation. For 144 causal clips, 64 frames span the previous 1.96875 seconds. Four spatial positions in each of 32 tubelets from block 12 give 128 sampled 1024-dimensional vectors per clip (12,288 fitting and 6,144 validation vectors). Features attend across the observed clip and become available at its endpoint, not at earlier tubelet times.

Our sparse-bottleneck pilot~\cite{gao2025scalingsae} uses 2048 latents, rectification/top-32 selection, unit-norm decoder columns, 600 optimizer steps, batch 256 and learning rate .001. Global training-centering/root-mean-square (RMS) scaling and per-vector normalization each have seeds 11/12/13. Checkpoints are not selected on validation. Exact dataset/model revisions and fixed positions accompany the numerical artifact.

\paragraph{Relevant features do not imply benefit.}
Compared with exact rank-32 principal component analysis (PCA), raw-coordinate reconstruction error is about 24.5--25.3\% lower; this matches active coordinates, not parameter count. Individual cross-seed matches are weak (median decoder cosines .37--.39). Training-energy groups share structure, but no semantic concepts are established. On eight separate 55-second validation clips, removing each dictionary's top-32 decoded contributions while preserving reconstruction residual changes encoder outputs more than magnitude-matched decoder-direction controls. A post-hoc common-support spatial-shift comparison has no zero-direction fallback. Across 448 cached encoder conditions, all remain representation probes, not new robot rollouts.

Fixed prediction heads trained on 192 pairs from the eight fitting participants predict the original same-view encoder vector two seconds later. Group removal increases mean error by 7.9--16.1\% for a residual head and .5--2.7\% for a direct head. Spatial ordering effects depend on the head; the residual head is worse than persistence on the eight intervention clips. Neither ablation loss nor reconstruction quality demonstrates that SAE features improve an unmodified predictor.

\paragraph{Original-input comparison.}
The benefit test uses 96 fitting and 48 validation clips from the 144-clip set. Session/time/frame hashes match the two-second future cache. Every visual input uses input-only PCA, standardized scores and direct ridge prediction; dimensions $\{8,32\}$ and penalty bases $\{1,10,100\}$ are selected by eight fitting-participant folds. SAEs and the upstream normalizer already saw those eight participants, so this is conditional head cross-validation, not end-to-end nested representation training. No final-validation participant fits an SAE, but that cohort was inspected previously. Both family choices and all final heads are saved before validation scoring.

\begin{table}[ht]
\centering\small
\setlength{\tabcolsep}{6pt}
\begin{tabular}{llrrr}
\toprule
Input representation & Normalization/seed & PCA dimension & Penalty base & Future error\\
\midrule
Final pooled encoder & --- & 8 & 10 & 0.048301 \\
Raw token mean & Global RMS & 32 & 100 & 0.058815 \\
Raw token mean & Per-vector RMS & 32 & 100 & 0.058703 \\
SAE latent mean & Global, seed 11 & 32 & 10 & 0.058891 \\
SAE latent mean & Global, seed 12 & 32 & 10 & 0.058528 \\
SAE latent mean & Global, seed 13 & 32 & 10 & 0.058510 \\
SAE latent mean & Per-vector, seed 11 & 32 & 10 & 0.059171 \\
SAE latent mean & Per-vector, seed 12 & 32 & 10 & 0.058197 \\
SAE latent mean & Per-vector, seed 13 & 32 & 10 & 0.058542 \\
Clock only & --- & --- & 100 & 0.063723 \\
Persistence & --- & --- & --- & 0.051363 \\
Training-target mean & --- & --- & --- & 0.064350 \\
\bottomrule
\end{tabular}
\caption{Original-input two-second representation prediction (exploratory). Actor-equal mean squared L2 error to the original unit-normalized future encoder vector; lower is better. Each row uses its training-head-selected parameters. The final penalty is the base multiplied by $96/72$. Predictions are not renormalized. All six SAE variants are retained; no validation-best seed is selected.}
\label{tab:sae-benefit}
\end{table}

Training-head selection chooses per-vector SAE seed 11 (.059171), versus final pooled features (.048301): 22.5\% higher error, and .8\% worse than the corresponding raw-token mean (.058703). All seeds are retained. Final pooled features use a later layer/all patches; raw and SAE means use the same 128 positions. Thus the overall baseline gap cannot be attributed only to the SAE transform. Position-aware pooling is not tested.

Four tests and a separate closed-form implementation reproduce 456 head-fold fits, 57 candidate score vectors, ten final heads and all validation predictions. Saved-output checks do not independently rerun every encoder pass. With four reused validation participants, no population inference or new failure-detection, policy, safety or neural-mechanism gain is claimed. Cross-view error labels remain excluded pending verified time mapping; full exploratory records are retained in the numerical artifact.

%% file: evidence_inventory_round2.tex
\clearpage
\section{Evidence units and uncertainty ledger}
\label{app:evidence-ledger}
The 25 released sets count five FIPER tasks, seventeen Sentinel release settings and three aggregate SAFE releases. They are not 25 independent tasks, cohorts or detector-score streams. The release ledger below preserves those different counting levels. No total of unique physical episodes across releases is asserted.

\begin{table}[ht]
\centering\small
\setlength{\tabcolsep}{5pt}
\begin{tabular}{p{.18\textwidth}p{.18\textwidth}p{.20\textwidth}p{.34\textwidth}}
\toprule
Evidence & Counting unit & Available episode count & Score/calibration boundary\\
\midrule
FIPER & 5 tasks & 1540 task-listed TEST episodes & Six first-alarm score exports; continuous-score AUROC unavailable. Native calibration/setting reproduction retained.\\
Sentinel & 17 release settings & 700 raw release-listed episodes & Score files map to 13 settings; variants can reuse episodes and calibration. Four settings have no detector-score files.\\
SAFE LIBERO & 1 release, 10 tasks & 500 episodes & Four training-free token metrics; learned SAFE detectors not rerun.\\
SAFE WidowX & 1 release, 8 tasks & 532 episodes & Fixed within-task duration; no learned SAFE rerun or score-calibration split used by our adapter.\\
SAFE Franka & 1 release, 13 tasks & 1464 episodes & Training-free metrics; full-length restriction and task mixture require separate accounting.\\
GR00T replication & 4 tasks, new collection & 816 episodes; 815 with logged secondary scores & Primary termination endpoint separate from exploratory landmark/proxy analyses.\\
GR00T frozen heads & 4 tasks, separate prospective collection & 416 episodes; 342 quarter-horizon risk-set episodes & DEV-only fitted heads; all-task primary risk-set gate fails. Nine heads share this cohort, not nine independent datasets (Supp.~S).\\
\bottomrule
\end{tabular}
\caption{Counts at their recorded level, not additive independent observations. FIPER/Sentinel PushChair may overlap. Calibration rows and score variants must not be added to TEST or raw-release counts. The per-setting availability/risk-set tables in Supp.~H,K remain authoritative for scoreable subsets.}
\end{table}

\paragraph{Resampling and scope.}
Original FIPER contrasts use outcome-stratified episode resampling with score training, calibration and selected settings held fixed. Ranking checks share episode draws across methods before restriction to each risk set; repeated variants and seeds are dependent. Sentinel CAL+TEST resampling in Supp.~N adds threshold uncertainty only for score streams whose released predictions can be reconstructed. It does not cover model training or setting selection. Historical GR00T replication intervals resample configuration groups; its landmark AUROC combines within-task comparisons weighted by success--failure pair counts, rather than an equal-task macro. The changing landmark populations and low success counts remain explicit. The new frozen-head study instead fixes an equal-task macro endpoint (Supp.~S).

The source inventory has 91 heterogeneous records: method exports, score files, task entries and length-only release tags. That record count is an interchange/availability count, not a new dataset count. Missing class support, missing scores, overlap uncertainty and absent end-to-end uncertainty remain reported limitations.

\paragraph{Computational stages.}
The new PushChair timing reanalysis is CPU-only NumPy on archived numeric arrays, with no model calls. This excludes the original cost of producing detector scores. FIPER seeds run in separate processes to limit memory use. Policy-rollout generation, visual-score fitting and released-output auditing have separate computational costs; an offline audit is not a policy-training benchmark. The paced replay in Supp.~X records PyTorch 2.11.0+cu130/CUDA and excludes actor inference and physical acquisition. Historical total GPU-hours are not reconstructed or presented as a matched training budget.

%% file: prospective_monitor_20261112.tex
\clearpage
\section{Prospective DEV-trained frozen outcome monitors}
\label{app:prospective-monitors}
\paragraph{Status and primary outcome.}
All 416 planned episodes were collected after locally freezing weights, references, thresholds, code and endpoint. This was not external registration; historical DEV/confirmation results had been inspected. The primary is \textbf{non-evaluable under its prespecified risk-set rule}: quarter-horizon Eggplant has four successes, below five outcomes per class in every task. Estimates are descriptive, confirming neither visual benefit nor its absence. No task, episode count or primary endpoint was changed. The later TEST-2 is a separate study (Supp.~\ref{sec:prospective-test2}); it does not repair or replace this failed gate, and its episodes are not pooled with TEST-1.

\paragraph{Question and population.}
The target is failure to achieve the task's recorded success condition by a fixed horizon, not physical-failure onset. The actor and simulator follow Supp.~O: the same frozen GR00T checkpoint and four ManiSkill Bridge tasks. New task-specific episode seeds start at 500000, distinct from earlier DEV/confirmation namespaces. Fixed counts are Carrot 64, Spoon 64, StackCube 160 and Eggplant 128. The count rule was set from prior DEV risk-set proportions, not a formal power guarantee for a .02 AUROC effect. These are fresh stochastic episodes in existing task/configuration families, not unseen-task or unseen-configuration generalization. Full horizons are 60/120 actions; all 416 episodes are retained. In total 155 achieve success and 261 do not; 74 successes occur by the quarter landmark and are excluded there, leaving 342 risk-set episodes (81 successes, 261 failures).

\paragraph{DEV partitions and fitting.}
The existing 640 DEV episodes are split by whole configuration groups within task. Groups are ordered by SHA256 of \texttt{prospective-v1:task:config}; the first $\lceil .2G\rceil$ groups form a reference bank (REF), the next $\lceil .2G\rceil$ calibration (CAL), and the rest fitting (FIT). Labels do not determine the final group order. An earlier modulo-hash allocation was inspected on DEV and replaced by fixed group counts before collection. FIT/CAL/REF contain 377/131/132 full episodes. At the quarter landmark FIT contains 324 episodes (115 successes, 209 failures), and CAL 109 (34 successes, 75 failures). CAL success counts by Carrot/Spoon/StackCube/Eggplant are 10/21/2/1, so pooled calibration cannot guarantee task-wise FPR.

At each of quarter/half/three-quarter horizons separately, three StandardScaler--logistic-regression heads use identical FIT risk sets and labels: $C=1$, balanced class weights, \texttt{lbfgs} solver, maximum 3000 iterations, seed 0. There is no hyperparameter or representation search. The pooled CAL-success 95th percentile (\texttt{higher}) fixes each head's threshold; alarm means strictly above it. These are threshold-calibrated scores, not calibrated posterior probabilities. TEST labels never fit a reference, transform, head or threshold. The nine heads are not nine independent datasets.

\paragraph{Causal information boundary.}
The collection logs state and RGB before action index $j$, and success flags after that action. At decision $L$ completed actions, state/RGB index $L$ and previous executed actions $[0,L)$ are available; any success in those $L$ completed actions removes the episode from the risk set. No action $[L]$, object pose, contact force, current post-action flag, terminal duration or configuration ID is an input.

Metadata has 6 features: task one-hot, $L/T$, $\log(1+L)$. State/action adds current end-effector xyz/gripper, sine/cosine of Euler angles, joint positions, xyz/gripper and joint displacements from initial state, last-three mean absolute finite differences, previous action and prefix action mean/standard deviation (69 total). Differences are per control step, not measured SI velocities. The states are simulator-computed robot kinematics/proprioception, not measured hardware telemetry. Actions already depend on the vision-using actor, so this baseline can carry vision-mediated information.

Visual adds one image-novelty feature (70 total): mean Euclidean distance from the current frozen ResNet-18 ImageNet1K\_V1 embedding to five nearest REF-success frames. Reference frames end at the earlier of the quarter horizon and first success; query episodes never enter REF. Logged $128\times128$ RGB is bilinearly resized to $224\times224$, normalized by ImageNet statistics and independently encoded in evaluation mode; saved 512-dimensional embeddings are float16. REF-success frame counts by Carrot/Spoon/StackCube/Eggplant are 224/384/64/565. All heads share FIT/CAL labels, but success labels additionally construct the visual reference bank. This is a comparison of specified pipelines, not a pure causal effect of image access under identical reference supervision.

\begin{table}[ht]
\centering\small
\begin{tabular}{llrrrr}
\toprule
Landmark & Task & Risk-set S/F & Task/time & + state/actions & + image novelty\\
\midrule
0.25 & Carrot & 27/37 & 0.500 & 0.793 & 0.801\\
0.25 & Spoon & 42/16 & 0.500 & 0.688 & 0.707\\
0.25 & StackCube & 8/152 & 0.500 & 0.741 & 0.757\\
0.25 & Eggplant & 4/56 & 0.500$^\dagger$ & 0.638$^\dagger$ & 0.638$^\dagger$\\
0.5 & Carrot & 10/37 & 0.500 & 0.805 & 0.824\\
0.5 & Spoon & 9/16 & 0.500 & 0.451 & 0.451\\
0.5 & StackCube & 4/152 & 0.500$^\dagger$ & 0.816$^\dagger$ & 0.816$^\dagger$\\
0.5 & Eggplant & 0/56 & --- & --- & ---\\
0.75 & Carrot & 3/37 & 0.500$^\dagger$ & 0.964$^\dagger$ & 0.964$^\dagger$\\
0.75 & Spoon & 6/16 & 0.500 & 0.667 & 0.667\\
0.75 & StackCube & 2/152 & 0.500$^\dagger$ & 0.990$^\dagger$ & 0.990$^\dagger$\\
0.75 & Eggplant & 0/56 & --- & --- & ---\\
\bottomrule
\end{tabular}
\caption{Per-task diagnostic AUROC from frozen heads on fresh episodes. $\dagger$: fewer than five outcomes of one class; not a valid benchmark cell under the planned gate. Dashes: AUROC mathematically undefined (no successes). Values in small cells are retained for transparency, not substituted for valid primary evidence.}
\label{tab:prospective-frozen}
\end{table}

\paragraph{Primary contrast and uncertainty.}
The planned primary is equal-task macro AUROC(visual)$-$AUROC(state/action) at the quarter horizon. With the gate failed, descriptive macro AUROC is .500/.715/.726 for metadata/state/visual. The descriptive visual difference is $+.01095$ with percentile 95\% interval $[-.00495,.02817]$. The point misses the .02 practical target; the interval contains both zero and .02, so neither practical benefit nor its absence is established. State minus metadata is descriptively $+.21491$ $[.08259,.32280]$, not a replacement primary confirmation.

We draw 2000 paired configuration-group bootstrap samples independently within tasks, using the same group multiplicities for every arm, RNG 20261111. A draw lacking either class in any task has undefined macro AUROC: 1967/2000 quarter draws are valid (33 excluded), exceeding the planned 1900 minimum. This does not cure the separate five-per-class gate failure. At later horizons Eggplant has zero successes, so all-four-task macro AUROC and contrasts are undefined; no three-task average is substituted. Training, references and calibration remain fixed; these intervals do not include refitting or task-acquisition uncertainty. No multiplicity-controlled confirmation is claimed for secondary quantities.

\begin{figure}[ht]
\centering
\includegraphics[width=.96\textwidth]{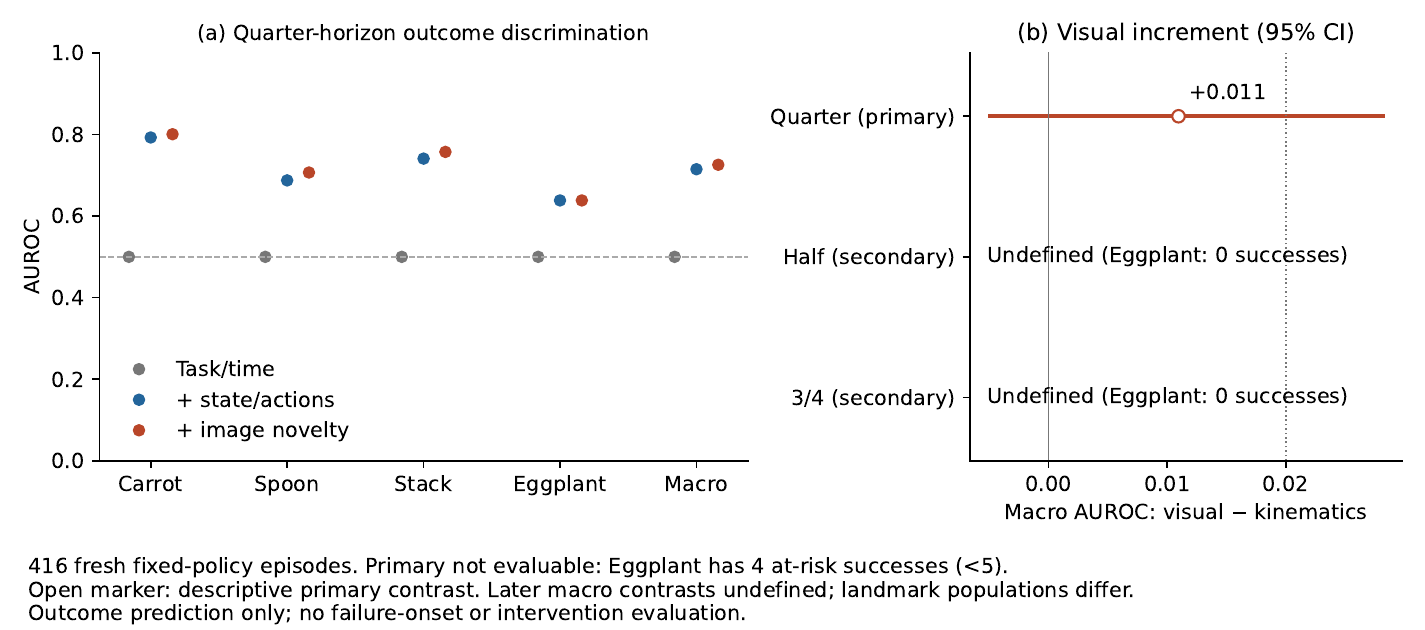}
\caption{Frozen-head prospective test, descriptive because the primary risk-set gate fails. Left: quarter-horizon per-task and macro point estimates, including the four-success Eggplant diagnostic. Right: planned paired contrast; open marker indicates ineligible confirmation. The dotted line marks the .02 practical target, not a significance cutoff. Later macro contrasts are undefined rather than interpolated; populations change across landmarks.}
\end{figure}

\paragraph{Checks and deviations.}
The local freeze bound 267 files before new collection (record digest prefix \texttt{4018565bd2f9}). All passed verification before collection and final evaluation. Six timing/partition tests passed. Independent DEV refitting reproduced all nine heads and thresholds exactly. Replacing four future frames left the actual encoder's first 16 outputs unchanged in a CPU DEV check. Independent failure--success pair counting with configuration-multiplicity quadratic forms reproduces every primary AUROC exactly and the confidence limits within $2\times10^{-17}$. All 416 horizons, task-specific episode IDs, seeds, finite state/action arrays and actor snapshot were checked.

Initial fitting stopped at an ID assertion before saving weights: historical embeddings had a documented $-200000$ ID conversion. DEV features were regenerated from raw frames. Strictly offline actor initialization failed at a tokenizer metadata check; final startup permitted that check while keeping transformer weight loading local and the original actor snapshot fixed. After all collection, socket cleanup redundantly tried to remove a socket already removed by Listener.close; the server had acknowledged shutdown, and no rollout was interrupted, replaced or regenerated. These deviations do not change the endpoint or fitted models.

\paragraph{Failure-onset boundary.}
A separate DEV-only feasibility check found grasp-loss proxies in successful as well as unsuccessful episodes, including 3/1/2/8 Carrot/Spoon/StackCube/Eggplant successful episodes before their first success. Grasp loss can recover or reflect placement; timeout is not verified physical-failure onset. No human onset annotations or reliable physical-onset truth were collected. Consequently this study supports no fixed-lead-time, real-time alarm-delivery, intervention or hardware-safety claim. Annotation and genuinely new-policy/scene validation remain separate work.

%% file: constructive_design_principles.tex
\clearpage
\section{Prospective design principles for attributable visual warnings}
\label{app:design-principles}
\paragraph{Status and scope.}
This section proposes designs and falsifiable tests, not a new detector or a validated causal-disentanglement method. Conditional predictive improvement, sensitivity to a sensor intervention, and improved robot outcomes are different claims. Learning a predictive representation does not by itself identify its causal variables~\cite{scholkopf2021causal}. The main-paper audit establishes neither a specific architecture's efficacy nor intervention utility.

\paragraph{1. Specify the endpoint before removing a shortcut.}
Fix what counts as success, failure onset, recoverable error and timeout. A geometric success flag, a released-and-settled placement, and an irreversible physical failure need not coincide. Changing the endpoint requires rebuilding its risk sets and calibration labels; retaining the old success-conditioned population can exclude the new endpoint's positive examples. Fix the landmark, horizon and class/configuration support rules on development data before collecting TEST. Record actual terminal states rather than auto-reset observations. Under a finite observation cap, state whether timeout is a negative, a task non-completion or an unobserved event; do not silently mix these targets. Exact landmarking removes variation in elapsed time within a task, but does not identify the mechanism of a remaining visual association.

\paragraph{2. Model the expected component explicitly.}
Let $Z_L$ contain task, elapsed time, known policy identity and the robot state/action history actually available at decision $L$. Let $V_{\le L}$ be the available visual prefix. Declare embedded timestamps, clip lengths, sampling schedules and audio access; do not give the visual branch undeclared timing channels. On the same task-wise landmark population $R_L$, compare a contextual baseline $p_0(Z_L)$ with a visual update. One candidate is a shrinkage-regularized log-odds correction,
\begin{equation}
 \operatorname{logit}p_1
 =\operatorname{logit}p_0(Z_L)+g_\theta(V_{\le L},Z_L).
 \label{eq:prospective-visual-update}
\end{equation}
Use out-of-fold baseline predictions when fitting the correction; fit preprocessing and reference banks without query-episode leakage. Include a capacity-matched context-only correction, so additional parameters or reference supervision are not credited to vision. Freeze all fitting, selection and probability-calibration procedures before TEST. Threshold calibration alone does not turn a class-balanced score into a posterior probability; the existing monitor results are not reinterpreted as such.

Keep one prespecified primary metric and treat additional losses as secondary. Report paired held-out AUROC differences and a probability-score contrast, such as negative logarithmic loss~\cite{gneiting2007proper},
\begin{equation}
 \Delta_{\rm pred}(L)
 =\mathrm E\!\left[\ell(Y,p_0)-\ell(Y,p_1)\mid R_L\right],
 \quad \ell(y,p)=-y\log p-(1-y)\log(1-p).
 \label{eq:prospective-predictive-increment}
\end{equation}
Use probabilities in $(0,1)$ and predeclare any numerical clipping. The target $Y$ and prediction horizon must be identical in both terms. Positive $\Delta_{\rm pred}$ supports an increment over the specified fitted comparator, not a causal contribution of vision or an additive decomposition of AUROC. Actions from a vision-using actor can already carry visual information; conditioning on them tests an additional detector channel, not the total effect of the robot's visual system. Residual training or adversarially suppressing time can remove useful task-phase information and supplies no causal guarantee. These are candidates to compare, not requirements that visual features contain no time information.

\paragraph{Information versus model capacity.}
Let $P_L=P(\cdot\mid R_L)$ be the risk-set distribution. At its ideal population predictors, the log-score gain has a precise interpretation:
\begin{equation}
\mathrm E_{P_L}\!\left[\log\frac{p_L(Y\mid Z_L,V_{\le L})}{p_L(Y\mid Z_L)}\right]
= I_{P_L}(Y;V_{\le L}\mid Z_L).
\end{equation}
This standard conditional-information identity motivates the comparison; finite fitted heads and their AUROC differences do not estimate that quantity automatically. Conversely, if a physical residual $r=f(Z_L)$ is deterministic given the full available state/history and declared nominal-model parameters already in $Z_L$, then $I_{P_L}(Y;r\mid Z_L)=0$. The residual can still help a restricted model through a useful representation. Matched-capacity context controls distinguish such computational advantages from an extra observation source. Neither conditional prediction nor conditioning on the still-running population identifies a causal effect on task outcomes.

\paragraph{Memory as a context control (prospective).}
Multi-Scale Embodied Memory (MEM) combines recent video with long-horizon language memory for action policies~\cite{torne2026mem}. This motivates separating short visual dynamics from compressed task history in a future monitor, not importing MEM's policy results as warning evidence. A memory of completed steps can also reveal progress; compare against a strong history-conditioned reference under equal input windows and supervision. Visually derived memory remains a visual-history input. Retain source pointers, observation/readiness times and uncertainty; a generated summary is not a new direct observation. Neither an eventual outcome nor a later-reviewed event may enter a summary before it becomes available. No MEM replication or memory-based warning gain is established here.

\paragraph{Performance conditions and joint ablations (prospective).}
In $\pi_{0.7}$, \texttt{Speed} bins episode length in 500-step intervals, \texttt{Quality} scores execution from 1 to 5, and \texttt{Mistake} marks human-annotated action segments~\cite{pi2026pi07}. These training annotations are legitimate supervision; desired inference conditions are not measurements of the current episode's eventual outcome. Our decision-time boundary forbids supplying a retrospective evaluation label as an online input, not learning from past labeled trajectories. Possible duration-related structure in a representation motivates a controlled monitor comparison, not a finding that $\pi_{0.7}$ or SAFE leaks outcomes. An input \texttt{Mistake} flag also supplies no calibrated error posterior by itself.

The joint no-metadata ablation lowers performance~\cite{pi2026pi07}, but neither isolates \texttt{Speed} nor measures duration decoding or monitor reliance. The no-eval-data variant instead excludes autonomous evaluation episodes from training, a separate data-source ablation.

\paragraph{Throughput and a conditional duration implication.}
Raw throughput in $\pi_{0.7}$ is successes per hour; Fig.~6 normalizes it to the specialist and Fig.~7 to $\pi_{0.7}$~\cite{pi2026pi07}. It combines success with wall-clock efficiency, rather than measuring successful-rollout length alone. For the main paper's any-step, no-patience clock ceiling, let primes denote a changed policy and retain the same task mixture, outcome definition and recorded-step unit. If
\begin{equation}
 F'_S(\tau)\ge F_S(\tau)\quad\text{for all }\tau,
 \qquad F'_F(\tau)=F_F(\tau),
 \label{eq:efficiency-duration-condition}
\end{equation}
then $D'^+=\max_\tau[F'_S(\tau)-F_F(\tau)]\ge D^+$, so the best any-step clock balanced accuracy cannot decrease. The proof is pointwise comparison followed by maximization. This is a conditional implication of Proposition~1, not an empirical result about $\pi_{0.7}$. It need not be strict: both ceilings already equal one when all successes end before a common failure cap. It also does not guarantee improvement of a clock threshold chosen by a particular calibration rule. Higher throughput can reflect higher success probability, lower overhead or a changed control rate without fewer recorded steps. Check success/failure length distributions, caps and step/wall-clock units within each task and policy or speed setting; the aggregate throughput and joint metadata ablation do not verify these premises.

\paragraph{Generated goals remain predictions (prospective).}
The same policy can use generated subgoal images from a BAGEL-initialized model~\cite{pi2026pi07}. A generated condition becomes usable only when ready, while its source observation retains its earlier timestamp; it is not a new direct observation. A desired goal differs from an action-conditioned forecast. Goal-distance scores require progress/history controls, timing and error-specific validation; neither goal distance nor our successful-reference novelty proxy automatically measures prediction error. Any mistake head or goal-conditioned warning comparison is future work, not a tested $\pi_{0.7}$ detector here.

\paragraph{Forecast scores and arrived residuals (prospective).}
A WAM forecast sourced at $s$, ready at $r$, and targeting $u=s+k$ can support a forecast-based score only at a decision $d\ge r$. An error against the real image at $u$ becomes feedback later: its readiness is at least $\max\{r,\tau^{\rm avail}(u)\}$ plus comparison time. Preserve the source pointer and matched executed actions; a desired goal image is a different object. A forecast of failure and an arrived forecast residual therefore have different availability times and need separate calibration and endpoint definitions. Neither can be backdated to $s$.

For a future image-residual monitor, compare fixed-span predictions re-anchored to each newly available real observation with continuously rolled-out predictions. Re-anchoring may limit compounded drift; it does not establish independence from episode time. Forecast span $k$ and elapsed episode time are different variables, and a fixed-span receding forecast need not become less accurate as the episode ages. Keep horizon, image sampling and memory budget controlled, and retain A1/A2 plus history/context controls: cached history may still convey time or task progress. No universal reset interval, cache-growth property or WAM warning gain is asserted here. Dens's state-space residual and its eventual-outcome increment are already measured; useful image-forecast warning remains a separate test.

\paragraph{Physics-derived state comparators (prospective).}
Analytical Disturbance-Aware Policy Training (ADAPT) uses a momentum observer with nominal robot dynamics and motor commands to provide disturbance-related inputs for humanoid locomotion~\cite{lyu2026adapt}. This motivates an additional comparator, not an established EgoOddball result: compare task/time/state/action history, then add an estimated generalized-disturbance residual, then add local vision on the same episodes and endpoint. A residual computed from state history and nominal dynamics is a derived physical feature, not necessarily an independent information source. Do not feed simulator contact truth or an injected test perturbation into the estimated-residual branch. Verify joint ordering, actuation semantics, nominal-model agreement, initialization transients and causal filtering before this comparison. Residuals can contain ordinary support/contact forces and model mismatch; they are not failure labels. ADAPT reports finite observer response time, so score actual residual/alarm availability rather than backdating its response to disturbance onset. A fixed-base arm requires its own observer and calibration, including removal of the humanoid-specific floating-base weight adjustment. These proposed controls do not imply that ADAPT's G1 robustness results transfer to WidowX or a wheel-legged robot, or that physical residuals or vision already improve early warning here.

\paragraph{3. Separate sensory reliance from physical intervention.}
In an offline paired sensor test, hold the saved state/action trajectory, frame availability, landmark and outcome fixed, and alter only the detector's visual input. The actor still receives its original observations. Compare aligned evidence with a prespecified task-relevant occlusion and an equally processed irrelevant-region or rendering placebo. Retain identical episode pairs and report score/alarm changes at thresholds fixed on CAL. A drop under relevant occlusion supports reliance on those pixels under that intervention; it does not establish correct physical understanding. Frame shuffling, blank images or mismatched clips can create out-of-distribution inputs, so a large response alone is insufficient.

A separate world-intervention test can start from validated checkpoints and randomize a bounded perturbation versus sham, with its scheduled onset randomized within prespecified pre-intervention checkpoint strata and hidden from the detector. Freeze the policy and detector before assignment. Follow the resulting closed-loop trajectories rather than forcing their future actions or states to match: these may legitimately mediate the perturbation's effects. Report all assigned episodes, reset failures, early outcomes and conditional warning populations; restricting analysis to post-treatment survivors changes the estimand. This tests responses to a specified physical manipulation. Attributing task benefit to the monitor requires another comparison: randomize a prespecified monitor-triggered response versus a matched clock/state-triggered response or shadow mode, and measure task outcomes, contacts and intervention burden. Offline detection accuracy cannot substitute for that experiment.

\paragraph{4. Train for future events, and score time honestly.}
An onset study needs an operational event definition and independently reviewed onset intervals $[a,b]$. Let $d_L$ denote actual alarm-delivery time for landmark $L$, in the same time unit as $a,b$ and the required lead $\delta$. A conservative warning must satisfy $d_L+\delta\le a$; a timestamp before rollout termination is insufficient. Also preserve the trained forecast origin: if an observation at $t_L$ predicts an event within $h$ seconds, require $b\le t_L+h$. Delivery delay consumes lead and must not extend that forecast to $d_L+h$. Separate events already begun, intervals crossing the decision, and events with inadequate lead. Train only on prefixes available by the specified decision, keep trajectory-level splits, and exclude terminal flags or future-derived features from detector inputs. Use a loss for a stated future event window rather than treating the first visible error as its own early-warning target. A hazard or sequence model is optional; success as a competing event, observation caps and loss to follow-up still require explicit treatment. This proposed event-conditioned evaluation does not retroactively change the main paper's eventual-outcome landmark estimand.

Thresholds must be calibrated on separate episodes for the exact inference cadence and false-alarm endpoint. Fixed-landmark false-positive control is not continuous-stream alarm control: repeated decisions create additional alarm opportunities. A streaming study must instead calibrate an episode-level first-alarm rule, or another prespecified sequential criterion, and include processing and delivery latency in the lead-time calculation. Report recall at a matched false-alarm burden, not only AUROC or time-before-termination.

The timing geometry itself is a diagnostic: with constant source-to-delivery delay $\ell$, the source times that could satisfy both conservative conditions lie in $[b-h,a-\delta-\ell]$. Its width is $h-\delta-\ell-(b-a)$; broad review intervals or latency can eliminate a usable window. Report such episodes, cadence and onset-interval widths without removing them from the declared event denominator. This is a feasibility limit, not an achieved recall or a reason to backdate alarms.

\paragraph{5. Use expectancy as a specificity probe, not a safety claim.}
The proposed expected/unexpected $\times$ correct/incorrect probe asks whether a score responds to task error, perceptual rarity or violated expectation. Define expectation through a prespecified context or training distribution, independently validate correctness, and match event time, task phase and visible content where feasible. A context change must not silently redefine the task's correct action. Unexpected but correct actions provide a negative control for failure-specific credit; expected errors test whether an erroneous action is missed because it is unsurprising. This is a conceptual probe rather than evidence that every surprising action is unsafe, that every successful outcome was risk-free, or that the detector implements mismatch negativity. Generalization requires new policies/configurations and naturally occurring errors in addition to staged probes.

\paragraph{Costly review is another endpoint.}
Prediction loss, future-event warning and the value of requesting a review must be evaluated separately. A common maximum number of requests does not match realized review cost. For a new selective-review study, choose operating points on DEV/CAL and freeze them before TEST: compare capture at a stated realized-cost budget, or test a prespecified noninferiority margin for capture while reducing cost. Report the whole declared cost--capture profile, processing/service delay, normal-case requests, repeated requests and episodes that exhaust the budget before an event. Do not retune the operating point on TEST or reinterpret a failed primary result using a favorable secondary tradeoff. Event-supervised risk can be learned before a reviewer is available; learning whether deferral improves a decision additionally requires that reviewer's observed errors and costs. An ideal reviewer in replay supplies neither those data nor an intervention benefit. A future robot study should first log shadow requests under a frozen policy, then separately assess reviewer quality and a controlled response. Supplement~W retains a negative selective-review pilot under this distinction.

\paragraph{A falsifiable validation sequence.}
Begin with task/time, add causally available state/action controls, then add vision under matched reference supervision and search budget. Report continuous-score discrimination alongside thresholded accuracy and TWA, and task-wise as well as pooled results. Freeze the primary incremental metric, support rule, probability/threshold calibration and inference path. Test on new episodes without selecting the best model or landmark again. If the primary population lacks either outcome, report non-evaluability rather than appending episodes, dropping a task or reusing a selected exploratory subset. Only after that predictive test should a sensor-intervention study support a reliance claim, and a separately randomized response study support deployment utility. This sequence gives constructive targets while retaining the audit's central boundary: what an alarm predicts, what evidence changes it, and what acting on it improves must each be demonstrated.

%% file: systems_representation_pilot_20261113.tex
\clearpage
\section{Exploratory resource and representation diagnostics}
\label{app:resource-representation-pilot}
\paragraph{Evidence boundary.}
The following engineering and representation studies use the same 48 development (DEV) episodes, not new independent warning trials. Human event-onset review is still pending for this cohort. No result below establishes incremental failure-warning accuracy, delivered pre-onset lead, closed-loop task benefit or safety. The main-paper audit results and estimands are unchanged.

\paragraph{Resource-aware execution without changing the evidence set.}
BaGuaLu-2 addresses dynamic activation-memory provisioning for large-scale mixture-of-experts training~\cite{zhai2026bagualu2}. We borrow the systems principle of measuring resource pressure and mitigation cost, rather than importing its training framework or claiming to reproduce it. Our independently implemented offline feature runner records memory, complete request duration and recovery attempts. It can reduce its batch size on a typed out-of-memory error, preserves all frames in order, and fails explicitly if even one image cannot be processed. It does not truncate temporal windows, generate replacement features or alter timestamps. This behavior is verified with software fault injection; actual GPU exhaustion was not intentionally induced. The frozen online warning path is unchanged.

We profiled 144 RGB images sampled at frames 0, 60 and 120 of each DEV episode on an NVIDIA RTX 5080 Laptop GPU. Table~\ref{tab:offline-resource} reports medians over three whole-request repetitions and peak allocated CUDA memory; these repetitions do not estimate a stable tail distribution. File loading and weight initialization are recorded separately. No camera, network or robot-loop timing is included. Batch 32 increased offline throughput by approximately $3.73\times$ relative to batch 3, while increasing peak allocated memory. This is a feature-extraction speedup, not a robot control-rate improvement. Relative feature differences versus batch 3 reached approximately $7.9\!\times\!10^{-4}$ for global and $6.4\!\times\!10^{-4}$ for spatial features; none of the fixed source-presence decisions crossed its threshold on these 144 images. This does not guarantee unchanged warning scores or justify changing a frozen inference configuration without revalidation.

\begin{center}
\begin{minipage}{\linewidth}
\centering
\captionof{table}{Offline feature extraction on the same 144 DEV images. GPU denotes graphics processing unit; MiB denotes $2^{20}$ bytes. Peak allocation is process-local, not total device use.}
\label{tab:offline-resource}
\begin{tabular}{rrr}
\toprule
Maximum batch & Median images/s & Peak allocated memory (MiB)\\
\midrule
1 & 511.0 & 68.2\\
3 & 1021.7 & 72.2\\
8 & 2836.5 & 96.3\\
16 & 3606.1 & 151.9\\
32 & 3809.5 & 257.1\\
\bottomrule
\end{tabular}
\end{minipage}
\end{center}

\paragraph{Straightness is a candidate prior, not a failure criterion.}
Temporal Straightening for Latent Planning regularizes the angle between consecutive latent displacements~\cite{wang2026temporal}. For $v_t=z_{t+1}-z_t$, its curvature term is $1-\cos(v_t,v_{t+1})$. Equal direction does not require equal speed; a low value does not make Euclidean distance exactly equal to physical reachability cost. A clock-only embedding can also be perfectly straight. Conversely, a meaningful contact transition can bend a trajectory. Thus we test prediction, information retention and time decodability rather than accepting reduced curvature as detector improvement.

\paragraph{Bounded adaptation and controls.}
We conducted a frozen-backbone small-adapter pilot, not a reproduction of the paper's action-conditioned planner. Fixed global/source/context RGB features (1536 dimensions) are projected to 64 dimensions using preprocessing fitted only within each training fold. A small residual adapter and predictor learn next-frame embeddings with a stop-gradient target. The predictor conditions on the current causal state/command-history/nominal-physics context. All variants also use an explicitly additional, unit-weight fixed-teacher anchor loss to discourage collapse. Three whole-configuration DEV folds, two seeds, 300 optimization steps and batch 256 are shared across variants. No opened new-configuration check or future warning TEST is reused.

The initial experiment compared prediction only, cosine weights 0.01 and 0.1, and an ordinary squared-second-difference penalty with weight 0.1. A diagnostic found a held-out context coordinate approximately 103 training standard deviations from its fitted mean, with unstable linear state predictions. A bounded follow-up therefore compared fixed clipping at $\pm3$ training standard deviations and constant-zero context, each with prediction-only and cosine-0.1 variants. These follow-ups were motivated by DEV diagnostics and are exploratory. Raw context remains in the record; clipping is not automatically enabled in the main warning model.

\begin{center}
\begin{minipage}{\linewidth}
\centering
\captionof{table}{Exploratory adapter results, averaged across three DEV folds and two seeds. MSE is measured after decoding predictions into the same fixed teacher coordinates within each fold, preventing a simple latent-scale reduction from masquerading as better prediction. Curvature is $1-\cos(v_t,v_{t+1})$ over moving triplets. Lower is better for both columns; neither is a warning metric.}
\label{tab:straightening-pilot}
\begin{tabular}{llrr}
\toprule
Predictor context & Regularization & Teacher next-frame MSE & Curvature\\
\midrule
Raw & Prediction only & 0.21436 & 1.15446\\
Raw & Cosine 0.01 & 0.21430 & 1.15408\\
Raw & Cosine 0.1 & 0.21387 & 1.15031\\
Raw & Second difference 0.1 & 0.20988 & 1.14950\\
Clipped & Prediction only & 0.17802 & 1.15454\\
Clipped & Cosine 0.1 & 0.17814 & 1.15040\\
Constant zero & Prediction only & 0.18604 & 1.15419\\
Constant zero & Cosine 0.1 & 0.18618 & 1.14985\\
\midrule
\multicolumn{2}{l}{Unchanged-frame persistence control} & 0.20391 & ---\\
\bottomrule
\end{tabular}
\end{minipage}
\end{center}

\paragraph{Interpretation and remaining test.}
Table~\ref{tab:straightening-pilot} retains all eight variants (48 trained adapters in total). Curvature regularization reduced curvature but did not improve the stronger clipped-context or constant-zero-context prediction baselines. Clipped-context prediction-only reduced fixed-teacher next-frame MSE by approximately 12.7\% versus persistence and 17.0\% versus raw-context prediction-only. These are descriptive DEV comparisons, not confirmatory improvements in failure detection. Effective rank remained approximately 35, near the fixed-teacher held-out rank; no rank collapse was detected by this diagnostic. Elapsed-time probe $R^2$ remained approximately 0.54, so straightening did not remove temporal information. Linear probes of simulator source--gripper relative position and latent changes around simulator grasp-flag transitions assess information retention; those privileged quantities are never representation-training inputs or reviewed failure labels. Grasp-flag changes can include normal release.

The justified next step is to compare raw and robust state controls with local vision under reviewed future-event supervision, then freeze preprocessing, model and thresholds before independent warning evaluation. Resource changes must record evidence age and delivery latency; representation changes must retain event-relevant information and beat stronger state controls. Clipping must also be tested for loss of genuine rare-event information rather than assuming outliers are noise. Neither systems throughput nor smoother latents can substitute for that evidence chain.

%% file: detailed_audit_estimates_20261113.tex
\clearpage
\section{Detailed paired audit estimates}
\label{app:detailed-audit}
The main table displays point estimates at a readable single-column size. This table preserves every previously reported interval and the same rounding, selection and bootstrap conditions; no analysis was rerun.
\begin{table}[t]
\centering
\small
\setlength{\tabcolsep}{4pt}
\begin{tabular}{lccccccc}
\hline
 & \multicolumn{2}{c}{FIPER} & \multicolumn{2}{c}{Clock} & \multicolumn{2}{c}{Clock $-$ FIPER [95\% CI]} & Landmark \\
Task & bAcc & TWA & bAcc & TWA & bAcc & TWA & FIPER bAcc \\
\hline
Sorting   & 0.661 & 0.542 & 0.726 & 0.511 & $+0.064$ & $-0.031$ & 0.547 \\ & & & & & [$+0.024$, $+0.105$] & [$-0.066$, $+0.006$] & [0.51, 0.59] \\[1.5pt]
Stacking$^\dagger$ & 0.732 & 0.617 & 0.500 & 0.500 & $-0.232$ & $-0.117$ & 0.601 \\ & & & & & [$-0.262$, $-0.205$] & [$-0.146$, $-0.091$] & [0.57, 0.63] \\[1.5pt]
PushT     & 0.712 & 0.554 & 0.971 & 0.525 & $+0.259$ & $-0.030$ & 0.533 \\ & & & & & [$+0.212$, $+0.305$] & [$-0.082$, $+0.023$] & [0.48, 0.59] \\[1.5pt]
Pretzel   & 0.850 & 0.676 & 0.850 & 0.498 & $+0.000$ & $-0.178$ & 0.669 \\ & & & & & [$-0.150$, $+0.150$] & [$-0.296$, $-0.030$] & [0.49, 0.84] \\[1.5pt]
PushChair & 0.969 & 0.836 & 1.000 & 0.833 & $+0.031$ & $-0.002$ & n/a \\ & & & & & [$+0.000$, $+0.091$] & [$-0.064$, $+0.071$] & \\
\hline
\end{tabular}
\caption{Expanded estimates underlying main Table~2 (exploratory FIPER audit). S-paper selection for each method; 95\% intervals from 2000 rollout-bootstrap draws stratified by outcome within task, selection held fixed. Landmark: balanced accuracy among test rollouts still running at $L=\lfloor 0.25\,T\rfloor$ (n/a: fewer than five successes or failures at risk). The clock's landmark balanced accuracy is 0.5 by construction. $^\dagger$Stacking TWA uses 120 for normalization although the longest test rollout is 100 steps.}
\label{tab:fiper-audit-detailed}
\end{table}

\subsection{Length-only clock ceilings}
\begin{table}[t]
\centering
\resizebox{\columnwidth}{!}{%
\begin{tabular}{lcccccc}
\hline
& & \multicolumn{3}{c}{Balanced accuracy} & \multicolumn{2}{c}{TWA}\\
Task & $D^+$ & ceiling & clock & FIPER & ceiling & FIPER\\
\hline
Sorting & 0.517 & 0.759 & 0.726 & 0.661 & 0.520 & 0.542\\
Stacking & 0.983 & 0.992 & 0.500 & 0.732 & 0.646 & 0.617\\
PushT & 0.990 & 0.995 & 0.971 & 0.712 & 0.570 & 0.554\\
Pretzel & 0.800 & 0.900 & 0.850 & 0.850 & 0.522 & 0.676\\
PushChair & 1.000 & 1.000 & 1.000 & 0.969 & 0.833 & 0.836\\
\hline
\end{tabular}}
\caption{Clock ceilings (main Proposition~1) from test-rollout lengths alone, against the calibrated clock and FIPER under the S-paper settings. Post hoc. The ceiling is not deployable: it picks the step on test lengths.}
\label{tab:supp-clock-ceiling}
\end{table}

%% file: physical_object_memory_20261113.tex
\clearpage
\section{Mechanism probes: object memory, visibility and prediction}
\label{app:object-memory}
\paragraph{Question and evidence boundary.}
This separate study asks whether a small privileged-input predictor assigns greater prediction error to physically impossible than matched possible videos. It uses the Physical Concepts data of Piloto et al.~\cite{piloto2022intuitive}; it is not a reproduction of PLATO, a task-correctness experiment, an EEG measurement or an early-warning result. A physical violation, a rare event, an expectation violation under a specified history, and a task error are different endpoints. We preserve the original exposed-probe negative results and report both helpful and adverse effects of the follow-up.

\paragraph{Preserved original experiment.}
A residual MLP with 64 hidden units was trained for 150 epochs at seeds 11/12/13 on 128 freeform videos, with 32 videos from another freeform training shard for development. Two past region summaries and past/current camera inputs predicted the next region summary: centroid, area, mean RGB and visibility. Inputs use ground-truth masks and associations. All currently visible regions, including background, contribute. The development MSEs are .007159/.007222/.007179 versus .006932 for persistence: none improves the baseline. Twelve quadruplets per concept (60 total, 240 videos) were subsequently scored. Probes never entered training or model selection. Each quadruplet contains two possible and two impossible videos, with matching RGB-frame multisets across the two conditions. Table~\ref{tab:physical-legacy} preserves every seed. Preference is the fraction of quadruplets whose mean impossible-video error exceeds their mean possible-video error, with half credit for exact ties; it is not failure-detection accuracy.

\begin{table}[h]
\centering\small
\begin{tabular}{lrrrr}
\toprule
Concept & Seed 11 & Seed 12 & Seed 13 & Persistence\\
\midrule
Continuity & .667 & .667 & .583 & .667\\
Solidity & .750 & .667 & .750 & 1.000\\
Directional inertia & .750 & .583 & .667 & .750\\
Unchangeableness & .458 & .458 & .500 & .583\\
Object persistence & .083 & .167 & .083 & .000\\
\bottomrule
\end{tabular}
\caption{Original exposed-subset probe results, unchanged. Only 12 quadruplets per concept; the three seeds do not create additional independent stimuli. For continuity and solidity, positive median contrasts coexist with negative mean contrasts at all three seeds: preference alone does not describe error magnitude.}
\label{tab:physical-legacy}
\end{table}

\paragraph{Why test persistent memory?}
The original sampler enumerates only objects visible at the current frame. It has two-frame summaries rather than a persistent hidden-object state. This motivates, but does not prove, a memory-based explanation for weak persistence results. The follow-up retains tracks already observed during a video, including their invisible intervals. It never backfills hidden coordinates from future frames. Labels first appearing in the future can allocate storage, but their features, losses and recurrent updates are masked until first observation. Background label 0 is excluded in the new common-support comparison; label 0 occupies the largest region in 5762/6000 inspected frames. RGB-frame matching is verified, but this does not establish flawless identity correspondence across every spliced mask sequence.

\paragraph{Fixed factorial comparison.}
Four models cross feedforward versus recurrent memory with zero-filled versus visibility-aware training. The feedforward network has 6021 parameters; the gated recurrent unit (GRU) model has 5991. Both use the same two-frame observation/camera inputs and train-only normalization; the recurrent model also carries its hidden state forward. All 12 fits use the original freeform TRAIN/DEV videos, seeds 11/12/13, 150 fixed epochs, AdamW learning rate .001 and track batches of 128. There is no epoch or hyperparameter selection on probes. Zero-filled training penalizes all seven target dimensions, including zeros when invisible. Visibility-aware training separates binary observed visibility from conditional visible-attribute error; invisible attributes carry no supervised target. Visibility is not physical existence. First appearances before a track has ever been observed are outside this endpoint, so it cannot establish comprehensive object creation/disappearance reasoning. A same-weight diagnostic clears recurrent hidden state on currently invisible frames.

The primary common-support score is next-frame visibility negative log likelihood plus visible-target attribute MSE. Known-track errors are averaged within each frame, then across frames and paired videos. A zero-filled score remains a declared sensitivity analysis. This foreground/known-track support differs from the original visible-region/background score: the two tables are not direct numerical before/after replacements.

On the 32-video DEV set, memory reduces the visibility-aware score from .148054 to .131254 under the same visibility-aware objective (11.3\%). However, conditional visible-attribute MSE rises from .003623 to .005279. Reappearance attribute error falls from .052589 to .047373, but only five DEV videos contribute to that endpoint. Clearing memory on hidden frames increases the memory model's DEV score to .145533 and its reappearance error to .103947. These are descriptive diagnostics, not independent mechanism confirmation or a guarantee of better object geometry.

\paragraph{Frozen fresh-shard check.}
After development, hashes of all 12 weights and preprocessing, alongside the primary score, contrast and sample count, were recorded before fetching the first 48 quadruplets per concept from probe shard 00001-of-00020. The check contains 240 new quadruplets/960 videos, all retained; no exact record-payload hash overlaps the original 60. Frame-multiset matching holds for all 240. This fixed prefix is not a representative random sample, and unknown generator-template overlap is not ruled out. No fitting or result-dependent extension occurs. Scores are aggregated across seeds within quadruplet, then equally across concepts.

\begin{table}[h]
\centering\small
\setlength{\tabcolsep}{5pt}
\begin{tabular}{lrrrrrr}
\toprule
Concept & FF/zero & FF/visibility & Memory/zero & Memory/visibility & Persistence & Constant velocity\\
\midrule
Continuity & .000 & .257 & .701 & .646 & .177 & .458\\
Solidity & .833 & .729 & 1.000 & .924 & .490 & 1.000\\
Directional inertia & .049 & .063 & .083 & .125 & .531 & 1.000\\
Unchangeableness & .500 & .517 & .986 & .986 & .500 & 1.000\\
Object persistence & .021 & .069 & .118 & .319 & .000 & .021\\
\midrule
Equal-concept mean & .281 & .327 & .578 & .600 & .340 & .696\\
\bottomrule
\end{tabular}
\caption{Fixed fresh-shard physical paired preference. FF denotes feedforward. Every method uses the same known-track support and visibility-aware score; model entries average three seeds. Constant velocity extrapolates the last observed object displacement and retains attributes; both motion baselines use fixed visibility probabilities .99/.01 according to current visibility, not calibrated probabilities.}
\label{tab:physical-memory-fresh}
\end{table}

The prespecified memory/visibility versus FF/zero contrast is +.319 [conditional 95\% CI: .278,.365]. The same-objective memory contrast versus FF/visibility is +.273 [.228,.322]. Visibility-aware training beyond memory/zero is only +.022 [-.019,.065]; the joint candidate remains below constant velocity by -.096 [-.144,-.049]. Intervals use 2000 concept-stratified paired quadruplet-bootstrap draws, conditional on fitted models and assuming quadruplet independence within concept; training and unknown template-cluster uncertainty are excluded. Object persistence (.319) and inertia (.125) remain below .5. These results support a bounded architectural comparison, not general physical understanding or superiority to motion controls.

\paragraph{Reproducibility and next boundary.}
Five causal-software tests cover future-prefix invariance, unseen future objects, hidden-target masking, retained hidden tracks and the reset control. An initial report-serialization failure occurred before fitting; its log is retained, and the corrected run uses a new identifier. Twelve fits completed; 180 exposed-probe metric cells were recomputed from saved per-track predictions, and 70 fresh concept/seed cells were recomputed from saved scores. Original data/result hashes remained unchanged. These checks are not independent retraining. A future task-conditioned oddball study must separately manipulate history and correctness, retain identical target clips across contexts, and measure pre-event warning before crediting deployment benefit. The present probes remain separate from the main duration-attribution audit.

\paragraph{Motion-anchored follow-up: development is not probe success.}
The exposed checks above motivated a follow-up that preserves a causal constant-velocity attribute anchor and learns either visibility corrections alone or both visibility and attribute residuals. Each GRU has 7431 parameters and receives the original 26 features plus 15 causal motion/memory features: the six predicted attributes, six last-seen attributes, two velocity components and time since last observation. The output correction starts at zero. A first six-fit run on the original 128 freeform videos had worse DEV prediction scores (.167938 for visibility-only and .170326 for attribute residuals) than constant velocity (.137228); all fits and results are retained, and no new probe was opened for that run.

The next bounded run adds 512 freeform training videos from training shard 00002, with exact record-hash disjointness from the original TRAIN/DEV records, for 640 training videos total. It retains the original 32-video DEV set, three seeds, learning rate .001, batch size 128 and a maximum of 150 epochs. Residual-attribute training adds a .1 squared-correction penalty. Checkpoints at epochs 25/50/75/100/125/150 are selected per seed using ordinary DEV next-frame prediction score, never physical-probe preference. A common correction multiplier per arm is chosen from $\{0,.01,.03,.1,.3,1\}$ by mean DEV score. Both arms select multiplier 1, and all six fits select epoch 25. The residual arm is selected as primary: DEV score .101135 versus .101737 for visibility-only. For a stronger control, constant-velocity visibility logits are calibrated on DEV over temperature $\{.25,.5,1,2,4\}$ and bias $\{-2,0,2\}$; the selected setting is the original temperature 1, bias 0. The raw control is retained. Training size and selection rules changed together, so this is not an isolated architecture ablation.

Before fetching probe shard 00002, the selected arm, score, sample count, code, preprocessing and all new fitted-weight hashes are frozen. The next check again contains exactly 48 quadruplets per concept, 240 total/960 videos, with no record-hash overlap with the 300 earlier exposed quadruplets. All candidates are evaluated without further selection or extension, using the unchanged visibility-aware score and support. Table~\ref{tab:physical-motion-fresh} compares every method on this new cohort; its numbers must not be subtracted from Table~\ref{tab:physical-memory-fresh} as if they shared stimuli.

\begin{table}[h]
\centering\small
\setlength{\tabcolsep}{5pt}
\begin{tabular}{lrrrrr}
\toprule
Concept & Previous FF & Previous memory & Motion/visibility & Motion/residual$^*$ & Constant velocity\\
\midrule
Continuity & .007 & .549 & .910 & .896 & .458\\
Solidity & .875 & .965 & 1.000 & 1.000 & 1.000\\
Directional inertia & .069 & .181 & .438 & .382 & 1.000\\
Unchangeableness & .497 & 1.000 & .951 & .924 & 1.000\\
Object persistence & .007 & .229 & .167 & .167 & .000\\
\midrule
Equal-concept mean & .291 & .585 & .693 & .674 & .692\\
\bottomrule
\end{tabular}
\caption{Next frozen-shard physical paired preference, with three-seed model averages. $^*$ denotes the arm selected on DEV before this check, not the best arm after viewing probe results. The raw and DEV-calibrated constant-velocity controls coincide. Persistence is .358 overall; all per-seed scores are retained.}
\label{tab:physical-motion-fresh}
\end{table}

The primary motion/residual contrast versus calibrated constant velocity is -.018 [conditional 95\% CI: -.061,.025], while its contrast versus the previous frozen memory model is +.089 [.042,.133]. Its seed-wise overall preferences are .642/.675/.704. The secondary motion/visibility arm scores .683/.671/.725 across seeds and .693 overall, only +.001 [-.042,.043] versus constant velocity: this is not established superiority, equivalence, or a post-test replacement of the primary arm. Previous-memory seed means are .613/.567/.575. These intervals use the same conditional paired bootstrap and limitations as above, without multiplicity adjustment. Lower ordinary prediction loss therefore does not automatically imply better physical-violation discrimination.

A post-test additive-score diagnostic finds that visibility NLL reverses an otherwise positive attribute contrast in all 81 inertia seed/quadruplet instances where constant velocity is correct but motion/visibility is wrong. For motion/residual, this occurs in 77 of 89 such instances. These are descriptive seed/stimulus instances, not additional independent quadruplets or causal interventions. They motivate separating motion and visibility evidence in future development; the present test score and primary selection remain unchanged. Ten software tests pass, and 75 saved model/concept score cells are independently recomputed in NumPy float64 (maximum video-score difference below $10^{-7}$), including checks that the motion anchor reproduces the original constant-velocity implementation. All six new fitted hashes match their prefetch commitment. No robot-warning, neural or closed-loop benefit follows from this follow-up.

A further bounded DEV experiment adds image-mask shape and a pooled message from previously observed neighbors, including hidden neighbors. Nine matched-architecture fits (three arms, three seeds) and three regularized readouts use the same 640 normal TRAIN videos, 32 ordinary DEV videos and exposed physical DEV quadruplets. Best eligible paired preferences are .736 for motion only, .563 with own shape and .538 with persistent-neighbor summaries, versus .783 for tuned constant velocity. The unchanged score and common support do not rescue these representations; no new probe is opened.

\subsection{Independent flexible-object and selective-review diagnostics}
\label{app:flexible-mechanism}
\paragraph{Different endpoints and provenance.}
The physical quadruplets above test relative error on matched possible/impossible videos. The following independent MuJoCo 3.3.7 cable simulations instead test normal-deformation forecasting and post-perturbation review timing. Neither supplies human-reviewed robot failure onsets or evidence of better robot task outcomes. The simulated cable, virtual tip forces and camera are independently specified; these are not external demonstration data. Bending and density values are known simulator settings, not measured material properties. A render-only panel changes visibility without contact forces. Held-out configurations vary within the same clamped-cable family and fixed camera; they are not unseen real objects.

\paragraph{Normal-deformation forecasting.}
Thirty configurations are split before collection into 18 TRAIN, 6 DEV and 6 TEST, with three episodes per configuration (90 episodes, 7200 frames). Frames and known force commands are recorded every 20\,ms; the prediction horizon is 100\,ms. Eligible decisions use a force command already committed over that horizon. A shared TRAIN-fitted, then frozen RGB-D encoder supplies estimated centerlines and visibility; true centerlines are targets, not forecast inputs. Three arms have the same registered architecture and training budget, with three seeds each: no material parameters, material parameters, and material parameters plus persistent memory. Checkpoints and baseline configurations are chosen on DEV; fitted hashes are recorded before loading TEST arrays. Table~\ref{tab:flexible-forecast} reports the unchanged primary endpoint: mean 3D centerline point error, averaged over decisions and episodes, then configurations. Mean seed performance is not an ensemble prediction.

\begin{table}[ht]
\centering\small
\begin{tabular}{lr}
\toprule
Predictor & TEST error (mm)\\
\midrule
RGB-D, without material parameters & 5.721\\
With material parameters & 3.825\\
With material parameters and memory & 3.475\\
Persistence of estimated shape & 6.395\\
Estimated constant velocity & 9.425\\
Small-deflection beam equilibrium & 3.311\\
Beam relaxation (DEV-selected) & 4.115\\
Ridge regression, same full features & 2.744\\
\bottomrule
\end{tabular}
\caption{Normal simulated deformation: lower point error is better. Learned rows average three seeds on six held-out configurations/18 episodes. Ridge is the strongest comparator selected on DEV, not on TEST. It and the beam outperform the memory network on this endpoint.}
\label{tab:flexible-forecast}
\end{table}

Memory minus material-only error is $-.350$\,mm [conditional 95\% CI: $-.534,-.182$]; memory minus the DEV-selected ridge comparator is $+.731$\,mm [$+.002,+1.607$], where positive means worse. These intervals resample six configurations, conditional on the shared encoder and fitted seeds; fitting and simulator-fidelity uncertainty are excluded. Future-visibility Brier score improves from .0497 to .0297 with memory, but normal 10\,mm realized forecast-residual exceedance rises from 5.75\% to 6.51\%. Ridge has no such exceedance on this small test. These exceedances are retrospective forecast errors, not detector false-positive rates. Better visibility prediction therefore does not establish fewer spurious warnings. Saved predictions, raw-target alignment and fitted hashes were checked; independent float64 recomputation changes reported mean errors by less than $4.1\times10^{-7}$\,mm.

A later TRAIN/DEV-only experiment uses a nominal beam anchor, causal command history and the same frozen RGB-D encoder. Across three seeds, full-memory DEV error is 2.068\,mm versus 2.087 for a same-input feedforward arm, 2.464 for full-input ridge, 3.952 for physics-only memory and 3.711 for the original frozen ridge on these same DEV episodes. The combined pipeline improves prediction, but changing inputs and anchor together prevents isolated attribution. Memory's incremental .93\% improvement fails the prespecified 5\% engineering gate. The original TEST above is not reused and no new TEST is collected.

\paragraph{Selective-review pilot.}
A separate cohort has 48 new configurations and four paired cells per configuration: external-force injection versus sham, crossed with clear versus occluded views (192 episodes). Known injection times vary within the declared range; sham and injection share the occluder schedule. Sixteen configurations train trigger readouts, eight select thresholds and timer phase, and 24 form TEST (96 episodes, 48 injections). Frozen predictors supply only causal image-derived estimates, commands, time and declared material/geometry metadata. State controls include estimated full centerline and velocity: they are stronger than sensor-free clocks or ordinary robot proprioception. Neither injected-force metadata nor true centerlines enter scores.

Each method can request at most two reviews. An ideal reviewer identifies an injected perturbation if the request uses a post-injection observation and its response arrives within 300\,ms of injection. Response delay is assumed to be 100\,ms, not measured on an LLM. Capture is post-onset detection, not early warning; requests do not change the replayed dynamics. The DEV selection objective is capture minus .10 times mean requests. Table~\ref{tab:selective-review-pilot} retains every first-round trigger. The nonlinear state head uses fixed gradient boosting; other fitted readouts use regularized logistic regression. Thus this round does not isolate every readout-capacity effect.

\begin{table}[h]
\centering\small
\begin{tabular}{lrr}
\toprule
Trigger & Timely capture (\%) & Mean requests\\
\midrule
DEV-selected periodic review & 58.33 & 2.000\\
Time/estimated state, linear & 45.83 & 1.156\\
Time/estimated state, nonlinear & 66.67 & 1.198\\
State plus simple ridge predictor & 72.92 & 1.604\\
State plus frozen memory predictor & 66.67 & 1.427\\
\bottomrule
\end{tabular}
\caption{Controlled-force review timing on 24 held-out configurations. The maximum request budget is common; actual costs differ. The memory trigger does not exceed the stronger controls. Ideal review and an assumed delay are not an LLM or robot closed loop.}
\label{tab:selective-review-pilot}
\end{table}

The prespecified advancement rule requires at least .10 greater capture, a paired configuration-bootstrap lower bound above zero, and a request-count difference upper bound no greater than zero versus the strongest DEV-selected control. It fails: memory ties the nonlinear state control's capture while using more requests. Memory requests review in 77.08\% of sham episodes. Four of its 16 missed injections exhaust the budget before injection; scheduling is a diagnostic contributor, not a tested repair. The joint truth manifest, including TEST metadata, was parsed at startup; TEST labels were joined only after head/threshold freeze and were not used in selection. We do not claim that all TEST metadata remained unread. Subsequent loaders require split-specific truth files.

\paragraph{Capacity-matched DEV follow-up and disposition.}
A bounded follow-up uses the same fixed nonlinear readout for all fitted arms and adds current CNN features and their causal five-frame differences. It uses only the original trigger TRAIN/DEV configurations. The selected visual-plus-ridge candidate has 81.25\% DEV capture with .750 requests, versus 87.50\% and .844 for the strongest state-plus-ridge control. This is a tradeoff, not superiority; the DEV advancement gate fails and no new TEST is collected. Raw pairing checks verify that the panel does not change physics and injection/sham prefixes match before injection; saved first-round schedules, budgets, hashes and paired intervals are recomputed. No robot event labels are supplied by these simulations. These studies retain both useful predictive changes and failed stronger-control comparisons as mechanism exploration; they do not support a robot-warning, neural-mechanism or intervention claim.

A separate DEV-only trigger comparison freezes these new predictors before fitting seven capacity-matched readouts. Visual-plus-beam-memory capture is 87.50\% with .750 requests, versus 87.50\%/.844 for state plus the old ridge and 100\%/1.781 for state plus the physics-only beam predictor. The last control maximizes the fixed utility. The joint superiority/no-added-cost gate fails, despite a useful cost tradeoff at selected DEV operating points. These eight DEV configurations are repeatedly exposed development data; no new TEST or closed loop follows. Improved normal forecasting therefore has not established independent visual early-warning gain.

\paragraph{Symmetric threshold search revises the cost interpretation.}
The apparent reduction from .844 to .750 requests at 87.5\% capture compares individually selected points on a coarse threshold grid. Giving every frozen readout all attainable scalar threshold cuts, with identical eligibility, cap and cooldown, changes the minimum cost at that capture to .625 for state plus the old ridge and .688 for the previous visual-memory candidate. The apparent 11\% saving is therefore not a robust method advantage. This DEV diagnostic also includes zero-, one- and two-query fixed schedules; fractional randomized-policy envelopes, when computed, are expected-cost summaries, not realized budgets.

\paragraph{Matched factors and stronger history controls.}
Twenty-four further fits cross output anchor (constant velocity or nominal beam), explicit ten-command lag features, and recurrent versus reset state, at three seeds. All register 54,700 parameters, receive the same 204 slots and use the same normal TRAIN/DEV and training budget; ablated slots are zeroed. Beam features remain available in both anchor arms. With memory and no explicit lag slots, beam-anchored DEV error is 1.965\,mm versus 3.518 for the matched constant-velocity anchor. Additional lag slots worsen the beam-memory result to 2.068\,mm. Six further control fits mask either just raw CNN features or all image-derived features. Retaining only estimated kinematic history, visibility and known physical inputs gives 1.871\,mm, versus 4.010 for the physics-only predictor. These descriptive DEV results motivate a simpler predictor, without establishing additional raw-visual information or real-object generalization.

\paragraph{Confirmation-aware scheduling remains a separate endpoint.}
A further ideal-reviewer diagnostic stops requests only after a positive response has arrived, with the same assumed 100\,ms delay for every method. It never uses a future response to cancel a current request. Thirteen frozen readouts receive identical exact-threshold searches. Under the .75 mean-call DEV ceiling, the selected extra-CNN candidate obtains 87.50\% capture at .625 requests; the strengthened kinematic-history predictor control obtains 93.75\% at .719. The primary visual candidate again fails the engineering filter. These repeatedly exposed eight DEV configurations do not confirm a visual-warning or actual-LLM cost gain; no new TEST is collected and prior failed gates remain unchanged.

\paragraph{Spiking temporal heads: a retained negative comparison.}
A bounded TRAIN/DEV follow-up compares leaky integrate-and-fire (LIF) spiking neural networks with analog heads and gated recurrent units (GRUs), holding the RGB-D frontend and 120 causal input slots fixed. A factorial design varies absolute versus difference encoding, fixed versus learned time constants, and fixed versus learned thresholds, with three seeds and 100-epoch budgets (54 new fits). Learnable membrane constants follow the motivation of Fang et al.~\cite{fang2021learnable}, rather than reproduce their classification experiments. Combined difference encoding and learned constants/thresholds reduce LIF DEV error from 2.688 to 2.088\,mm under the matched budget. However, the strongest previously frozen GRU obtains 1.871\,mm; tuned LIF also has higher occluded error (2.419 versus 2.292\,mm) and nonzero normal 10\,mm residual exceedance. Its engineering gate fails. Checkpoints and arms were selected on the exposed six DEV configurations; no new TEST, event-risk training, neural-mechanism validation or measured energy benefit follows.

%% file: evidence_ladder_diagnostics_20261114.tex
\clearpage
\section{Portable audits and bounded development diagnostics}
\label{app:portable-diagnostics}
\paragraph{Evidence status.}
The analyses below reuse exposed releases or development episodes. They improve reproducibility and identify failure modes; they are not new independent validation of visual early warning. The three evidence levels in the main text remain distinct. In particular, the new 48-episode grasp-loss review cohort has no eligible imported full-episode human labels at this writing, so no corresponding event-risk model, CAL threshold or TEST comparison is fitted. Earlier author-confirmed annotations have a different protocol and remain separate. No machine event proposal is promoted to human-reviewed truth.

\subsection{A portable native-alarm and reviewed-onset interface}
The duration interface accepts episode identity, task, grouping, outcome, observed length, declared horizon and each method's native first-alarm index. It preserves imported alarm rules rather than replacing every method's calibration with a generic threshold. Optional continuous score prefixes enable AUROC; absent scores are not imputed from binary alarms. Reports distinguish all-episode common-prefix truncation $[0,T_{\min})$ from strict landmark populations $T_i>L$, retain invalid task cells, and state the normalization used for TWA. A length-only maximum is labeled descriptive, not a deployable calibration rule.

On the 1,540 original FIPER episodes, 50 seed/quantile imports reproduce 150 preserved full-rollout and landmark reference cells to $1.2\times10^{-16}$. The replicates are not additional independent episodes. A second check compares 7,510 metric cells, including the unchanged shortest-prefix results. Their largest difference is zero. These are numerical reproduction checks, not an independent replication of the original detector training.

The separate reviewed-onset interface requires complete episode review, explicit present/absent/uncertain event status, common source windows, aligned clock domains, and arm-specific input-readiness, decision and delivery times. A missing image uses the declared state/physics fallback exactly; the whole streaming rule still needs separate calibration. If source time $t$ predicts within $h$, delivered time $d$ must satisfy both $d+\delta\le a$ and $b\le t+h$ for conservative credit on reviewed onset interval $[a,b]$. Latency does not extend the forecast origin. Unknown onsets, expired forecasts and state-fallback alarms remain visible. The supplied implementation uses maxima over negative configuration groups for its finite-sample CAL threshold, targeting group alarm probability under exchangeability; it separately reports realized TEST episode and group burdens. Hashes and boolean gate declarations check supplied-record consistency, not actual human attention or historical freeze authenticity.

\subsection{Setting reselection and a prior-control comparison}
The original confidence intervals fix settings. A new exploratory reanalysis resamples original episodes within task, recorded outcome and fixed filename-hash fold, using 5,000 paired draws. Model seeds and calibration quantiles are averaged. Each draw reselects the native style/window by the same across-task mean TWA objective; detector fitting and calibration remain fixed. Acquisition clustering is unknown. Direct repetition of the native episode arrays verifies 1,617 metric cells from 11 declared draw indices to $8.9\times10^{-16}$.

Sorting was chosen because its earlier ranking reversal was already known. Within each twofold split, a method is selected using the full-rollout or quarter-risk objective on one fold and evaluated on the other fold's identical quarter-risk population. Full-rollout selection chooses clock or action RND; quarter-risk selection chooses RND-OE or action RND. The mean held-out quarter bAcc difference is $+.0449$ [exploratory 95\% CI: $.0022,.1077$]. This is an exposed-data procedure-sensitivity result, not untouched validation or intervention utility.

A post-hoc comparison, added after inspecting that positive result, gives the prior shortest-prefix control the same selection opportunity. The prefix is the shortest length in the selection fold only, recomputed in every bootstrap selection sample; held-out lengths never select it. Shortest-prefix and quarter-risk selection choose the same methods at the point estimate: quarter minus shortest-prefix bAcc is $0$ [$-.0363,.0306$]. Thus this example does \emph{not} establish an advantage over prior truncation. It supports horizon-dependent selection and retains a concrete limit on the claimed incremental benefit.

\subsection{Frozen visual-input sensitivity on exposed DEV}
The frozen heads of Supplement~S are evaluated on the original 640 DEV episodes at three fixed landmarks (1,920 source windows). State/action inputs, heads, reference bank and historical thresholds are unchanged. Paired conditions replace the detector's extra image with an unflagged black image, an image five control steps old, or a flagged missing image with exact state-score fallback. On Eggplant, a fixed purple-color heuristic also defines a source rectangle and an equal-area nonoverlapping control rectangle. This is neither verified object segmentation nor a guaranteed empty-background control. Localization failures are retained in coverage counts (15 Eggplant source windows); region comparisons use the same available pairs.

\begin{table}[ht]
\centering\small
\begin{tabular}{lrrr}
\toprule
Quarter-landmark input change & Paired windows & Mean absolute score change & Alarm flips (\%)\\
\midrule
Unflagged black image & 546 & .2174 & 28.21\\
Five-control-step stale image & 546 & .0154 & 2.56\\
Flagged missing image; state fallback & 546 & .0207 & 2.56\\
Purple source rectangle & 80 & .0391 & 1.25\\
Matched non-source rectangle & 80 & .0080 & 0.00\\
\bottomrule
\end{tabular}
\caption{Input sensitivity on the native-outcome quarter-risk DEV population. Changes are relative to the same single-frame clean input. Alarm flips are not improvements or false-positive rates. Native success-geometry limitations remain; no physical onset is scored. The diagnostic fallback retains the old visual threshold and is not a newly calibrated streaming policy.}
\label{tab:dev-input-sensitivity}
\end{table}

The source-minus-control difference in absolute score change is $.0311$ [$ .0243,.0376$], using a paired conditional bootstrap of 47 configuration groups. This suggests differential sensitivity under the chosen masks, not useful causal reliance on object evidence. Unflagged blanking can cause a much larger harmful response. An independent NumPy forward calculation checks 36,477 numeric/decision cells, with maximum score difference below $10^{-15}$. Changing the historical batch encoder to this single-frame path flips one original threshold decision across all 1,920 windows; this path therefore cannot inherit deployment calibration without checking. No weight/threshold search, new TEST, reviewed failure-onset evaluation or robot response occurs in this diagnostic.

\subsection{Four-hour warm software replay}
A predeclared four-hour run paces the original DEV corpus at 5\,Hz through a bounded producer queue, a real single-frame ResNet-18/state worker, and a local receipt sink. The run completes in 14,401.75\,s: 72,001 source frames cover all 640 recorded episode identities, with repetition. These are zero new independent robot episodes. Scores are emitted only at the heads' three trained landmarks (2,760 paired state/visual outputs); the other 69,241 frames supply feature-readiness telemetry, not warning predictions. State packets are emitted and received before visual packets at every landmark. Independent trace recomputation checks 299,098 timing cells exactly, including chronology, saved summaries, source/model/code hashes and five missing-image landmark fallbacks.

At trained landmarks, actual-release-to-sink latency is 1.93/3.32\,ms (median/99th percentile) for state/action and 4.32/8.04\,ms for visual output. Scheduled-source-to-sink visual latency is 4.69/8.50\,ms, with maximum 11.42\,ms. However, the complete source stream has \textbf{11 scheduled 200\,ms deadline misses}, with maximum 463.97\,ms; none occurs at a trained landmark. Post-release latency alone would hide this producer-scheduling tail. No message is lost and the worker completes, but this does not establish a uniformly met input deadline.

During the second half, 140 missing, 140 unflagged black and 140 five-step-stale images are injected without changing state/action history. The timing run never scores their warning accuracy. Warmup, physical acquisition, remote transport, actor inference, display, review and actuation are excluded. CPU verification and low-frequency resource monitoring are logged concurrent loads, so this is shared-machine timing. Peak post-warmup PyTorch-allocated GPU memory is 68.2\,MiB; this is not total deployment memory. Later device-wide memory readings range from 368 to 4,738\,MiB; these are not model-only allocations. This partial-run monitor is not a full-run energy measurement. Neither the 5\,Hz replay nor its fast landmark outputs establishes 50\,Hz robot control, human-reviewed pre-event lead or useful visual warning.

\begin{figure}[ht]
\centering
\includegraphics[width=.96\textwidth]{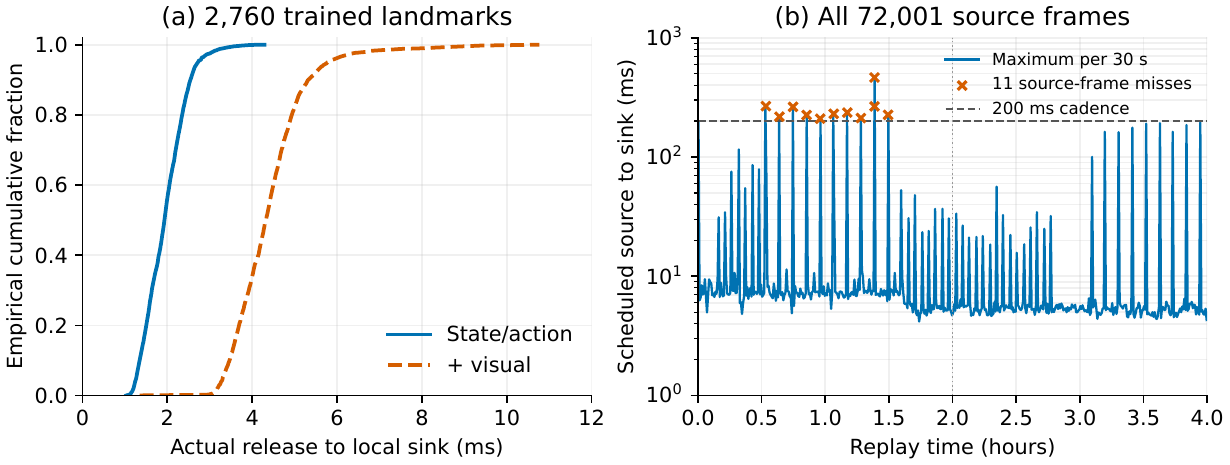}
\caption{Warm software timing, not physical warning performance. Left: actual local delivery at the 2,760 trained landmarks. Right: all source frames, showing each 30-second maximum and every scheduled-input deadline miss. The dotted vertical line marks the start of diagnostic image faults, not a randomized comparison. Replayed frames and repeated episodes are not independent trial counts.}
\label{fig:software-soak}
\end{figure}

%% file: scope_monitor_signals_20261114.tex
\clearpage
\section{Scope: which monitor signals this audit covers}
\label{sec:supp-scope}
The audit can be adapted to timestamped score streams when the required outcomes, rollout limits, calibration data and alarm rules are available; A3 additionally needs the candidate settings and selection criterion. Which checks can be reconstructed depends on the release. Table~\ref{tab:supp-scope} summarizes the evidence here, rather than an exclusive taxonomy or a claim of useful early warning. ``Not covered'' makes no claim for or against that signal family. Bare numeric section, figure and equation references are to the main paper; supplementary sections use letters and supplementary figures and equations use an S prefix.

\begin{table}[!ht]
\centering
\small
\setlength{\tabcolsep}{5pt}
\renewcommand{\arraystretch}{1.25}
\begin{tabular}{@{}p{0.19\linewidth}p{0.22\linewidth}p{0.25\linewidth}p{0.24\linewidth}@{}}
\toprule
\raggedright \textbf{Signal family} &
\raggedright \textbf{Examples here} &
\raggedright \textbf{How duration can enter} &
\raggedright \textbf{Status in this paper} \tabularnewline
\midrule
\raggedright Policy-associated scores and external VLM judgments &
\raggedright FIPER RND-OE and ACE~\cite{romer2025fiper}; STAC~\cite{agia2024stac}; SAFE token metrics~\cite{gu2025safe}; released VLM verdicts &
\raggedright Outcome-dependent exposure; cumulative aggregation; time or task-progress information in context &
\raggedright Offline release-specific checks (Secs.~5.1--5.3). Rank and task-mixture extensions are exploratory. SAFE learned detectors are not rerun; VLM judgments are not policy-internal features. \tabularnewline
\raggedright State-space prediction surprise &
\raggedright Dens state-transition score (Sec.~4.1) &
\raggedright Residual dependence on task phase and windowing; outcome-dependent observation lengths &
\raggedright Prespecified Octo contrast versus task/time/kinematics: $\Delta$AUC $+.0099$, below the $.02$ target (Sec.~5.4). Classifiers are cross-validated within confirmation, not transferred frozen. \tabularnewline
\raggedright Image-space novelty or distance &
\raggedright GR00T image-distance proxy; frozen ResNet-18 nearest-neighbour OOD score &
\raggedright Outcome-dependent exposure; reference-bank distance may reflect task progress or elapsed time &
\raggedright Octo secondary OOD contrast $+.0257$, without multiplicity control; GR00T landmark results exploratory (Fig.~2c). TEST-1 (416 episodes) fails its risk-set gate (Supp.~S). Separate TEST-2 (1,248 episodes) yields $+.0035$ [$+.0007,+.0062$], below the $.02$ target, on Carrot/Spoon at 24 fixed configurations each (Supp.~\ref{sec:prospective-test2}). \tabularnewline
\raggedright Pixel-space world model (WM) or world-action model (WAM) prediction error &
\raggedright Predicted versus later observed images; no tested instance here &
\raggedright Compounded forecast or context effects may affect scores; forecast span and elapsed episode time are distinct &
\raggedright \textbf{Not covered.} Match clock and monitor update opportunities (e.g., chunks). Credit an error only after the matching real image arrives and comparison completes (Eq.~1). \tabularnewline
\bottomrule
\end{tabular}
\caption{Evidence coverage by monitor-signal family. Coverage does not establish a method's usefulness, and an uncovered family is not shown to fail. The rows summarize the studies above, not a new evaluation or a mutually exclusive classification.}
\label{tab:supp-scope}
\end{table}

For variable-duration chunks, retain an elapsed-time reference as well as chunk index. Fixed-span re-anchoring may reduce compounded forecast drift, but does not establish independence from episode time (Supp.~T).

%% file: prospective_test2_20261116.tex
\section{Prospective TEST-2: bounded visual increment}
\label{sec:prospective-test2}

\paragraph{Frozen scope.}
TEST-2 was designed after the earlier 416-episode frozen-monitor test and is reported separately rather than pooled with it. The study launched 1,248 new GR00T rollouts: 576 for Carrot and 672 for Spoon. Every rollout used a fixed 60-step horizon; first-success time was stored separately. Inference is limited to these two tasks and their 24 fixed configurations per task. Eggplant and StackCube were excluded from TEST-2 using DEV evidence before the test videos or outcomes were inspected.

\paragraph{Primary result.}
At the frozen quarter-horizon risk set, the state/action model achieved macro AUROC .77172 and the model adding visual novelty achieved .77519. The paired macro increment was +.00347 (within-configuration episode-bootstrap 95\% CI [.00073, .00615], 5,000 draws). Each bootstrap draw resamples episodes within every fixed configuration, preserving all 24 configurations per task and then reapplying the risk-set rule; the two task increments receive equal weight. The interval is conditional on the frozen heads, reference bank and configuration set. The lower confidence bound exceeds zero, satisfying the frozen \emph{statistical-positive} rule. The point estimate is nevertheless well below the prespecified +.02 practical-effect target, so the prespecified \emph{confirmatory-positive} rule is not met. This is evidence for a small bounded predictive increment, not evidence of practically useful early warning, reviewed failure-onset lead time, visual reliance, or intervention benefit.

\begin{table}[h]
\centering
\small
\begin{tabular}{lrrrrr}
\toprule
Task & Launched & Risk set & State/action & +Visual & $\Delta$AUROC \\
\midrule
Carrot & 576 & 576 & .80565 & .80433 & $-.00131$ \\
Spoon  & 672 & 623 & .73780 & .74604 & $+.00825$ \\
\midrule
Macro & -- & -- & .77172 & .77519 & $+.00347$ \\
\bottomrule
\end{tabular}
\caption{Frozen TEST-2 quarter-horizon results. Carrot contains 260/316 and Spoon 441/182 successes/failures in the eligible risk sets. Macro averaging gives tasks equal weight.}
\label{tab:test2}
\end{table}

\paragraph{Procedural disclosures.}
The exact freeze file existed before launch, but the corresponding repository revision was pushed after launch; the remote Git timestamp is therefore post-launch. The same-environment preflight covered policy-server startup, a fixed synthetic batch-size-16 request, shutdown, and cleanup. It did not cover the simulator, embedding extraction, score preparation, or analysis. A first launch failed during policy-server initialization before any rollout (0/1,248 episodes) and was retained as INCOMPLETE. After an infrastructure-only amendment (local model resolution and idempotent socket cleanup; tasks, seeds, configurations, horizon and analysis unchanged) and a passing preflight, a single second launch produced all reported episodes; no episode was replaced or rerun. The primary and independent analysis outputs were byte-identical (SHA-256 \texttt{2337c700f9e75d5c62520e10a92da603f34b6e200bf464b6ed9d7d6014f4ecd6}).

\paragraph{Interpretation.}
The task-level signs differ: visual input is slightly worse on Carrot and better on Spoon. The macro interval therefore supports only a small average increment over this fixed task/configuration set. TEST-2 narrows the uncertainty left by TEST-1 but does not establish that the visual signal is early enough to act on, causal with respect to vision, or transferable to new tasks, policies, configurations, or real robots.

The frozen visual pipeline therefore provides a small average predictive increment on the specified tasks and configurations, but does not meet the prespecified practical target. Statistical positivity, useful advance warning and intervention benefit remain separate claims.